\documentclass{article}

\usepackage[english]{babel}
\usepackage[utf8]{inputenc}

\usepackage{color,xcolor,ucs}

\usepackage{floatrow}
\usepackage{tabularx}
\usepackage{booktabs}
\usepackage{float}
\usepackage{amsfonts}
\usepackage{helvet}         
\usepackage{courier}        
\usepackage{type1cm}        
\usepackage{amsmath}
\usepackage{amssymb}
\usepackage{makeidx}         
\usepackage{comment}         
\usepackage{graphicx}        
\usepackage{multicol}        
\usepackage[bottom]{footmisc}
\usepackage{bm}

\usepackage{cite}
\usepackage{url}

\usepackage[unicode=true, bookmarks=true,colorlinks=true]{hyperref}
\usepackage{xr-hyper}

\usepackage{epsfig}
\usepackage{epstopdf}
\usepackage{caption}
\usepackage{subcaption}
\usepackage{xspace}
\usepackage{rotating}

\usepackage{tikz}
\usepackage{tkz-graph}
\usetikzlibrary{arrows,shapes,shadows,positioning,calc}

\usepackage{graphicx}
\usepackage{threeparttable}
\usepackage{multirow}
\usepackage[font=scriptsize,labelfont=bf]{caption}

\usepackage[margin=0.75in,bottom=0.85in]{geometry}

\usepackage{algpseudocode,algorithm,algorithmicx}
\usepackage{pdfpages}

\definecolor{vadim}{rgb}{0.0,0.0,1.0}
\definecolor{vova}{rgb}{0.05,0.5,0.05}

\newcommand{\prob}[1]{\ensuremath{{\mathbb{P}({#1})}}}

\algrenewcommand\algorithmicrequire{\textbf{Input:}}
\algnewcommand{\LineComment}[1]{\State \(\triangleright\) #1}

\newcommand{\poses}{\ensuremath{{\cal X}}\xspace}
\newcommand{\observations}{\ensuremath{{\cal Z}}\xspace}

\newcommand{\classes}{\ensuremath{{C}}\xspace}

\newcommand{\likelihoods}{\ensuremath{{\cal L}}\xspace}
\newcommand{\motionmodel}{\ensuremath{{\cal M}}\xspace}

\newcommand{\hisnew}{\ensuremath{{\Delta \cal H}}\xspace}

\newcommand{\his}{\ensuremath{{\cal H}}\xspace}

\newcommand{\appropto}{\mathrel{\vcenter{
			\offinterlineskip\halign{\hfil$##$\cr
				\propto\cr\noalign{\kern2pt}\sim\cr\noalign{\kern-2pt}}}}}

\title{Distributed Consistent Multi-Robot Semantic Localization and Mapping}

\author{Vladimir Tchuiev$^1$ and Vadim Indelman$^2$ 
	\thanks{The authors are with the Department of Aerospace Engineering, Technion - Israel Institute of Technology, Haifa 32000, Israel. {\tt\{vovatch$^1$, vadim.indelman$^2$\}@technion.ac.il}.
	This work was partially supported by the Israel Ministry of Science \& Technology (MOST).}
}

\date{}

\graphicspath{{figures/}}

\makeatletter
\def\endthebibliography{%
	\def\@noitemerr{\@latex@warning{Empty `thebibliography' environment}}%
	\endlist
}
\makeatother

\begin{document}
	
	\maketitle
	
	

\begin{abstract}

We present an approach for multi-robot consistent distributed localization and semantic mapping in an unknown environment, considering scenarios with classification ambiguity, where objects' visual appearance generally varies with viewpoint. Our approach addresses such a setting by maintaining a distributed posterior hybrid belief over continuous localization and discrete classification variables. In particular, we utilize a viewpoint-dependent classifier model to leverage the coupling between semantics and geometry. Moreover, our approach yields a consistent estimation of both continuous and discrete variables, with the latter being addressed for the first time, to the best of our knowledge. We evaluate the performance of our approach in a multi-robot semantic SLAM simulation and in a real-world experiment, demonstrating an increase in both classification and localization accuracy compared to maintaining a hybrid belief using local information only.

 
	
\end{abstract}


	
	
	\section{Introduction}
	\label{sec:introduction}


Deployment of multi-robot systems allow for fast information gathering, and can be used in a wide variety of applications, for example: search and rescue, autonomous driving, and agriculture. A significant part of ongoing research is multi-robot Simultaneous Localization and Mapping (SLAM), where a group of robots localize themselves and cooperatively map the environment. Multi-robot SLAM is utilized in a variety of navigation tasks such as cooperative search and rescue, underwater navigation, or warehouse management. SLAM itself is a widely researched problem (see e.g. \cite{Cadena16tro}) in the robotics community. In particular, semantic SLAM reasons about objects within the environment with richer information, such as object's class, compared to geometric SLAM. Yet, often when observed from certain viewpoints, inferring the correct class of an object can be challenging, i.e.~an object may visually appear similar to representative objects from different classes. This induces a viewpoint dependency for classifier outputs and requires information from different viewpoints for maintaining a belief over classification variables. 


In this paper we present the first distributed multi-robot approach for semantic localization and mapping in the above setting. Our approach maintains a hybrid belief over continuous variables (object and camera poses) and discrete variables (object classes), while considering the coupling between classification and localization, and enforcing consistent, double-counting-free estimation.

In contrast, existing approaches for multi-robot semantic SLAM utilize most-likely class measurements to solve data association. Moreover, these approaches do not maintain a belief over classification variables, nor model the coupling between semantic and geometric information.

As each robot uses information from other robots, it must not use measurements more than once, otherwise it will lead to erroneous and overconfident estimates, i.e.~it will double count information. To address this key problem, multiple approaches were proposed, all considering continuous  variables: from complex book-keeping (e.g. \cite{Bahr09icra}) to information removal techniques (e.g. \cite{Cunningham13icra}).
In this work we address consistent inference of a hybrid belief that consists  of continuous and discrete variables. To the best of our knowledge, the latter has not been addressed  thus far.

To summarize, our main contributions are as follows. (i) we contribute a multi-robot approach that maintains a hybrid belief over robot and object poses, and object classes in a distributed setting, while addressing the coupling between semantic and geometric information via viewpoint-dependent classifier model;  (ii) we address estimation consistency aspects considering both continuous and discrete random variables; (iii) we demonstrate the strength of this approach in simulation and real-world experiment, comparing to single robot and distributed multi-robot with double counting. This paper is accompanied with supplementary material  which provides further details and results.

\section{Related Work}

Various works have utilized sequential classification with a classifier model for a single robot.
Omidshafiei et al. \cite{Omidshafiei16arxiv} presented a sequential classification approach that used a Dirichlet distributed classifier model. 
The classifier model was not modeled as viewpoint-dependent. Kopitkov and Indelman \cite{Kopitkov18iros} presented an approach to train a viewpoint dependent classifier model.
Feldman and Indelman \cite{Feldman18icra} proposed a sequential object classification that utilizes a viewpoint dependent classifier with known relative poses a-priori. Tchuiev et al. \cite{Tchuiev19iros} maintained a hybrid belief with a viewpoint dependent classifier to disambiguate between data association realizations. 
These works, \cite{Tchuiev19iros}, address only sequential classification and do not consider the coupled problem with SLAM. To our knowledge, our work is the first to address the coupled problem in a distributed setting.


There are different approaches for distributed multi-robot SLAM; 
Walls et al. \cite{Walls15icra} proposed a distributed geometric SLAM approach that communicates factors between robots.  Other approaches for geometric SLAM include Extended Kalman Filter (such as \cite{Roumeliotis02tra}) or Particle Filter based methods (such as \cite{Howard06ijrr}). 
Choudhary et al. \cite{Choudhary17ijrr} presented an approach for distributed semantic SLAM which communicates relative poses between robots and uses object class information for data association.
The geometric approaches do not reason about object classes, while the semantic approaches consider only most likely classification, i.e. do not maintain a belief over class variables. Our semantic approach maintains a belief over object classes and considers the coupling between the continuous and discrete variables.

Consistent estimation is a key issue in a distributed setup, with multiple approaches proposed to address it.
Bahr et al. \cite{Bahr09icra} proposed a distributed algorithm for under-water vehicles, with an approach for using all measurements without information loss. Indelman et al. \cite{Indelman12ijrr} proposed a graph based method that calculated cross-covariance terms that represent the correlation between measurements from different robots, utilizing it for consistent estimation.
Cunningham et al. \cite{Cunningham10iros} presented the DDF-SAM distributed SLAM algorithm that avoided double counting by creating two maps for each robot: local and global. The global map is updated with condensed local maps. A later work by Cunningham et al. \cite{Cunningham13icra} introduced DDF-SAM2, where each robot maintains only the global map. To avoid double counting, the old information during communication is filtered out via down-dating by each robot. These approaches consider continuous random variables. In contrast, we reason about discrete variables as well.

	
	\section{Notations and Problem Formulation}
	\label{sec:preliminaries}

Consider a group of robots operating in an unknown environment represented by object landmarks. All of the robots aim to localize themselves, and map the environment geometrically and semantically within a distributed multi-robot framework. In this work we consider a closed-set setting, where each of the objects is of one of $M$ possible classes. The number of objects in the environment prior to the scenario is unknown.

We denote states  inferred by robot $r$ with a superscript $\square^r$. Set $R$ is the set of all robots communicating with robot $r$ (including itself), either directly, or relayed through other robots. Note that $R$ can increase its size with time. Let $x_k$ denote robot pose at time $k$, $x^o_n$ and $c_n$ denote the $n$'th object pose and class respectively. Let $\poses^o \doteq \{x^o_n\}_n$ and $\classes \doteq \{c_n\}_n$ denote poses and classes of objects, and $\poses_k \doteq \{x_{0:k},\poses^o_k\}$ denotes all poses up to time $k$. Subscript $new,k$ representing the objects newly observed at $k$.


\begin{table}\scriptsize{
	\caption{Main notations used in the paper.\label{table:Notation}}
	\begin{tabularx}{\textwidth}{p{0.12\textwidth}X}
		\toprule
		\textbf{Parameters} \\
		$x$   & Robot pose \\
		$x^o_n, c_n$ & n'th object pose and class \\
		$\poses^o_k$ & Poses of objects observed up to time $k$ \\
		$\poses^o_{new,k}$ & Poses of objects newly observed at time $k$ \\
		$\poses_k$ & Robot and object poses up to time $k$ \\
		$\classes_k$ & Object seen up to time $k$ class realization \\
		$\classes_{new,k}$ & Classes of objects newly observed at time $k$ \\
		$\mathcal{Z}_k$ & Measurements at time $k$ including geometric and semantic \\
		$\motionmodel_k$ & Motion model from $x_{k-1}$ to $x_k$ \\
		$\likelihoods_k$ & Measurement likelihood of $\mathcal{Z}_k$ \\
		$\his_k$ & History of measurements and action up to time $k$ \\
		$b_k$ & Conditional continuous belief at time $k$ \\
		$w_k$ & Discrete weight at time $k$ \\
		$\xi_k$ & Continuous object marginal belief at time $k$ \\
		$\phi_k$ & Discrete marginal belief at time $k$ \\
		$N_k(\cdot)$ & Number of objects observed by a robot or a group up to time $k$ \\
		\textbf{Superscripts} \\
		$r$ & States of robot $r$ \\
		$R$ & States of robots communicating with $r$, directly and indirectly, including itself \\
		\bottomrule
	\end{tabularx}}
\end{table}

Let $\mathcal{Z}^r_k$ be the set of measurements robot $r$ receives at time $k$ by its own sensors. $\mathcal{Z}^r_k$ is composed of geometric and semantic measurements $\mathcal{Z}^{geo,r}_k$, and $\mathcal{Z}^{sem,r}_k$ respectively. We assume independence between geometric and semantic measurements, as well as between different time steps. 

We assume Gaussian and known identical motion  $\motionmodel_k \doteq \mathbb{P}(x_k|x_{k-1},a_{k-1})$ and geometric  $\mathbb{P}(z^{geo,r}_k|x^r_k,x^{o,r})$ models for all robots. At each time step, there is a subset of object poses involved in the geometric and classifier model that is determined by data association (DA). Unlike our previous work \cite{Tchuiev19iros}, herein, DA is assumed to be externally determined.


Additionally, we use a viewpoint-dependent classifier model that "predicts" classification scores (a vector of class probabilities). This model couples classifier scores with viewpoint dependency between object and camera; this coupling can be used to improve pose inference performance \cite{Tchuiev19iros}. The viewpoint dependency is modeled as a Gaussian with parameters that depend on the relative viewpoint from the camera to the object $x^{o,r} \ominus x^r_k$ and object's class $c$: 
\begin{equation}\label{eq:Semantic_Model}
	\mathbb{P}(z^{sem,r}_k|x^r_k,x^o,c) \!=\! \mathcal{N}(h_c(x^r_k, x^{o,r}),\Sigma_c(x^r_k, x^{o,r})),
\end{equation}
where $h_c(\cdot)$ and $\Sigma_c(\cdot)$ can be learned offline via a Gaussian Process (GP) \cite{Feldman18icra} or a deep neural network \cite{Kopitkov18iros}. Note that for $M$ candidate classes,  $M$ viewpoint-dependent models have to be learned. 

Let $\likelihoods^r_k \doteq \prob{\observations^r_k|\poses^r_k,\classes^r_k}$ be the local measurement likelihood of $r$ that consists of  geometric and classifier models:
%
\begin{equation}\label{eq:Measurement_Likelihood}
\likelihoods^r_k \doteq   \prod_{x^{o,r},c^r}
  \prob{\observations^{geo,r}_k|x^r_k,x^{o,r}}
\prob{\observations^{sem,r}_k|x^r_k,x^{o,r},c^r},
\end{equation}
where $x^{o,r} \in \poses^{o,r}_{\beta_k}$ and $c^r \in \classes^r_{\beta_k}$; the term $\beta_k$ represents the local DA of robot $r$ at time $k$, i.e. the correspondences between observations and object IDs. Denote $\poses^{o,r}_{\beta_k}$ the set of all poses of objects that observed by $r$ at time $k$, and similarly denote $\classes^r_{\beta_k}$ for object classes. For the reader's convenience, Table~\ref{table:Notation} presents the important notations used in the paper, some will be defined in the next section.

\paragraph*{Problem formulation}
For each robot $r$ we aim to maintain the following hybrid belief:
\begin{equation}\label{eq:Hybrid_Base}
	\mathbb{P}(\mathcal{X}^R_k,C^R|\his^R_k),	
\end{equation}
where $\his^R_k \doteq \{ \mathcal{Z}^{r'}_{1:k} , a^{r'}_{0:k-1} \}_{r' \in R}$ is the history of measurements of robot $r$ itself and transmitted information to $r$, as well as actions from all robots in $R$. The belief in Eq.~\eqref{eq:Hybrid_Base} is a hybrid belief  over both continuous (camera and object poses), and discrete (object classes) random variables. We aim to update this hybrid belief per each robot in a recursive manner, using both local measurements and information sent from other robot in the neighborhood, as well as sending information by itself. We aim to keep estimation consistency by avoiding double counting, i.e. using every measurement only once.
	
	
	\section{Approach}
	\label{sec:approach}

We present a framework for distributed classification, localization, and mapping. As with many multi-robot distributed frameworks, over-confident estimations, due to double counting, is a key issue; We propose a framework that simplifies the book-keeping that allows relaying of information (e.g. robot 1 sends information to robot 2, then 2 sends to 3 information that also includes the received from robot 1). This framework requires the maintenance of a local belief $\mathbb{P}(\mathcal{X}^r_k,C^r|\his^r_k)$ per each robot that can be sent and relayed to other robots. From multiple local beliefs a distributed belief can be constructed. The local beliefs are stored by each robot, and updated accordingly when new information arrives, and the receiving robot filters out the old information, thus avoiding double counting.

In the next sections we derive a recursive formulation for maintenance of the local belief, the distributed hybrid belief, and the information stack each robot holds and transmits.


\subsection{Local Hybrid Belief Maintenance}
\label{sec:Ind_Belief_Inference}

Our formulation for maintaining local hybrid beliefs builds upon our previous work \cite{Tchuiev19iros}, with the main differences being that here we assume the DA is solved, and the number of objects is unknown a-priori. In this section we present an overview of this approach. 

We maintain the hybrid belief of robot $r$ only from local information. This belief can be split into continuous and discrete parts as in:
%
\begin{equation}\label{eq:Initial_Expansion}
	\prob{\poses^r_k,C^r_k|\his^r_k} =
	\underbrace{\prob{\poses^r_k|C^r_k, \his^r_k}}_{b^r_k}
	\underbrace{\prob{C^r_k|\his^r_k}}_{w^r_k}.
\end{equation}
To maintain this hybrid belief, we must maintain a set of continuous beliefs conditioned on the class realization of all objects observed in the scene by robot $r$ thus far. 

The continuous part can be updated as follows:
\begin{equation}
\begin{array}{c}\label{eq:Cont_Belief_Main}
	b^r_k  \propto
	b^r_{k-1} \cdot 
	\likelihoods^r_k \cdot
	\motionmodel^r_k \cdot
	\prob{\poses^{o,r}_{\text{new},k}},
\end{array}
\end{equation}
where $\prob{\poses^{o,r}_{\text{new},k}} = \frac{\prob{\poses^{o,r}_k}}{\prob{\poses^{o,r}_{k-1}}}$ is the prior over object poses newly observed at time $k$. As opposed to \cite{Tchuiev19iros}, this formulation also supports an increasing number of objects known at each time step, with both $\poses^{o,r}_k$ and $C^r_k$ increasing in dimension. Note that in general $b^r_k$ is different for each class realization, as models \eqref{eq:Semantic_Model} are different for each class.

The discrete part is the weight associated to its corresponding continuous belief. As our measurement models depend on continuous variables, we  use  Bayes rule on $\prob{\classes^r_k|\his^r_k}$ and marginalize the measurement likelihood as follows:
%
\begin{equation}\label{eq:Discrete_Int}
w^r_k \propto w^r_{k-1}
\prob{\classes^r_{\text{new},k}} 
\int_{\mathcal{X}^r_k} \! \!
\likelihoods^r_k \cdot 
b^r_{k-1} \cdot
\motionmodel^r_k d\mathcal{X}^r_k,
\end{equation}
where $\prob{\classes^r_{\text{new},k}} = \frac{\prob{\classes^r_k}}{\prob{\classes^r_{k-1}}}$ is the prior over classes of new objects locally observed by $r$ at time $k$. 
We compute the integral in Eq.~\eqref{eq:Discrete_Int} by sampling the continuous variables that participate in $\mathbb{P}(\mathcal{Z}^r_k|\mathcal{X}^r_k,C^r_k)$, i.e. the last robot pose $x^r_k$ and the poses of  observed objects $\mathcal{X}^{o,r}_{\beta_k}$ at time $k$. These variables are sampled from the propagated belief $b^r_{k-1} \cdot \motionmodel^r_k$.
Variables that do not participate in $\likelihoods^r_k$ can be marginalized analytically.

 
\subsection{Distributed Hybrid Belief Maintenance}\label{sec:Joint_Hybrid}

In this section we extend the formulation presented in Sec.~\ref{sec:Ind_Belief_Inference} to include updates from other robots, considering a distributed multi-robot setting. As will be seen, our formulation uses each measurement only once, thus keeping estimation consistency and avoiding double counting. Similarly to \eqref{eq:Initial_Expansion}, we factorize the distributed hybrid belief \eqref{eq:Hybrid_Base} 
%
\begin{equation}\label{eq:MR-case-Factorized}
	\prob{\poses^R_k,C^R_k|\his^R_k} = 
	\underbrace{\prob{\poses^R_k|C^R_k,\his^R_k}}_{b^R_k}
	\underbrace{\prob{C^R_k|\his^R_k}}_{w^R_k}.
\end{equation}
As in the single robot case, maintaining this belief requires managing multiple hypotheses of class realizations. Compared to the single robot case, the number of objects observed will be equal or greater for distributed belief, therefore the number of possible realizations increases as well. Importantly, information transmitted by other robots impacts both $b^R_k$ and $w^R_k$. Furthermore, the classifier viewpoint-dependent model induces coupling between localization uncertainty and classification of different robots.

We present a recursive formulation for maintaining each of the parts in \eqref{eq:MR-case-Factorized}. The distributed measurement history $\his^R_k$ can be split to a prior part, and a new part, defined as $\hisnew^R_k$, that consists of measurements and actions from time $k$, s.t:
$\his^R_k = \his^R_{k-1} \cup \hisnew^R_k$. Similarly, let $\his^r_k \doteq \his^r_{k-1} \cup \{ \observations^r_k , a^r_{k-1} \}$ for the single robot case. Note information in $\hisnew^R_k$ transmitted by other robots can potentially be from earlier time instances (as each robot during communication transmits to robot $r$ its own stack of local beliefs of other robots, see Section \ref{sec:Stack}).
Crucially, each measurement must be used once to avoid double counting. We also denote history \emph{without} local measurements and action at time $k$ as
%
\begin{equation}\label{eq:Def_HRm}
\his^{R-}_k \! \! \doteq \! \his^{R}_k \backslash \{ \observations^r_k , a^r_{k-1} \!\}  \!\ ,  \!\ \hisnew^{R-}_k \!\! \doteq \! \hisnew^{R}_k \backslash \{ \observations^r_k , a^r_{k-1} \!\}.
\end{equation}
Using the above notations, one can observe $\his^{R-}_k = \his^R_{k-1} \cup \hisnew^{R-}_k$.
%
%
Next, we detail our approach for maintaining both the conditional continuous part $b^R_k$ and the discrete part $w^R_k$ recursively for a realization of object classes $\classes^R_k$.


\subsubsection{Maintaining $b^R_k$}

Using Bayes rule, we rewrite $b^R_k$ as:
\begin{equation}\label{eq:Joint_Cont_Init}
	b^R_k = \eta \cdot
	\likelihoods^r_k \cdot
	b^{R-}_k
\end{equation}
where $\eta \doteq \prob{\observations^r_k|\classes^r_k,\his^R_k \backslash \observations^r_k}^{-1}$ is a normalization constant the does not participate in inference of the continuous belief. The local measurement likelihood, $\likelihoods^r_k$, is defined in Eq.~\eqref{eq:Measurement_Likelihood}.  

The term $b^{R-}_k \doteq \prob{\poses^R_k|\classes^R_k,\his^R_k \backslash \observations^r_k}$ is the distributed propagated belief that is conditioned on information transmitted by other robots at time $k$, and on the latest action of robot $r$ but not on its local measurement. 
%
%
During update, $b^{R-}_k$ is saved to be used in maintenance of $w^R_k$, as seen in the next subsection.  Using  chain rule, we can extract the motion model of the latest action as well:
\begin{equation}\label{eq:Joint_Cont_Dev_1}
\begin{array}{c}
	b^{R-}_k =
	\motionmodel^r_k \cdot
	\prob{\poses^R_k \backslash x^r_k|\classes^R_k,\his^{R-}_k}.
\end{array}	
\end{equation}
%
%
We can express $\prob{\poses^R_k \backslash x^r_k|\classes^R_k,\his^{R-}_k}$ in  terms of the distributed continuous prior $b^R_{k-1} \doteq \prob{\poses^R_{k-1}|C^R_{k-1},\his^R_{k-1}}$, and the new information received from other robots (see supplementary material Sec.~2): 
\vspace{-0.4cm}
\begin{equation}\label{eq:Joint_Cont_Dev_4}
	\prob{\poses^R_k \backslash x^r_k|\classes^R_k,\his^{R-}_k} = 
	b^R_{k-1} \cdot
	\frac{\prob{\poses^{o,R}_k|\classes^{o,R}_k,\hisnew^{R-}_k}}
	{\prob{\poses^{o,R}_{k-1}}}
\end{equation} 
Finally, we substitute Eq.~\eqref{eq:Joint_Cont_Dev_4} to Eq.~\eqref{eq:Joint_Cont_Dev_1} and in turn to Eq.~\eqref{eq:Joint_Cont_Init}, and get the following recursive formulation:
\begin{equation}\label{eq:Joint_Cont_Final}
	b^R_k \propto 
	b^R_{k-1} \cdot
	\likelihoods^r_k \cdot
	\motionmodel^r_k \cdot
	\prob{\poses^{o,R}_{\text{new},k}}
	{\color{blue} \frac{\prob{\poses^{o,R}_k|\classes^{o,R}_k,\hisnew^{R-}_k}}{\prob{\poses^{o,R}_{k}}}},
\end{equation}
where the measurement likelihood $\likelihoods^r_k$ accounts for the new local measurement, $\motionmodel^r_k$ accounts for the latest action of robot $r$, and $\prob{\poses^{o,R}_k|\classes^{o,R}_k,\hisnew^{R-}_k}$ (shown in {\color{blue}blue}) accounts for new information sent to $r$ by other robots in $R$ at time $k$. This pdf is only over object poses ($\poses^{o,R}_k$), while the other robots' poses are marginalized out. Thus, robots communicate the environment states, which are implicitly affected by the robots' pose estimation.  Computation of the {\color{blue}blue} part is further discussed in Sec.~\ref{sec:Stack}. Compared to the local belief update \eqref{eq:Cont_Belief_Main}, the blue part is the main difference. The expression $\prob{\poses^{o,R}_{\text{new},k}}$ represents pose prior of objects newly known by $r$ at time $k$.

The distributed belief has at worst $M^{N_k(R)}$ continuous beliefs with corresponding weights, where the number of objects $N_k(R)$ known by $r$ can increase with time. Naturally, a multi-robot system will observe more objects than a single robot, therefore the computational burden for distributed belief will be larger than for the local belief. Therefore, the significance of pruning beliefs with small weight grows. We set a threshold for the ratio between a weight and the largest weight in the distributed hybrid belief.


\subsubsection{Maintaining $w^R_k$}

To maintain $w^R_k$, we use a similar derivation to the weight update via local information only, presented in Sec.~\ref{sec:Ind_Belief_Inference}. We use  Bayes rule to extract the last local measurement likelihood:
\begin{equation}\label{eq:Joint_Disc_Initial}
	w^R_k = \eta \cdot w^{R-}_k \cdot
	\prob{\observations^r_k|\classes^R_k,\his^R_k \backslash \observations^r_k},
\end{equation}
where $w^{R-}_k \doteq \prob{\classes^R_k|\his^R_k \backslash \observations^r_k}$ is the posterior distributed weight without accounting for the latest local measurements, 
and $\eta \doteq \prob{\observations^r_k|\his^R_k \backslash \observations^r_k}^{-1}$ is a normalization constant that is identical in all realizations of $C^R_k$, thus does not participate in weight inference. As we use a viewpoint dependent classifier model that utilizes the coupling between relative viewpoint and object class, we need to marginalize $\prob{\observations^r_k|\classes^R_k,\his^R_k \backslash \observations^r_k}$ over the involved poses in this likelihood: the last robot pose $x^r_k$, and poses of objects observed at time $k$. We denote the latter by $\poses^{o,r}_{\beta_k}$, and to shorten notations  denote $\poses^r_{\text{inv},k} \doteq \{x^r_k,\poses^{r,k}_{\beta_k}\}$, and by $\lnot \poses^r_{\text{inv},k}$. Thus, $\prob{\observations^r_k|\classes^R_k,\his^R_k \backslash \observations^r_k}$ is marginalized as
\begin{equation}\label{eq:Joint_Disc_Part_1}
\prob{\observations^r_k|\classes^R_k,\his^R_k \backslash \observations^r_k} = 
\int_{\poses^r_{\text{inv},k}}
\! \! \! \! \likelihoods^r_k \cdot
\prob{\poses^r_{\text{inv},k}|\classes^r_k,\his^R_k \backslash \observations^r_k} 
d\poses^r_{\text{inv},k},
\end{equation}
where $\prob{\poses^r_{\text{inv},k}|\classes^r_k,\his^R_k \backslash \observations^r_k}$ is computed by marginalizing $b^{R-}_k$ over the uninvolved variables $\lnot \poses^r_{\text{inv},k}$, with $\mathcal{X}^R_k = \poses^r_{\text{inv},k} \cup \lnot \poses^r_{\text{inv},k}$, as
%
\begin{equation}\label{eq:Joint_Disc_Preblue}
	\prob{\poses^r_{\text{inv},k}|\classes^r_k,\his^R_k \backslash \observations^r_k} =
	\int_{\lnot \poses^r_{\text{inv},k} }
	b^{R-}_k
	d\lnot \poses^r_{\text{inv},k}.
\end{equation}
The propagated distributed belief $b^{R-}_k$ is given to us from the continuous belief with Eq.~\eqref{eq:Joint_Cont_Dev_1}, and includes the external information, shown in {\color{blue}blue}.

In practice, we sample the involved variables $\poses^r_{\text{inv},k}$ in the current measurement likelihood and compute its value. As $b^R_k$ and $\likelihoods^r_k$ are Gaussian, $\eta$ does not play a role in the sampling process. Despite the classifier outputs being modeled as Gaussian, we integrate over poses; In the general case, expectation and covariance of the classifier model are a function of the relative viewpoint, thus we need to sample $\poses^r_{\text{inv},k}$ as presented in Sec.~\ref{sec:Ind_Belief_Inference} at Eq.~\eqref{eq:Discrete_Int}.

The other term we will address from Eq.~\eqref{eq:Joint_Disc_Initial} is $w^{R-}_k$. 
We express $w^{R-}_k$ in terms of $w^R_{k-1}$:
\begin{equation}\label{eq:Joint_Disc_Part_2}
\begin{array}{c}
w^{R-}_k \propto  w^R_{k-1} \cdot
\prob{C^R_{k-1}}^{-1}  \cdot
\prob{\classes^R_k|\hisnew^{R}_k \backslash \observations^r_k}.
\end{array}
\end{equation}
Finally, we substitute Eq.~\eqref{eq:Joint_Disc_Part_1} and \eqref{eq:Joint_Disc_Part_2} to Eq.~\eqref{eq:Joint_Disc_Initial} to reach our final recursive form for the discrete belief update:
\begin{equation}\label{eq:Joint_Disc_Final}
\begin{array}{c}
	w^R_k \propto w^R_{k-1} \cdot
	\prob{\classes^R_{\text{new},k}}
	{\color{red} \frac{\prob{\classes^R_k|\hisnew^{R}_k \backslash \observations^r_k}}{\prob{C^R_{k}}} }
	\int_{\poses^r_{\text{inv},k}}
	\likelihoods^r_k \cdot \\ \cdot
	\prob{\poses^r_{\text{inv},k}|\classes^r_k,\his^R_k \backslash \observations^r_k}
	d\poses^r_{\text{inv},k},
\end{array}
\end{equation}
with $\prob{\poses^r_{\text{inv},k}|\classes^r_k,\his^R_k \backslash \observations^r_k}$ computed via Eq.~\eqref{eq:Joint_Disc_Preblue}. This is a recursive formulation that includes the discrete prior $w^R_{k-1}$, external updates for the class probability from other robots (shown in {\color{red}red}), and the external updates for the continuous belief contained within the integral.

\emph{Remark}: One might be tempted to infer the class of each object separately, but it is not accurate due to the coupling between relative viewpoint and object class, as each object class is possibly implicitly dependent on all poses: robot and objects (see supplementary material Sec.~3).

\subsection{Communication Between Robots}\label{sec:Stack}

In Sec.~\ref{sec:Joint_Hybrid} we presented a framework to maintain a hybrid belief of $r$ given information obtained from other robots in $R$. That information was represented by the continuous {\color{blue}blue} expression in Eq.~\eqref{eq:Joint_Cont_Final} and implicitly in Eq.~\eqref{eq:Joint_Disc_Final}, and the discrete {\color{red}red} expression in Eq.~\eqref{eq:Joint_Disc_Final}. In this section, we present our approach for computing these parts, thus describing the management of this information and what each robot sends when communicating. We aim to achieve two goals:
\begin{enumerate}
	
	\item Simple double counting prevention when maintaining the distributed belief without complex bookkeeping.
	
	\item Distributed belief inference also via data not directly transmitted (e.g. robot $r_1$ sends data to $r_2$, $r_2$ to $r_3$, and $r_3$ is using data from $r_1$).
	
\end{enumerate}
As will be shown next, the {\color{blue}blue} and {\color{red}red} terms in Eqs.~\eqref{eq:Joint_Cont_Final} and \eqref{eq:Joint_Disc_Final} can be expressed via local information transmitted by different robots in $R$ to robot $r$.  To that end, each robot $r$ maintains and broadcasts a \emph{stack} of local hybrid beliefs of other robots it is aware of. In contrast to \eqref{eq:Initial_Expansion}, these local beliefs are marginal beliefs over object poses and classes, i.e.~robot poses are marginalized out.
Each slot for robot $r'$ in the stack of robot $r$ contains $N_k(r')$ continuous and discrete marginal beliefs (defined below as $\xi^{r,r'}_k$ and $\phi^{r,r'}_k$), one pair per class realization, following a factorization similar to \eqref{eq:Initial_Expansion}. Additionally, each slot includes a time-stamp that indicates on what data the local hybrid belief is conditioned upon. All in all, every stack contains $\sum_{i=1}^{|R|} N_k(r_i)$ continuous and discrete beliefs. Eq.~\eqref{eq:Slot_Full} presents the stack of robot $r$ as a set of slots, where each slot contains a hybrid belief of a particular robot $r_i \in R$ over object poses and classes, normalized by their priors.
\begin{equation}\label{eq:Slot_Full}
	\mathcal{S}_k^r \doteq \left\{   \left( \frac{\prob{\poses^{o,r_i}_{k_i}|\classes^{r_i}_{k_i},\his^{r_i}_{k_i}} \prob{C^{r_i}_{k_i}|\his^{r_i}_{k_i}}}
	{\prob{\poses^{o,r_i}_{k_i}}\prob{\classes^{r_i}_{k_i}}}, k_i \right)  \right\}_{r_i \in R},
\end{equation}
where $k_i$ is the time-stamp when robot $r$ received information about $r_i$. In general, time $k_i$ is not synchronized with $k$. The marginal continuous and discrete beliefs that robot $r$ has about robot $r_i\in R$ are denoted $\xi^{r,r_i}_k \doteq \prob{\poses^{o,r_i}_{k_i}|\classes^{r_i}_{k_i},\his^{r_i}_{k_i}} 
/ \prob{\poses^{o,r_i}_{k_i}}$ for the continuous part, and $\phi^{r,r_i}_k \doteq \prob{\classes^{r_i}_{k_i}|\his^{r_i}_{k_i}} 
/ \prob{\classes^{r_i}_{k_i}}$ for the discrete part.

With these definitions of $\xi^{r,r_i}_k$ and $\phi^{r,r_i}_k$, it is possible to show that  the {\color{blue}blue} part in Eq.~\eqref{eq:Joint_Cont_Final} can be expressed as (see full derivation in supplementary material)
\begin{equation}\label{eq:Stack_Cont_Final}
	{\color{blue}\frac{\prob{\poses^{o,R}_k|\classes^R_k,\hisnew^{R-}_k}}
	{\prob{\poses^{o,R}_{k}}}} = 
	\prod_{r_i \in R}
	\frac{\xi^{r,r_i}_{k}}{\xi^{r,r_i}_{k-1}} 
\end{equation}
Similarly, the {\color{red}red} term in Eq.~\eqref{eq:Joint_Disc_Final} can be expressed as (see full derivation in supplementary material):
\begin{equation}\label{eq:Stack_Disc_Final}
	{\color{red}\frac{\prob{\classes^R_k|\hisnew^{R}_k \backslash \observations^r_z}}
	{\prob{\classes^R_k}}} =
	\prod_{r_i \in R}
	\frac{\phi^{r,r_i}_k}{\phi^{r,r_i}_{k-1}}.
\end{equation}
Eqs.~\eqref{eq:Stack_Cont_Final} and \eqref{eq:Stack_Disc_Final} present the external update as a product of local beliefs, with only the updates from $k-1$ for robot $r$ are present. This formulation avoids double counting by removing old information, $\xi^{r,r_i}_{k-1}$ and $\phi^{r,r_i}_{k-1}$, in each communication and uses measurements only once. Specifically for $\xi^{r,r_i}_{k-1}$, we use the approach presented in \cite{Cunningham13icra}. Doing so by maintaining stacks of individual information does not require complex book-keeping, only time-stamps for each slot; Thus we fulfill the first goal.
Robots can also relay information transmitted to them, thus the distributed belief can be updated by information from robots that did not transmit to the inferring robot, thus fulfilling the second goal.

Robot $r_i$ sends the entire stack during information broadcast. When robot $r$ receives information, it integrates the broadcast in as follows: recall that $r_i$'s stack is divided to slots, with a time stamp per each slot. Robot $r$ compares time stamps with the received information per slot, and replaces the information within the slot if the received time stamps is newer. If $r$ receives information from more than one other robot at the same time, it will select the newest information per slot. This procedure is dependent on the relations between time-stamps, thus it is not necessary to synchronize time between the robots.

In the following section we  discuss double counting aspects of discrete random variables, corresponding to Eq.~\eqref{eq:Stack_Disc_Final}.

\subsection{Double Counting of Discrete Random Variables}\label{sec:disc_DC}

Double counting leads to over-confident estimations, and if an erroneous measurement is counted multiple times, it may lead to a large error in the state's estimation in turn. While the implications of double counting on continuous random variables (e.g. camera poses and objects) have been investigated, it is not so for discrete random variables. Both cases have a common thread: measurements counted multiple times will 'push' the posterior estimation to a certain direction while leading to lower uncertainty than when double counting is appropriately avoided (i.e.~each measurement is used at most once). 
In the continuous Gaussian case, it manifests in a covariance matrix with smaller eigenvalues. Comparatively, in the discrete case the highest probability category will have its probability increase while the probability of not being in this category decreases.

To illustrate the above, consider an example with a categorical random variable $c$; we receive two sets of data $Z_a = \{ z_1 , z_2 \}$, and $Z_b = \{ z_2 , z_3 \}$, with a common measurement $z_2$. Considering a measurement likelihood $\prob{z | c}$, the 
posterior over $c$ is (see e.g.~Bailey et al.~\cite{Bailey12fusion}):
\begin{equation}\label{eq:C_rand_example}
\prob{c|Z_a,Z_b} \! \! \propto \! \prob{c} \prob{Z_a,Z_b|c} \! = \! \prob{c} 
\frac{\prob{z_1|c} \prob{z_2|c}^2 \prob{z_3|c}}
{\prob{z_2|c}}.
\end{equation}
If the common data (measurement $z_2$) is not removed via the denominator in Eq.~\eqref{eq:C_rand_example}, it will be double counted. Compared to Eq.~\eqref{eq:Stack_Disc_Final}, the above nominator and denominator correspond, respectively,  to the terms $\phi^{r,r_i}_k$ and $\phi^{r,r_i}_{k-1}$. 

Denote $\prob{z_2|c=i} \doteq a_i$, and to shorten the notations $\prob{c=i} \prob{z_1|c=i} \prob{z_3|c=i} \doteq \mathcal{L}_i$.  The normalized posterior can be written as:
\begin{equation}
	\prob{c=i|Z_a,Z_b} = 
	\frac{a_i \mathcal{L}_i}
	{\sum_{j=1}^m a_j \mathcal{L}_j} = \frac{a_i^2 \mathcal{L}_i}
	{\sum_{j=1}^m a_j \mathcal{L}_j \cdot a_i }
\end{equation}
where $m$ is the number of candidate categories. Double counting, i.e. without the denominator in Eq.~\eqref{eq:C_rand_example}, gives after normalization $\frac{a_i^2 \mathcal{L}_i}
{\sum_{j=1}^m a_j^2 \mathcal{L}_j  }$.

The largest $a_i$ is denoted $a_{max}$, with $i_{max}$ being the category corresponding to $a_{max}$, and subsequently the product of all other terms for $i_{max}$ is denoted $\mathcal{L}_{max}$. Double counting of $\prob{z_2|c_i}$ will increase the probability of $i_{max}$:
\begin{equation}
	\prob{c=i_{max}|Z_a,Z_b} = 
	\frac{a_{max}^2 \mathcal{L}_{max}}
	{\sum_{j=1}^m a_j \mathcal{L}_j \cdot a_{max}} \leq 
	\frac{a_{max}^2 \mathcal{L}_{max}}
	{\sum_{j=1}^m a_j^2 \mathcal{L}_j}.
\end{equation}
Similarly, it can be shown that with higher power (i.e. counting the data more) can increase the posterior probability even further; In addition, the reverse can be shown for the lowest probability in $a$. This increase in influence can be disastrous if the category of the highest probability likelihood is not correct, possibly leading to pruning of the correct class hypothesis when maintaining the hybrid belief \eqref{eq:Hybrid_Base}.

\begin{figure}
	
	\begin{subfigure}[b]{0.25\textwidth}
		\includegraphics[width=\textwidth]{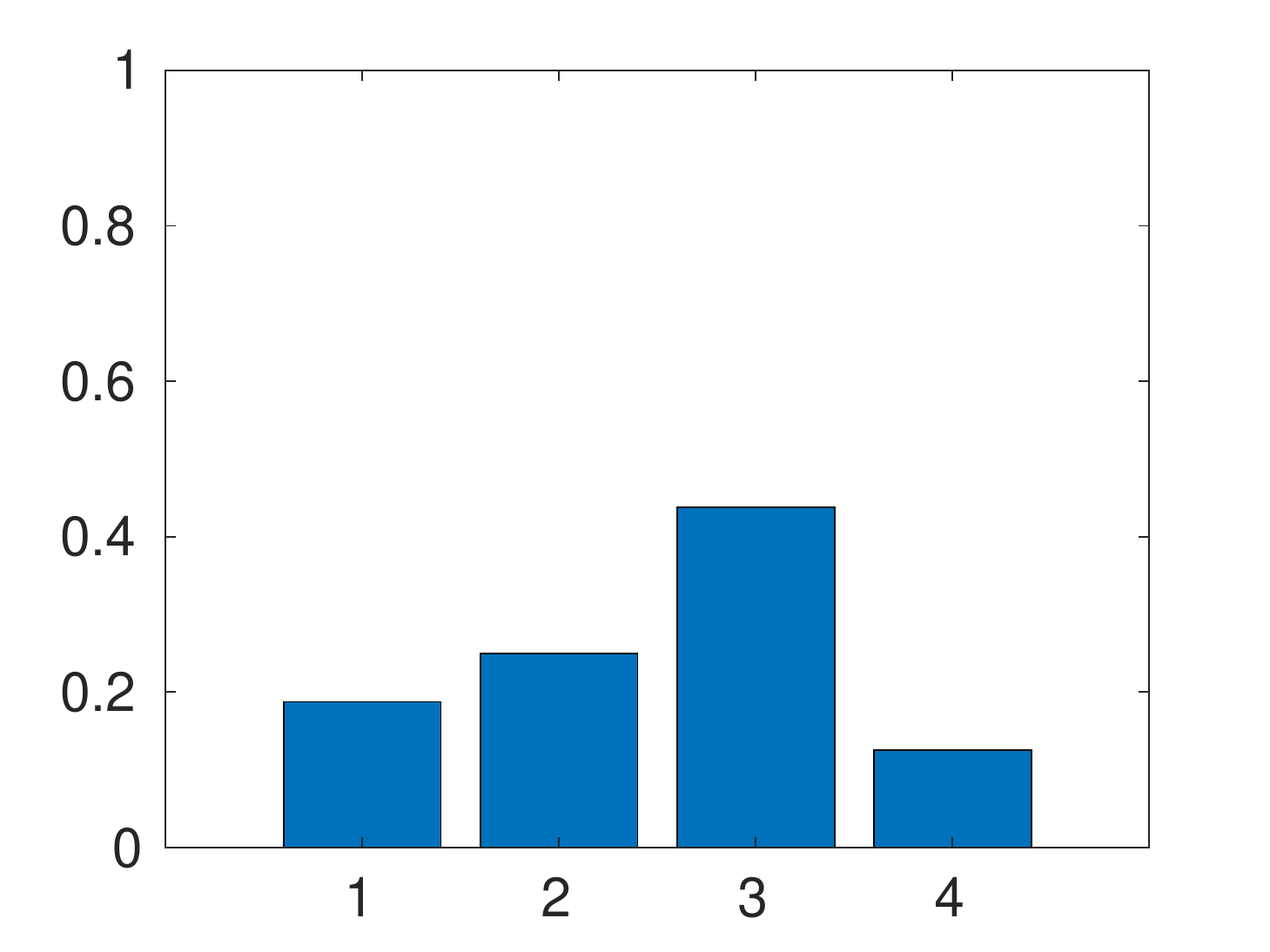}
		\caption{}\label{fig:Bar1}
	\end{subfigure}
	\begin{subfigure}[b]{0.25\textwidth}
		\includegraphics[width=\textwidth]{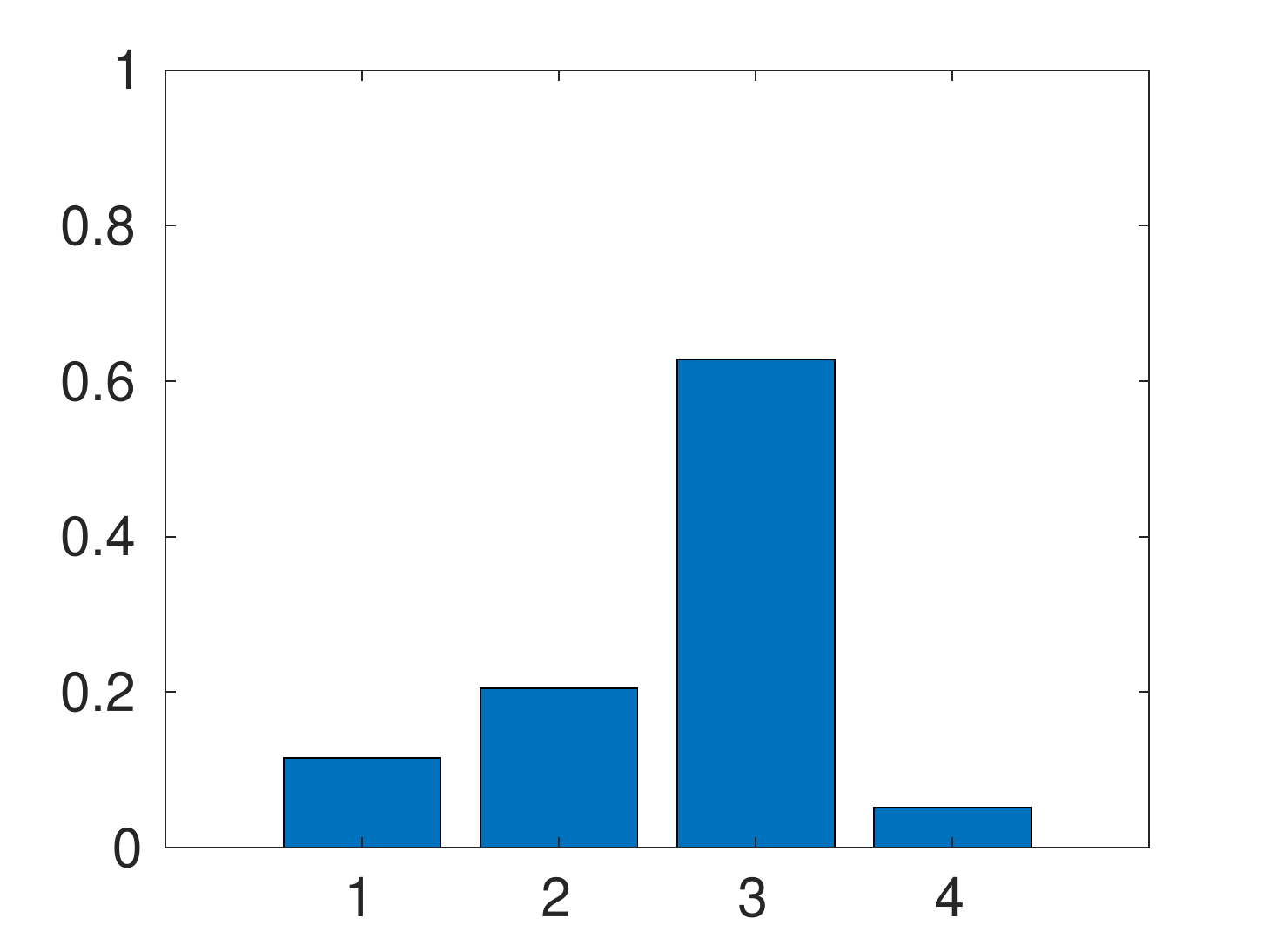}
		\caption{}\label{fig:Bar2}
	\end{subfigure}
	\begin{subfigure}[b]{0.25\textwidth}
		\includegraphics[width=\textwidth]{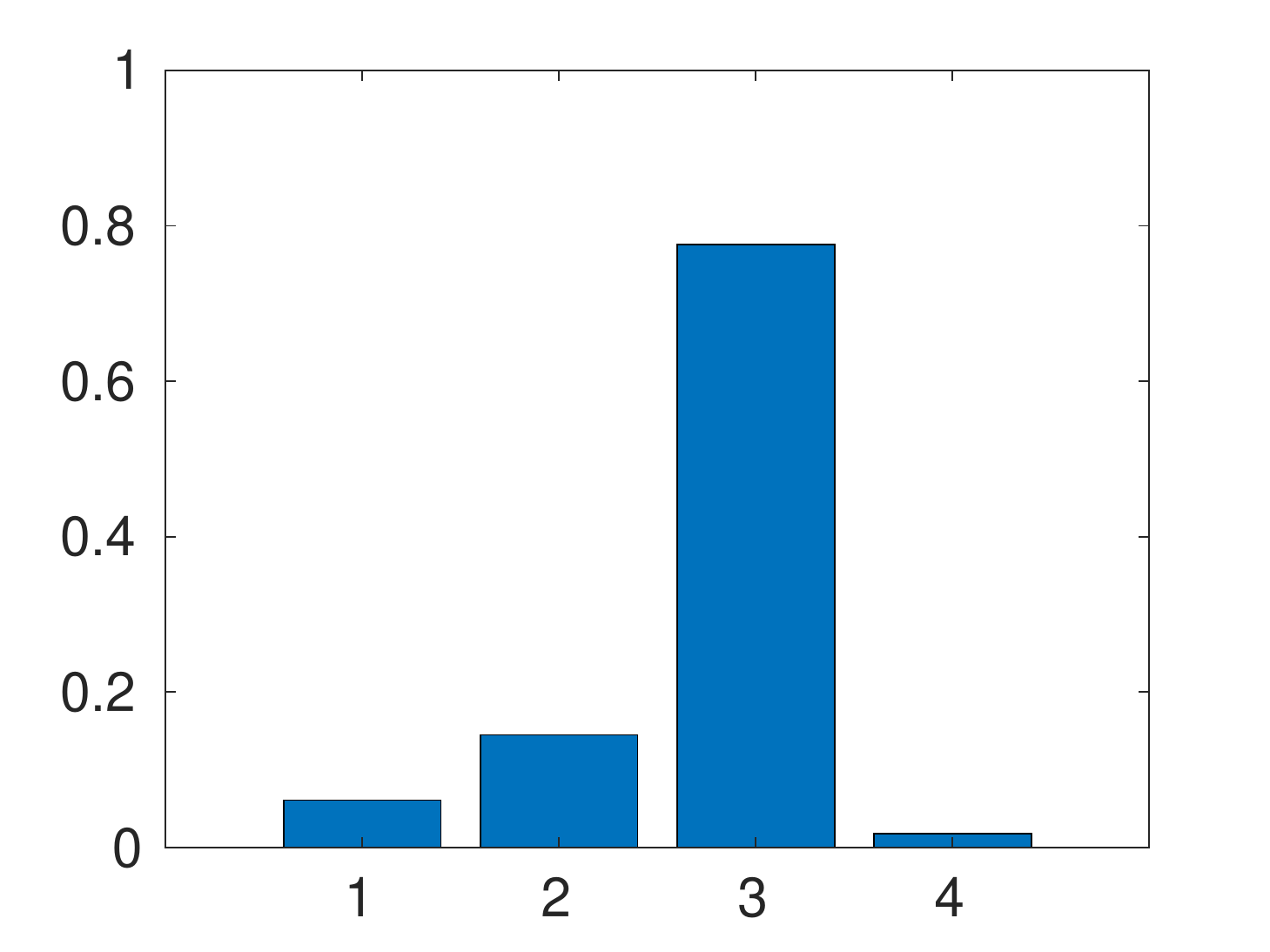}
		\caption{}\label{fig:Bar3}
	\end{subfigure}
	
	\caption{Conceptual demonstration of the effects of double counting on discrete random variables. Consider 4 possible categories with an uninformative prior over them. \textbf{(a)} is the measurement likelihood for the categories. Considering the uninformative prior, it is the posterior distribution as well. \textbf{(b)} and \textbf{(c)} counts the same likelihood twice and thrice respectively.}
	\label{fig:Bars}
\end{figure}

A visualization can be seen in Fig.~\ref{fig:Bars}, where there are 4 categories with uninformed prior and a measurement likelihood; in Figs.~\ref{fig:Bar1}, \ref{fig:Bar2} and \ref{fig:Bar3} the likelihood is counted once, twice and thrice respectively. Evidently, the strongest category's probability (cat. 3) is increased when counted more times while all other have their probability diminish.

	\section{Experiments}
	\label{sec:experiments}
	
\begin{figure*}[!htbp]
	
	\begin{subfigure}[b]{0.18\textwidth}
		\includegraphics[width=\textwidth]{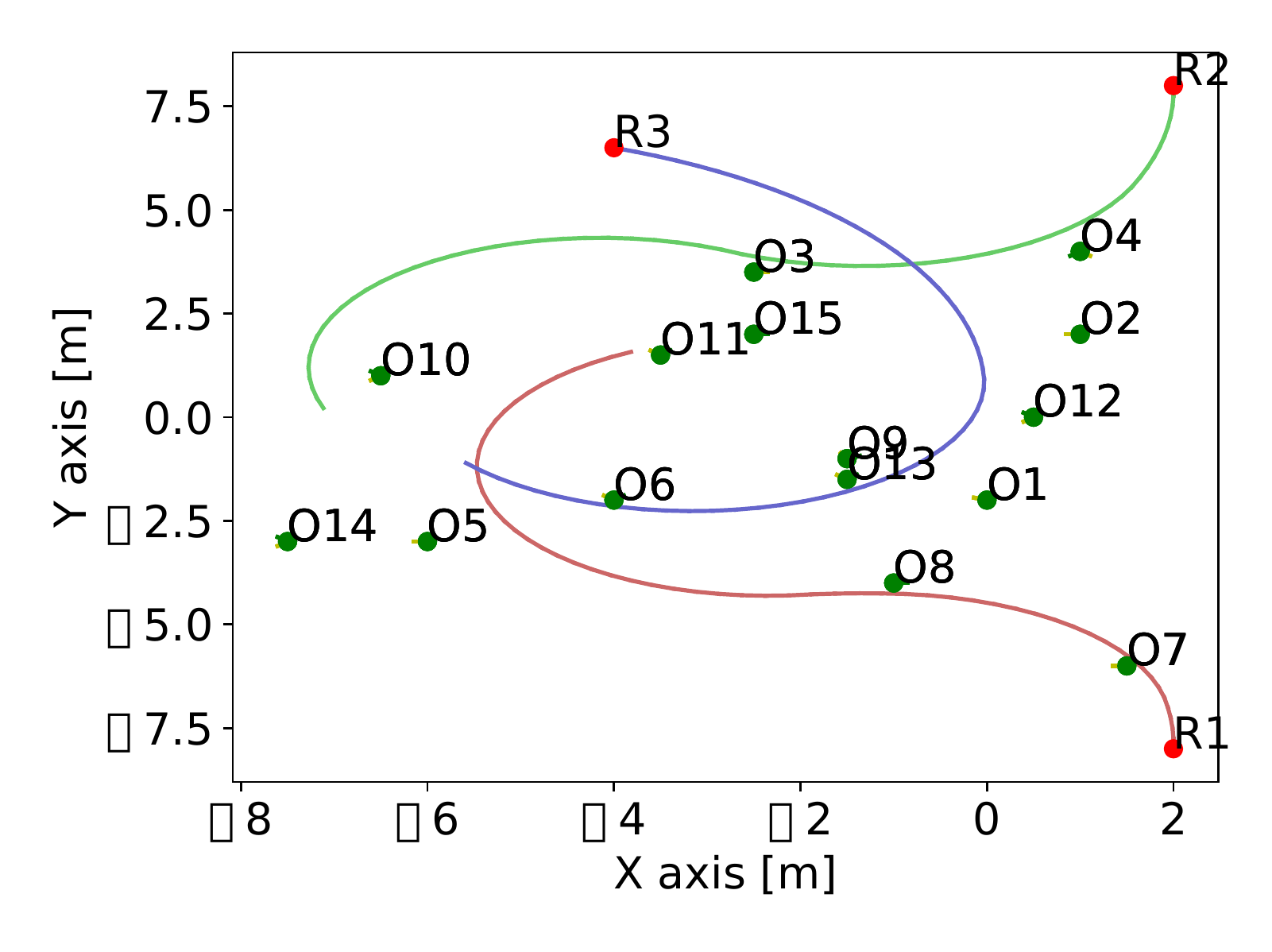}
		\caption{Ground Truth}\label{fig:GT_paths_sim}
	\end{subfigure}
	\begin{subfigure}[b]{0.18\textwidth}
		\includegraphics[width=\textwidth]{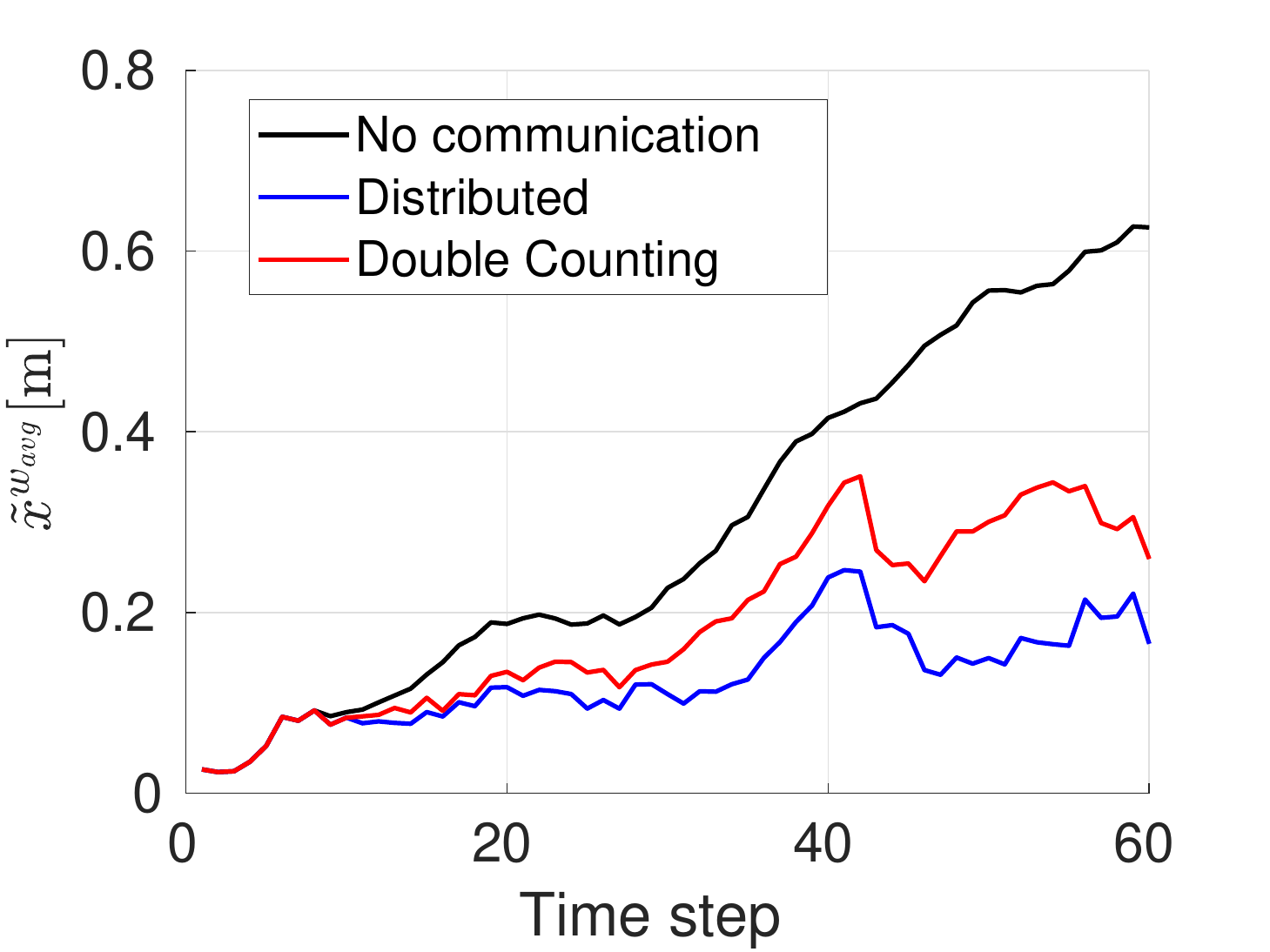}
		\caption{Robot position}\label{fig:x_wmax}
	\end{subfigure}
	\begin{subfigure}[b]{0.18\textwidth}
		\includegraphics[width=\textwidth]{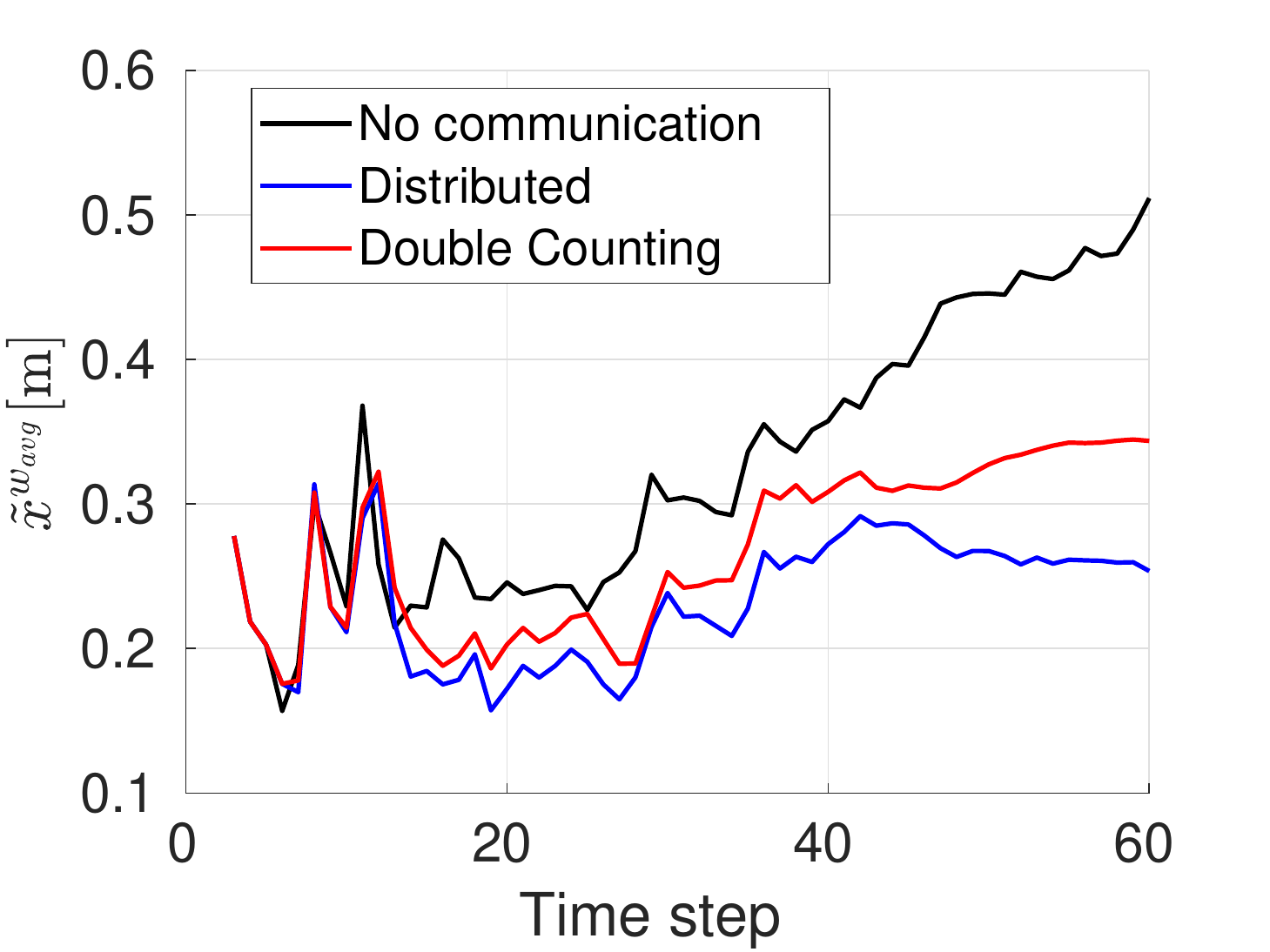}
		\caption{Object position}\label{fig:x_wmax_obj}
	\end{subfigure}
	\begin{subfigure}[b]{0.18\textwidth}
		\includegraphics[width=\textwidth]{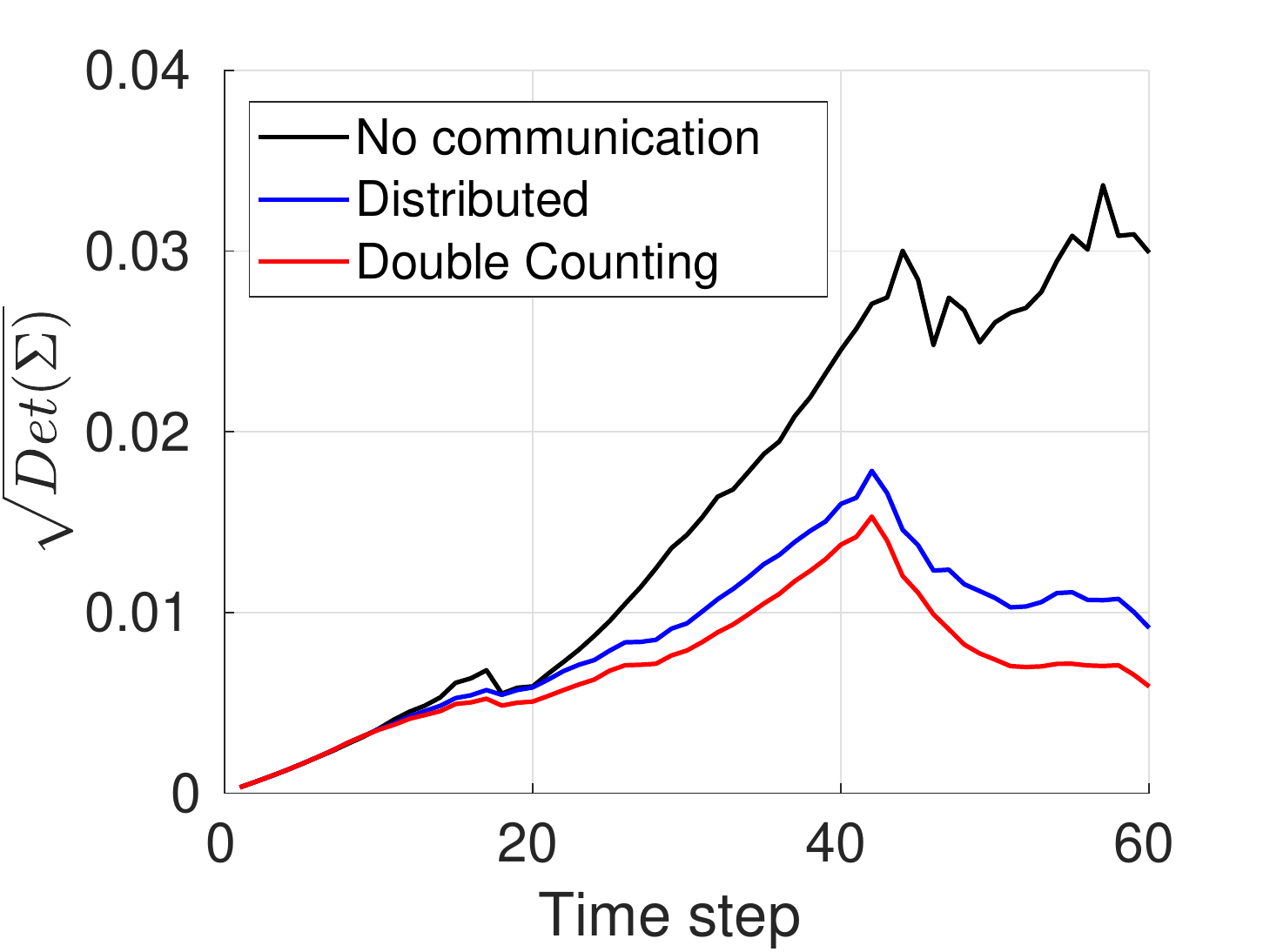}
		\caption{Robot covariance}\label{fig:Cov_avg}
	\end{subfigure}
	\begin{subfigure}[b]{0.18\textwidth}
		\includegraphics[width=\textwidth]{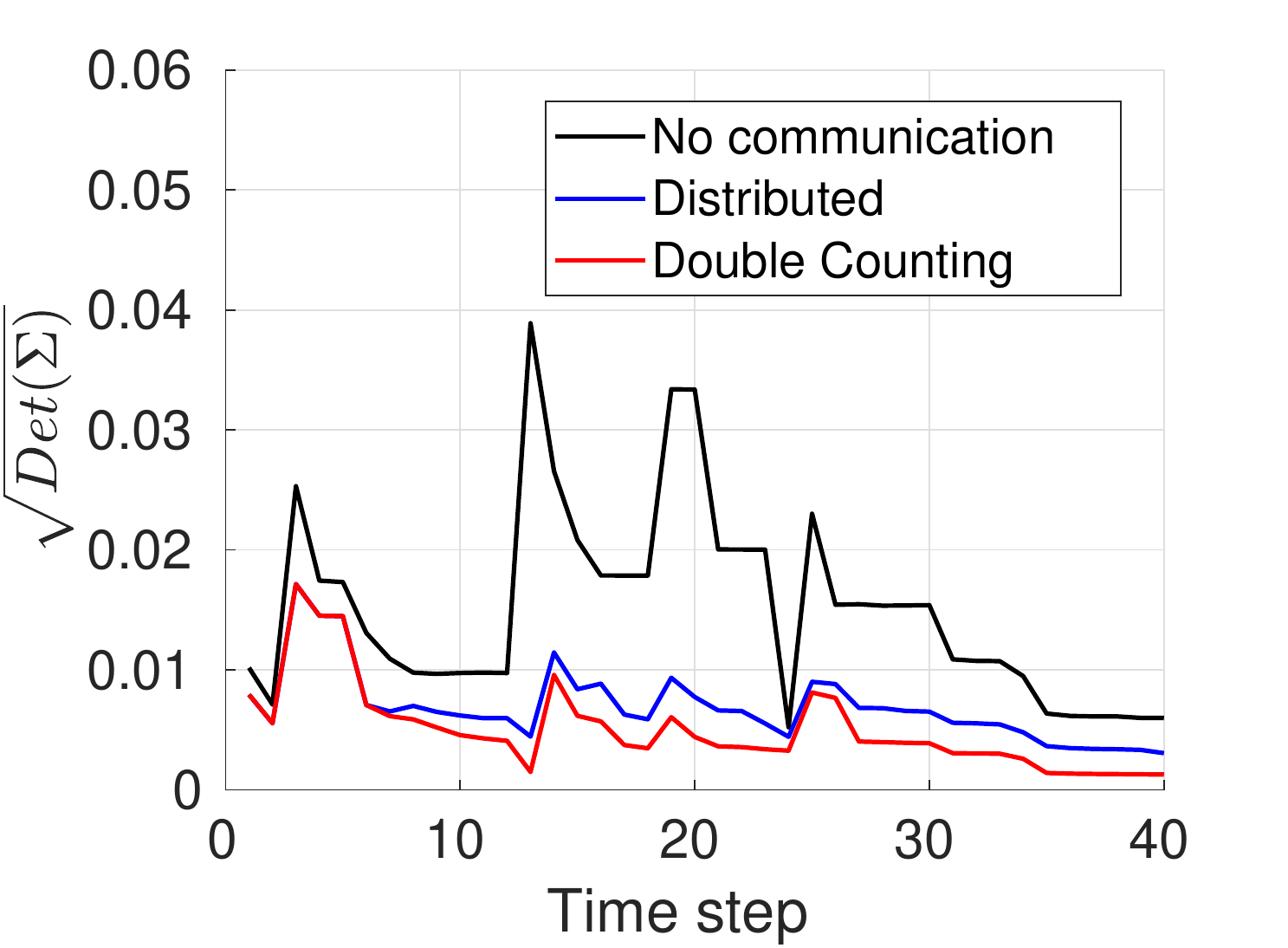}
		\caption{Object covariance}\label{fig:Cov_avg_obj}
	\end{subfigure}
	\caption{Simulation figures; \textbf{(a)} present the ground truth of the scenario. Red points represent the initial position of the robots, with different colored lines represent different robots. The green points represent the object poses. \textbf{(b)} and \textbf{(c)} represent the average $\tilde{x}^{w_{\text{avg}}}$ for robot and object position respectively as a function of time. \textbf{(d)} and \textbf{(e)} present the corresponding square-root of the position covariance for the robot and object average respectively.}
	\label{fig:x_wmax_figures}
\end{figure*}

We evaluated our approach in a multi-robot SLAM simulation and with real-world data where we consider an environment comprising several scattered objects observed by multiple mobile cameras from different viewpoints. Fig.~\ref{fig:GT_paths_sim} and Fig.~\ref{fig:GT_paths_real} present the ground truth for simulation and experiment respectively. Our implementation uses the GTSAM library \cite{Dellaert12tr} with a python wrapper.
The hardware used  is an Intel i7-7700 processor running at 2.8GHz and 16GB RAM, with GeForce GTX 1050Ti with 4GB RAM.

\subsection{Simulation Setting, Compared Approaches and Metrics}

Consider 3 robots, denoted $r_1$, $r_2$, and $r_3$, moving in a 2D environment represented by $N=15$ scattered objects. We consider a closed-set setting and assume, for simplicity, $M=2$ classes, where each object can be one of the two. In this scenario the maximum number of possible class realizations is $M^N=32768$. 

Our approach is evaluated for both classification, and pose inference accuracy, as we maintain a hybrid belief.
We consider an ambiguous scenario where the classifier model cannot distinguish between the two classes from a certain viewpoint, thus requiring additional viewpoints to correctly disambiguate between the two classes. The robots communicate between themselves, increasing performance for discrete and continuous variables, i.e. classification and SLAM. Additionally, the distributed setting extends the sensing horizon, allowing robots to reason about objects that are not directly observed, while keeping estimation consistency.

Each robot only communicates with robots within a 10 meter communication range, relaying the local information stored in its stack. In particular, initially $r_2$ and $r_3$ share information with each other, then $r_1$ and $r_2$, relaying information from $r_3$ through $r_2$. For a complete table of communication in the considered scenario, see supplementary material.  Further, we assume the robots share a common reference frame (this assumption can be relaxed as in \cite{Indelman16csm}). We simulate relative pose odometry and geometric measurements, and we crafted a classifier model that simulates perceptual aliasing.

In the evaluation we compare between three approaches: local estimations, our approach, and our approach with double counting, i.e. $\xi^{r,r_i}_{k-1} = 1$ and $\phi^{r,r_i}_{k-1} = 1$ in Eq.~\eqref{eq:Stack_Cont_Final} and \eqref{eq:Stack_Disc_Final} respectively. In all benchmarks we average the results for each robot. The parameters are presented in the supplementary material Sec. 6.

\begin{figure*}[!htbp]
	
	\begin{subfigure}[b]{0.18\textwidth}
		\includegraphics[width=\textwidth]{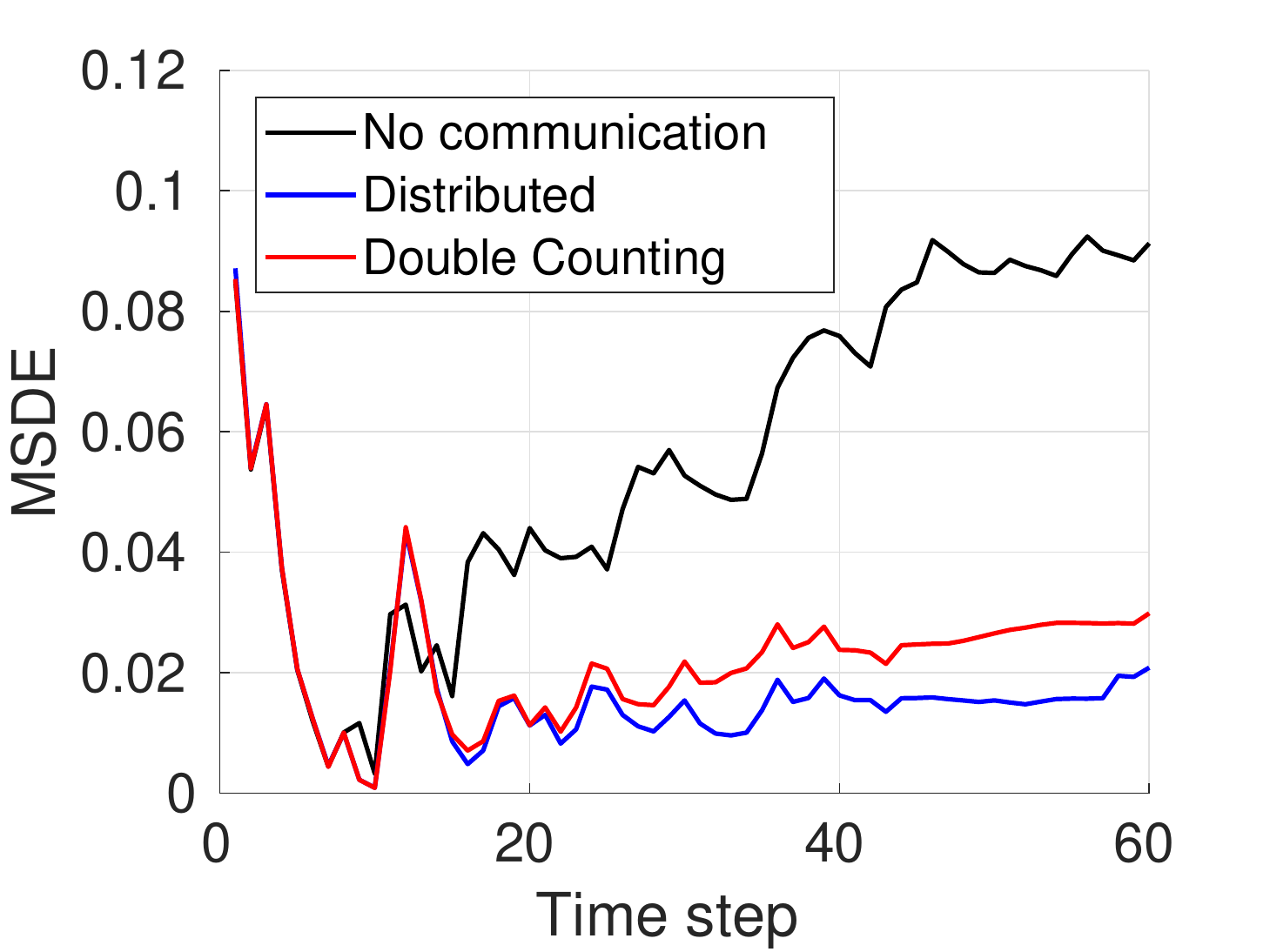}
		\caption{MSDE}\label{fig:MSDE_statistical}
	\end{subfigure}
	\begin{subfigure}[b]{0.18\textwidth}
		\includegraphics[width=\textwidth]{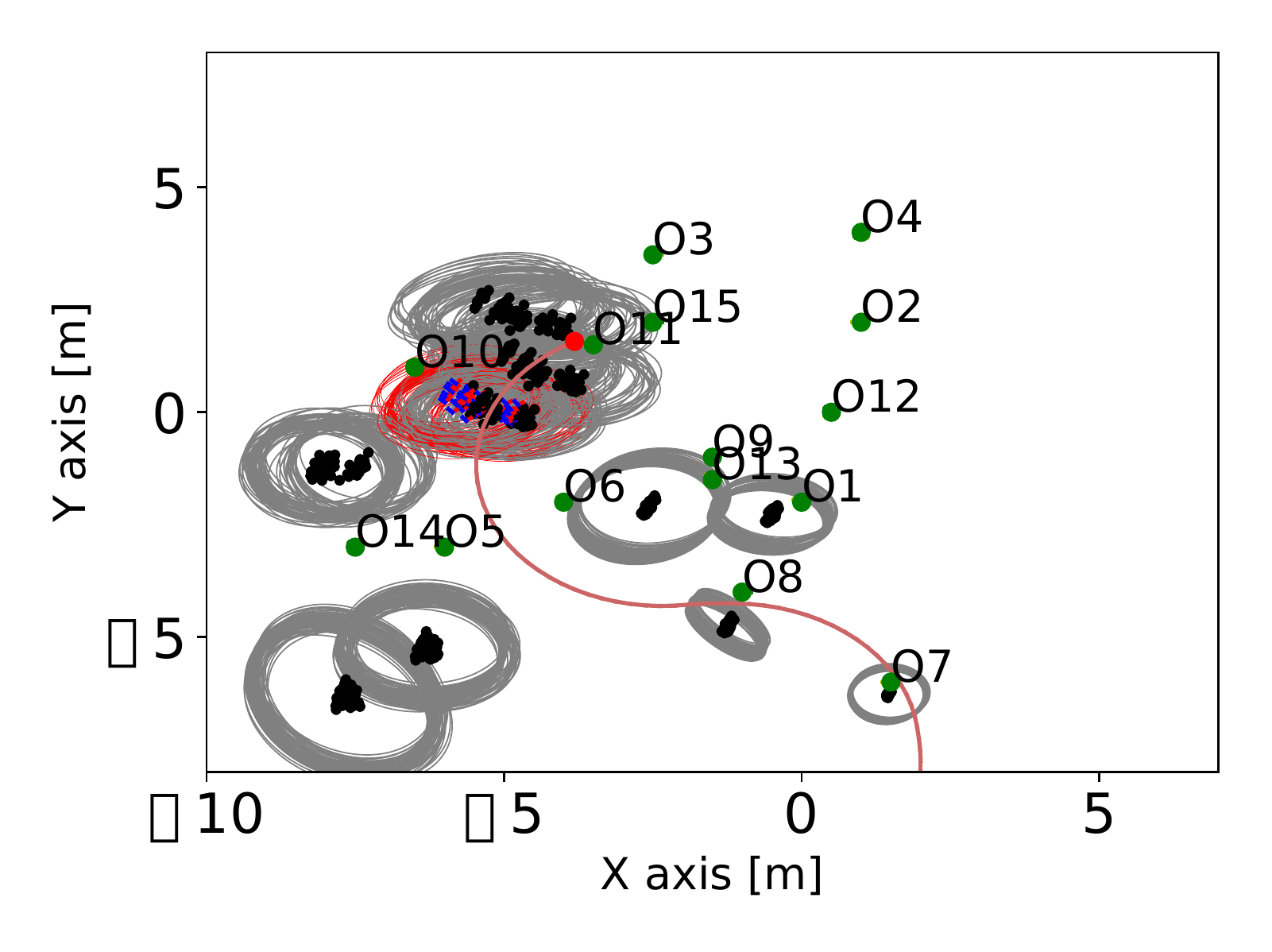}
		\caption{Local}\label{fig:cls_real_1}
	\end{subfigure}
	\begin{subfigure}[b]{0.18\textwidth}
		\includegraphics[width=\textwidth]{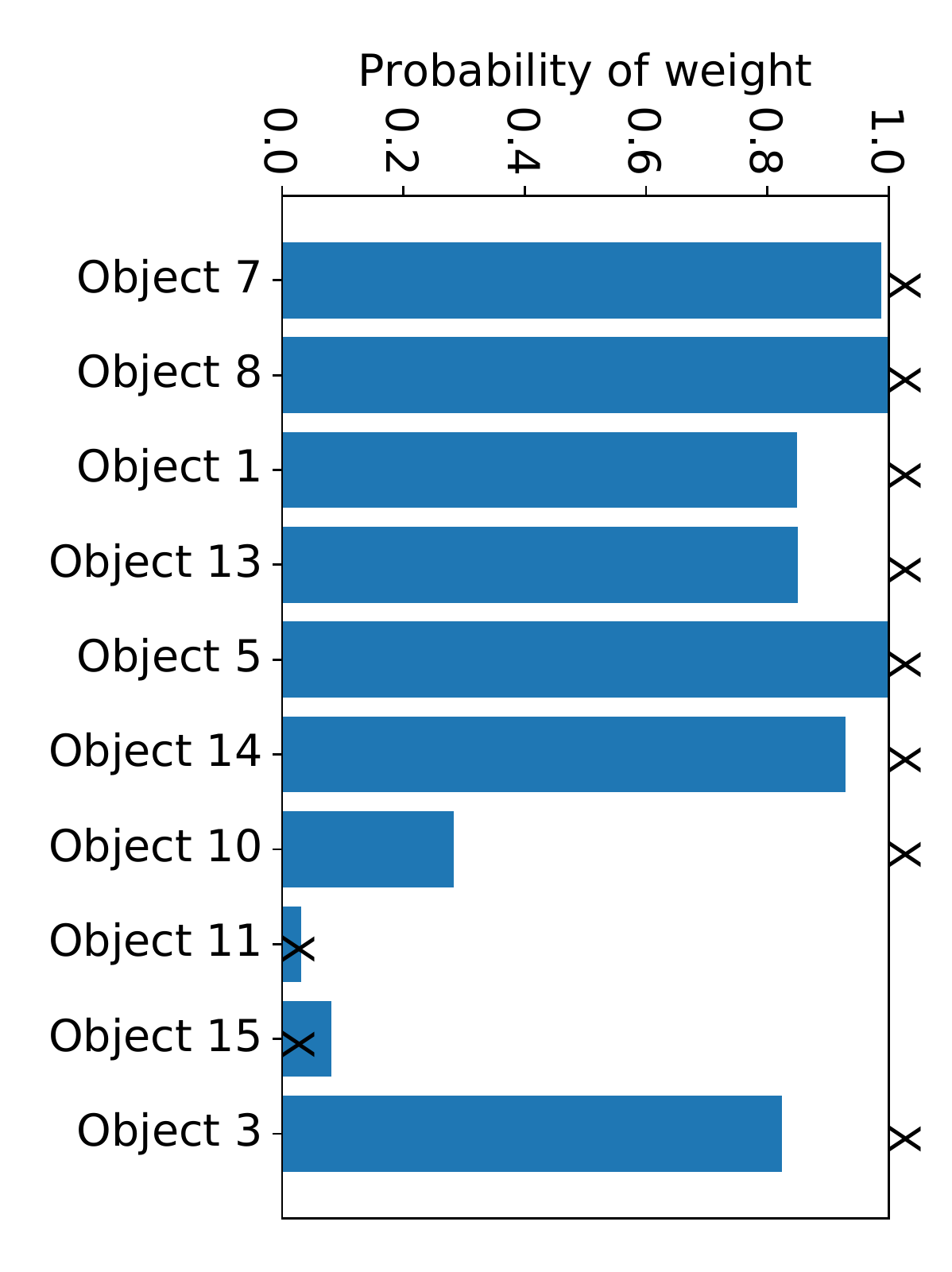}
		\caption{Local}\label{fig:cls_bars_1}
	\end{subfigure}
	\begin{subfigure}[b]{0.18\textwidth}
		\includegraphics[width=\textwidth]{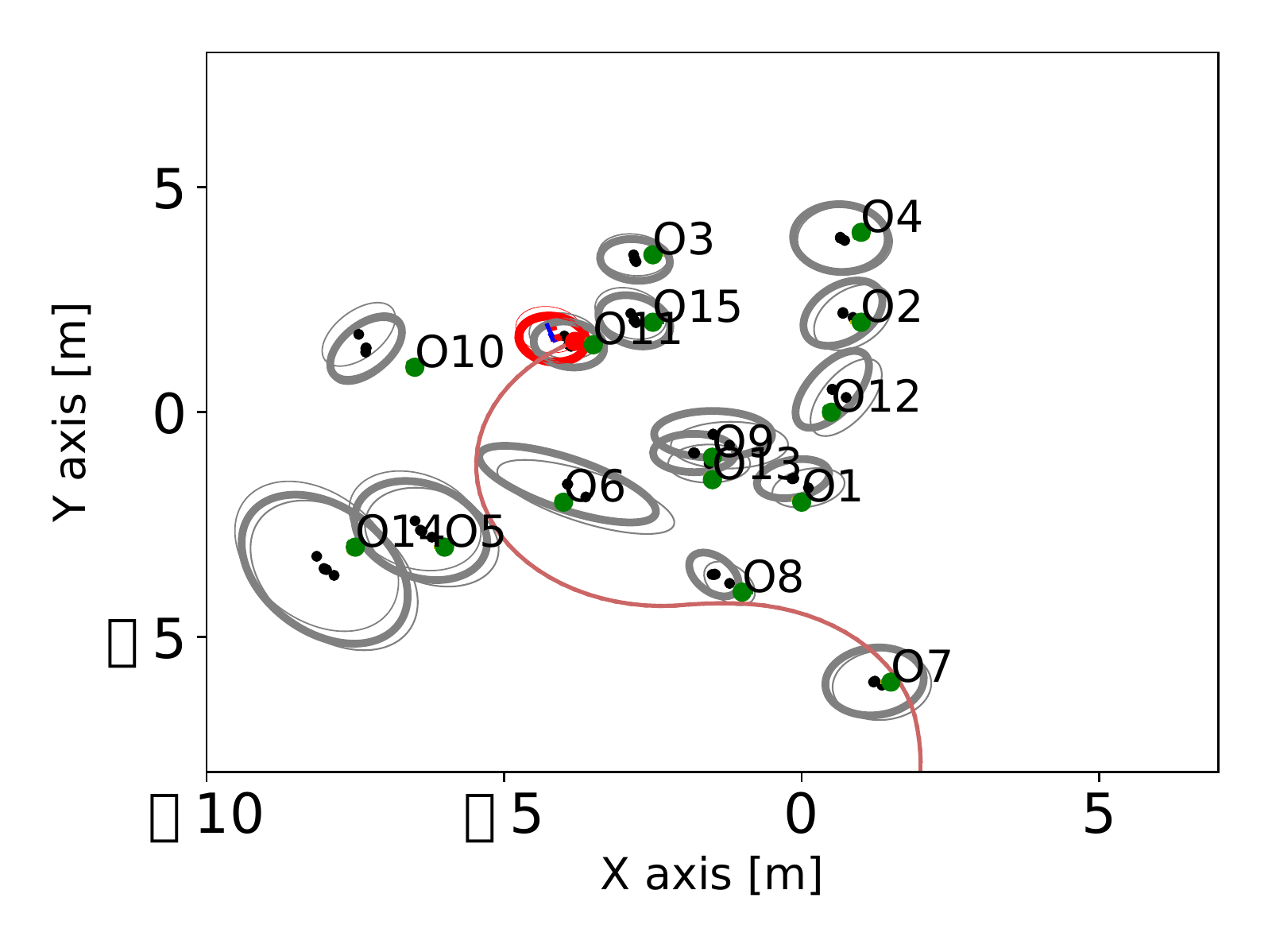}
		\caption{Distributed}\label{fig:cls_real_2}
	\end{subfigure}
	\begin{subfigure}[b]{0.18\textwidth}
		\includegraphics[width=\textwidth]{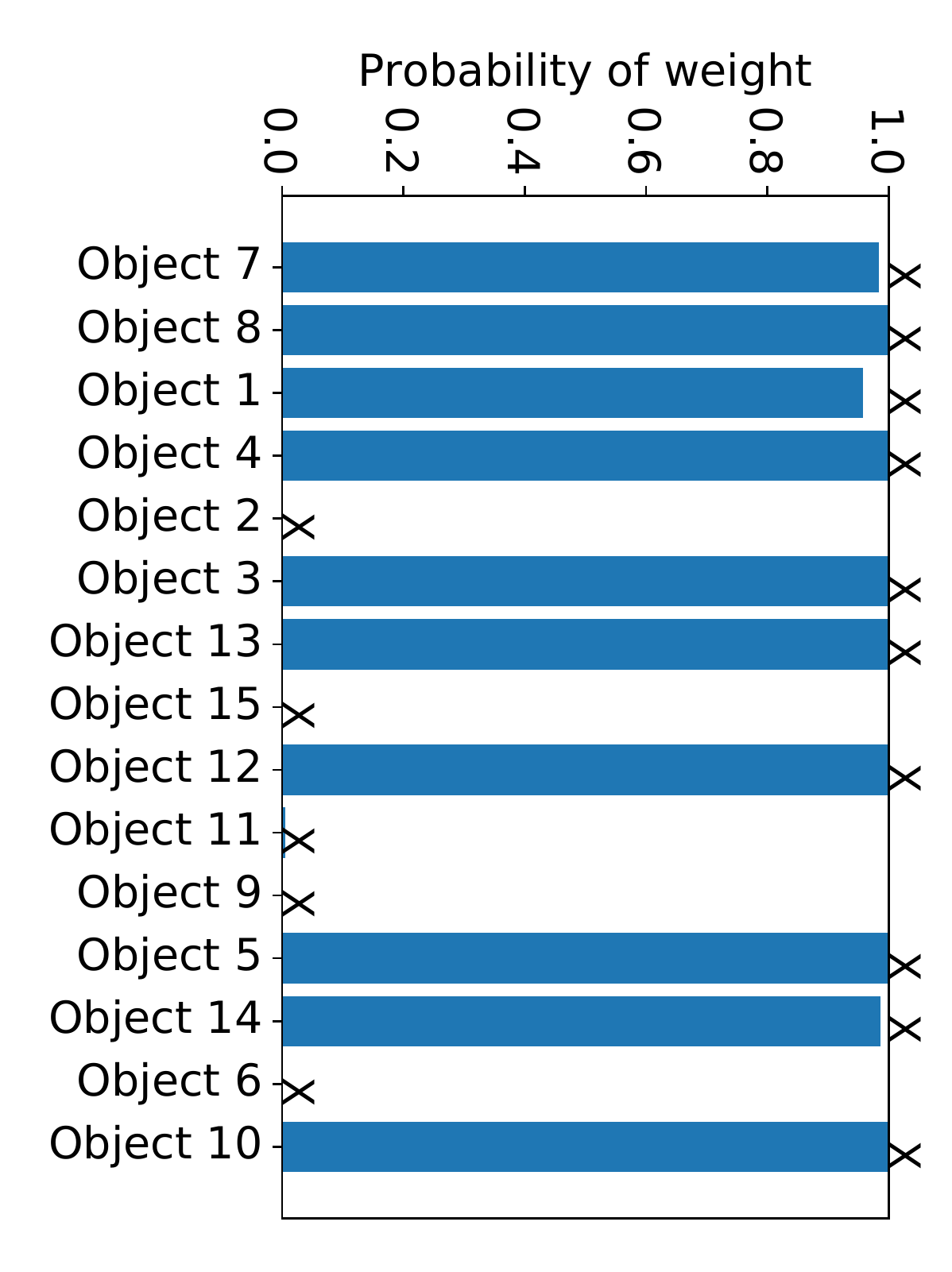}
		\caption{Distributed}\label{fig:cls_bars_2}
	\end{subfigure}
	\caption{ \textbf{(a)} presents average MSDE for the robots over 100 runs with different measurements. The rest are figures for time $k=60$ of $r_1$. \textbf{(b)} and \textbf{(d)} represent multiple SLAM hypotheses for local and distributed setting respectively; Black dots with gray ellipse represent object pose estimation, red \& blue signs with red ellipse represent robot pose estimation. Green and red points represent ground truth for object and robot positions respectively. \textbf{(c)} and \textbf{(e)} represent class probabilities for $c=1$ for objects observed thus far for local and distributed respectively. The $X$ notations represent ground truth (1 for class $c=1$, 0 for class $c=2$).}
	\label{fig:cls_graphs_sim}
\end{figure*}

As explained in Sec.~\ref{sec:disc_DC}, when double counting occurs, the posterior class probability will converge to extreme results quicker, and may result on either completely right or wrong classifications. Therefore, reasoning about a single run is insufficient, and a statistical study is required.
To quantify classification accuracy, we sample 100 times different geometric and semantic measurements, and perform a statistical study over the results. For that, we use mean square detection error (MSDE) averaged over all objects, robots, and runs (also used by Teacy et al. \cite{Teacy15aamas} and Feldman \& Indelman \cite{Feldman18icra}). We define MSDE per robot and object as follows:
\begin{equation}
MSDE \doteq \frac{1}{m} \sum_{i=1}^m (\mathbb{P}_{gt}(c=i) - \prob{c=i|\his^R_k})^2,
\end{equation}
where $\mathbb{P}_{gt}(c=i)$ represents the classification ground truth and can be either 1 for the correct class or 0 for all other classes. Therefore $MSDE = 1$ for completely incorrect classification, thus allowing us to perform statistical study of the effects of double counting of discrete random variables. To quantify localization accuracy, we use estimation error $\tilde{x}^{w_{avg}}$ which is the weighted average of Euclidean distance between the estimated and ground truth poses.

\subsection{Simulation Results}

Fig.~\ref{fig:x_wmax_figures} presents results for continuous variables, i.e.~robot and object poses. Figs.~\ref{fig:x_wmax} and \ref{fig:x_wmax_obj} show a clear advantage to our approach, where the localization error is the smallest for robots and objects respectively after the first 10 time steps. In Figs.~\ref{fig:Cov_avg} and \ref{fig:Cov_avg_obj} the estimation covariance is presented, where the double counted approach has the smallest values as expected. Fig.~\ref{fig:Cov_avg_obj} shows 'spikes' in the average objects' position covariance; these correspond to new object detections where the localization uncertainty is still high.

Fig.~\ref{fig:cls_graphs_sim} visualizes classification and estimations at time $k=60$ for local only and for distributed beliefs of robot $r_2$. At that time, robot $r_2$ communicated earlier with $r_3$, and for the first time communicates with $r_1$. When comparing Fig.~\ref{fig:cls_real_1} (local) to Fig.~\ref{fig:cls_real_2} (distributed), the number of possible class realizations is reduced. In addition, the estimate of $r_2$'s pose, as well as the objects, is more certain and accurate. When comparing Figs.~\ref{fig:cls_bars_1} and ~\ref{fig:cls_bars_2}, the latter presents a larger map, i.e. more objects observed, and the class estimations (classification) are closer to the ground truth.

Fig.~\ref{fig:MSDE_statistical} presents the average MSDE over 100 runs, where as a whole our approach shows lower MSDE values, i.e. statistically stronger classification results.
In supplementary material we present additional classification and SLAM results.

\subsection{Experiment Setting}

\begin{figure}[!htbp]
	
	\begin{subfigure}[b]{0.2\textwidth}
		\includegraphics[width=\textwidth]{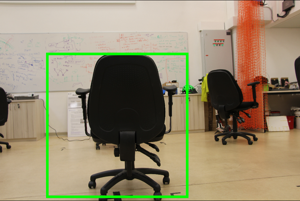}
		\caption{}\label{fig:Chair1}
	\end{subfigure}
	%
%
	\begin{subfigure}[b]{0.32\textwidth}
		\includegraphics[width=\textwidth]{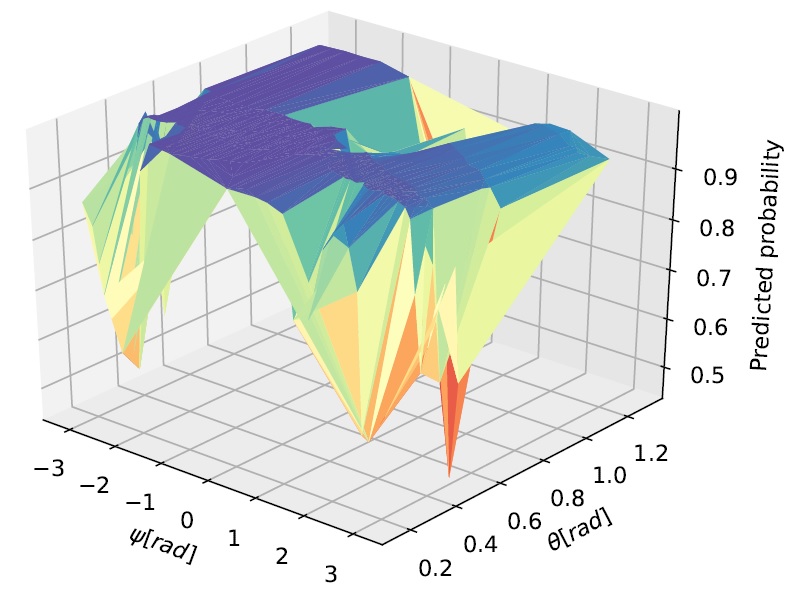}
		\caption{}\label{fig:c1_1}
	\end{subfigure}
	\begin{subfigure}[b]{0.32\textwidth}
		\includegraphics[width=\textwidth]{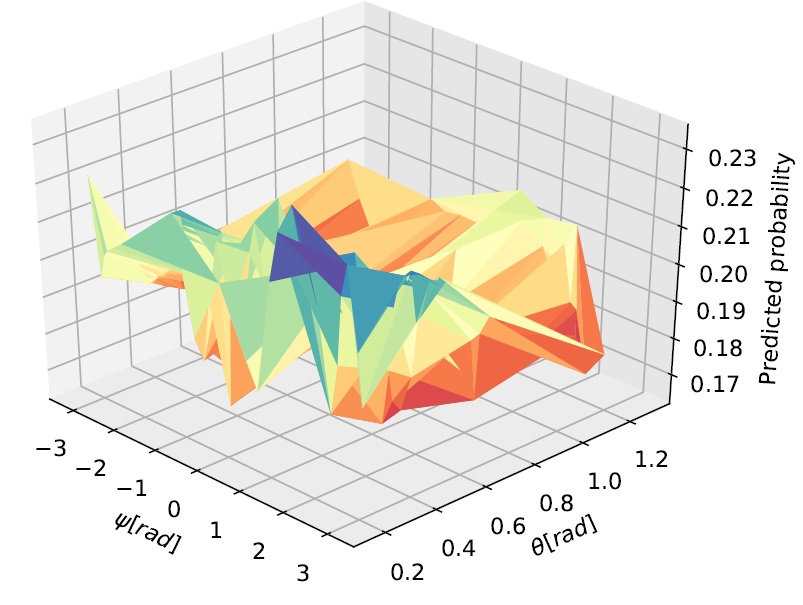}
		\caption{}\label{fig:c1_2}
	\end{subfigure}
	\caption{\textbf{(a)} is an image used in the experiment, with corresponding the bounding box. \textbf{(b)} and \textbf{(c)} are class probability expectation for class $c=1$ for classifier models of $c=1$ and $c=2$ respectively.}
	\label{fig:cls_model}
\end{figure}

In our scenario 3 robots are moving within an environment with multiple objects within it. We scattered 6 chairs within the environment and photographed them using a camera on a stand, keeping a constant height. In Fig.~\ref{fig:Chair1} we show an image from the scenario with the corresponding bounding box. The chairs were detected with YOLO3 DarkNet detector \cite{Redmon18arxiv}, which provided bounding boxes, and then each bounding box was classified using a ResNet50 convolutional neural network \cite{He16cvpr}. We considered 3 candidate classes out of 1000: 'barber chair', 'punching bag', and 'traffic light', as $c=1,2,3$ respectively with $c=1$ being the ground truth class. 
We trained three viewpoint-dependent classifier models using three sets of relative pose and class probability vector pairs, with the spatial parameters being the yaw and pitch angles from camera to object; The models are presented in the supplementary material Sec.~9. For the ground truth class we photographed an objects from multiple viewpoints, and then classified it using ResNet 50. For the other two classifier models, we sampled class probability vectors with larger probability for the corresponding class of the model, and used the same relative poses as the first model. Fig.~\ref{fig:c1_1},~\ref{fig:c1_2} presents expectation of $c=1$ for two of the classifier models as a function of the spatial parameters.

In the experiment (deployment phase), we utilized both geometric and semantic measurements, using the corresponding (learned) measurement likelihood models. Relative pose geometric measurements for odometry and between camera and objects were generated by corrupting ground truth with Gaussian noise, while the semantic measurements are provided by YOLO3 and ResNet from real images. For parameter details, see supplementary material Sec.~9. The same metrics as the simulation are used here.

\subsection{Experimental Results}

\begin{figure*}[!htbp]

	\begin{subfigure}[b]{0.18\textwidth}
		\includegraphics[width=\textwidth]{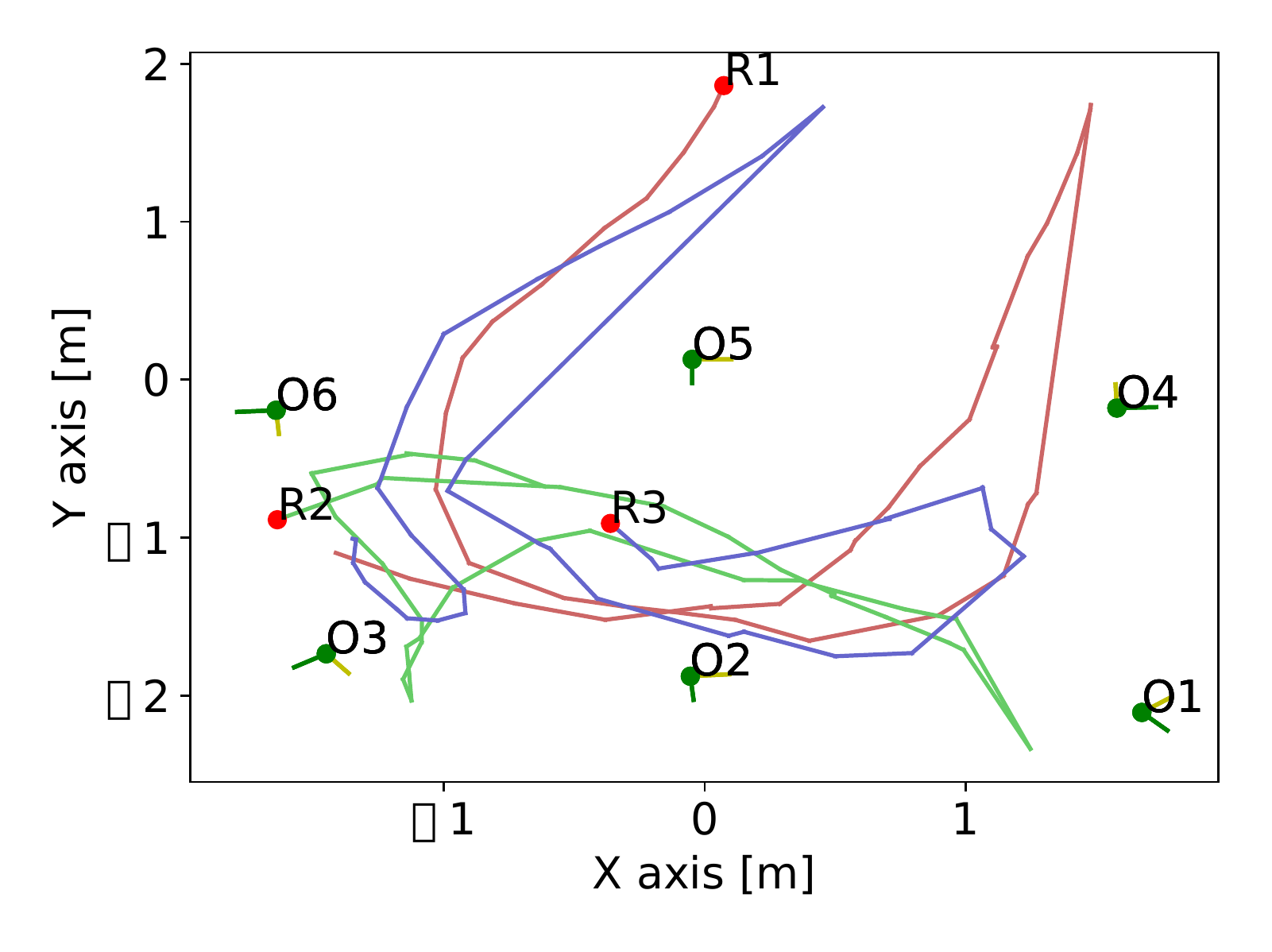}
		\caption{Ground Truth}\label{fig:GT_paths_real}
	\end{subfigure}
	\begin{subfigure}[b]{0.18\textwidth}
		\includegraphics[width=\textwidth]{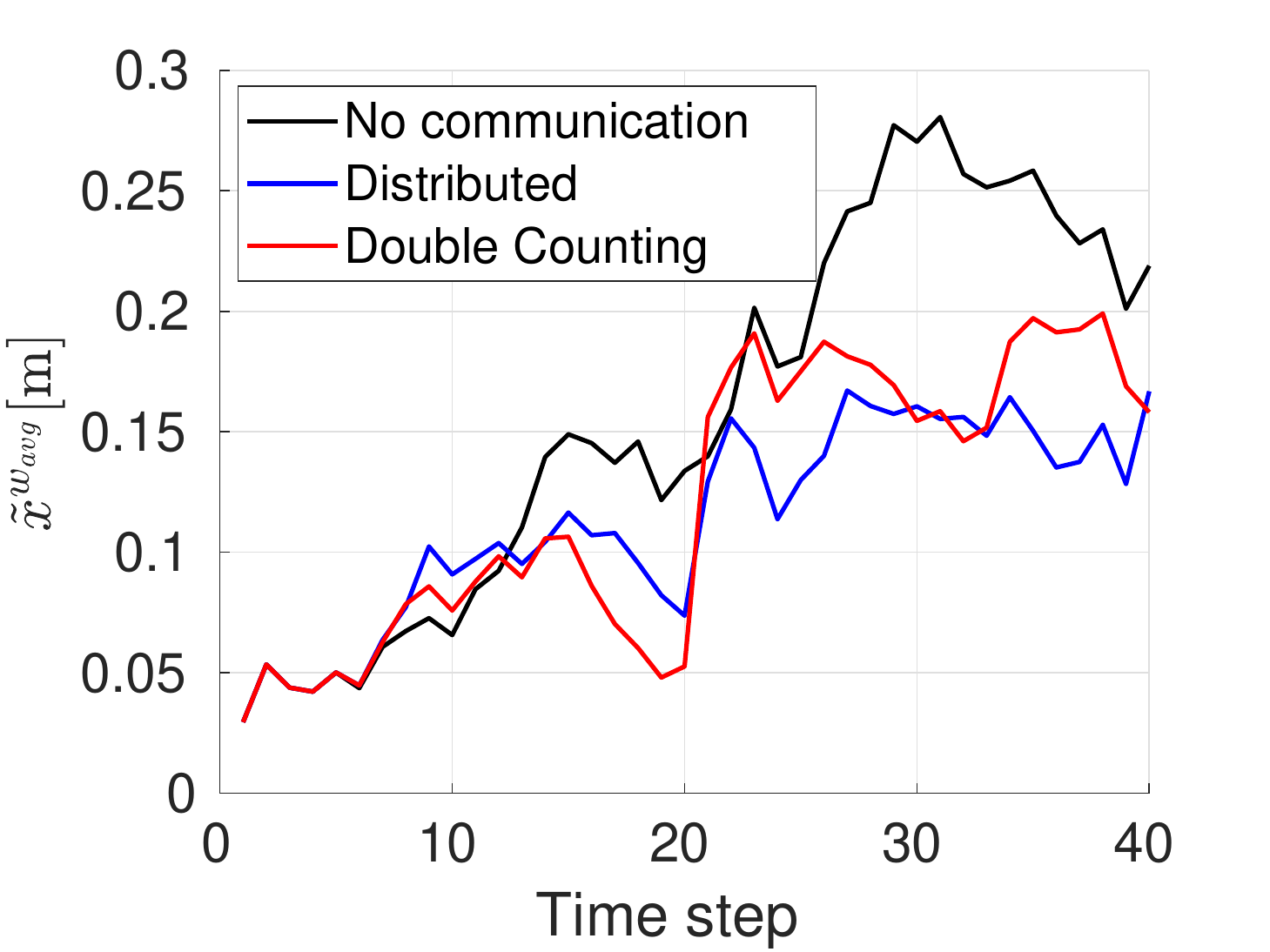}
		\caption{Robot position}\label{fig:real_x_wmax}
	\end{subfigure}
	\begin{subfigure}[b]{0.18\textwidth}
		\includegraphics[width=\textwidth]{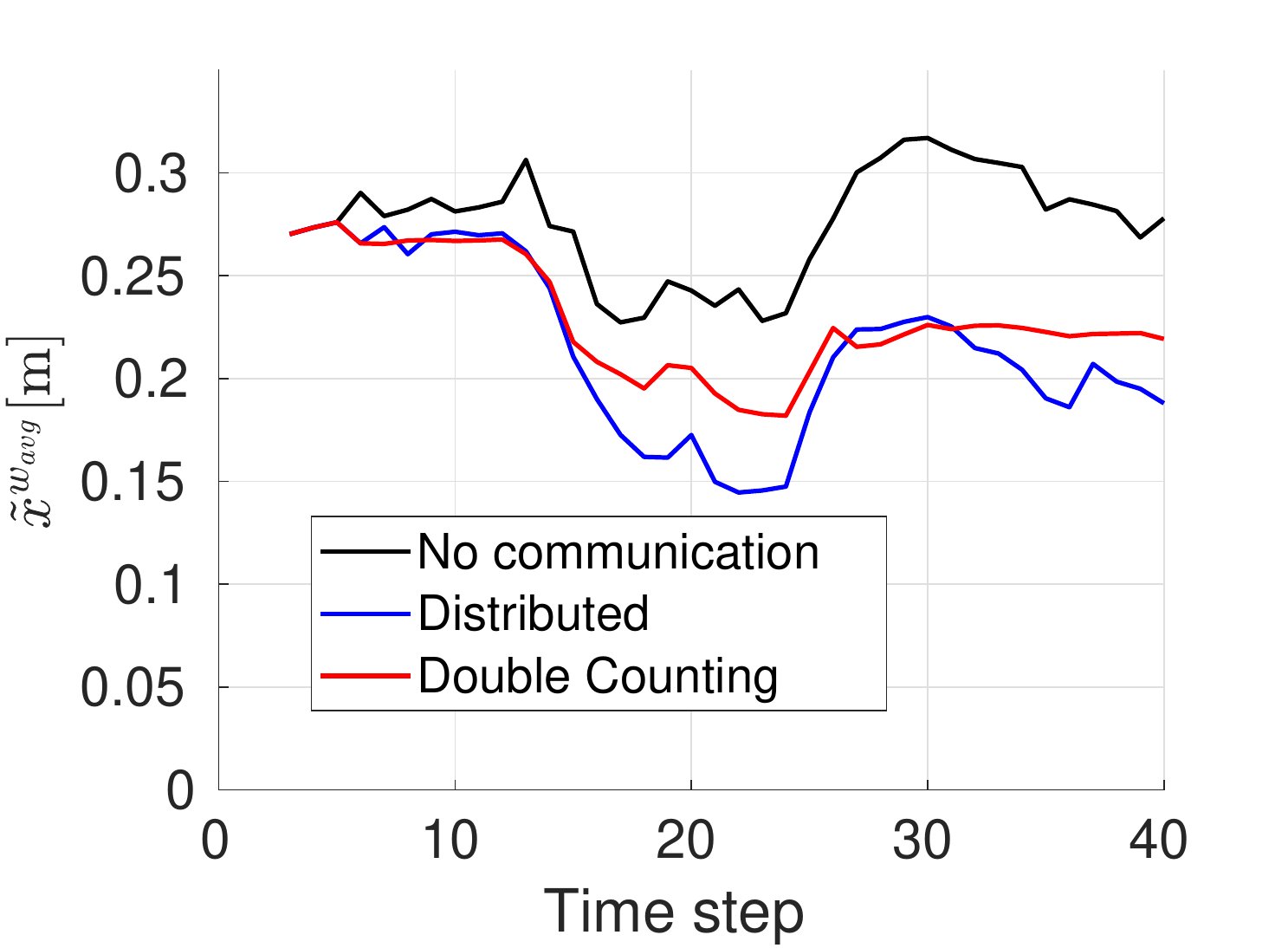}
		\caption{Object position}\label{fig:real_x_wmax_obj}
	\end{subfigure}
	\begin{subfigure}[b]{0.18\textwidth}
		\includegraphics[width=\textwidth]{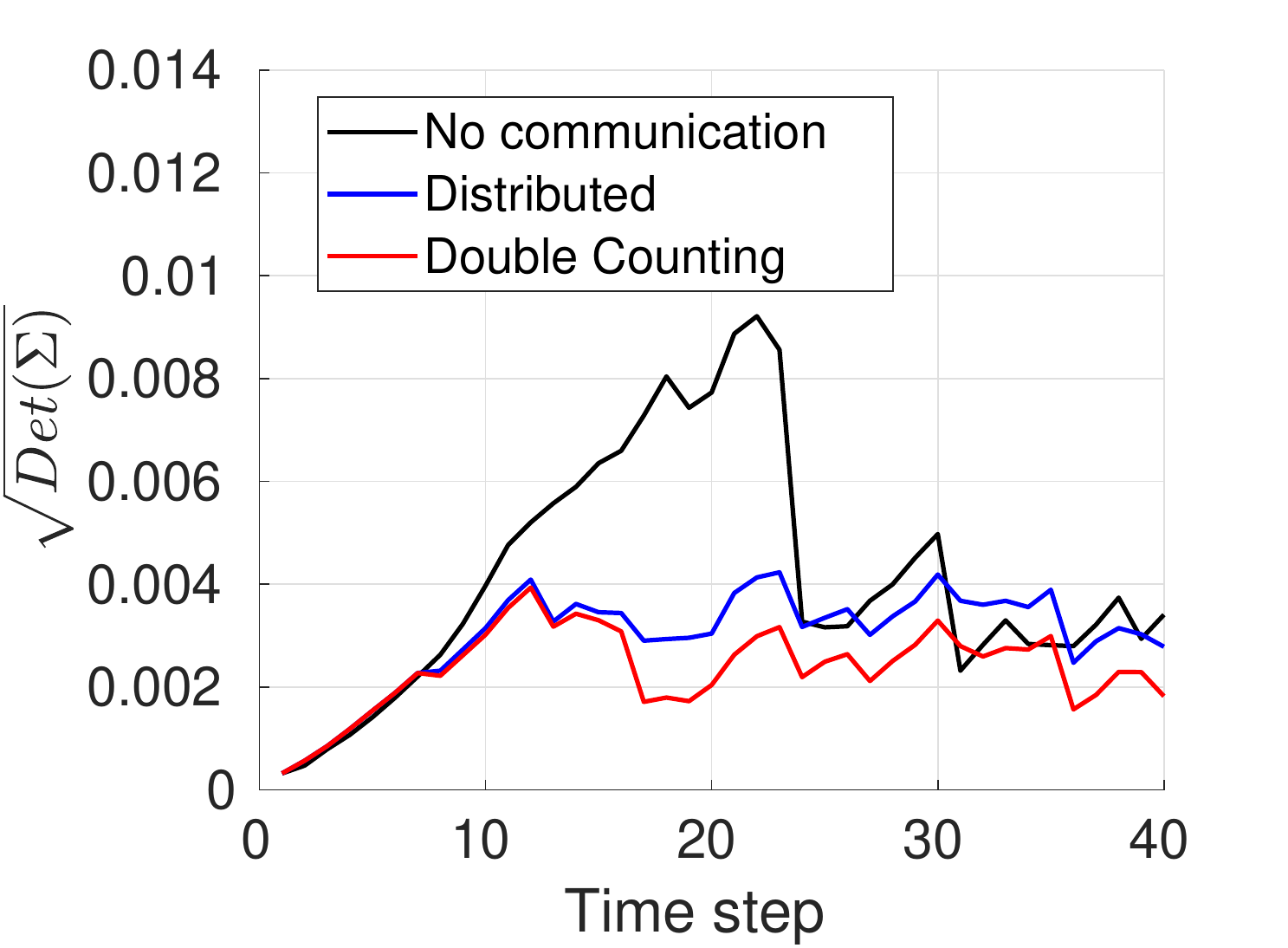}
		\caption{Robot covariance}\label{fig:real_Cov_avg}
	\end{subfigure}
	\begin{subfigure}[b]{0.18\textwidth}
		\includegraphics[width=\textwidth]{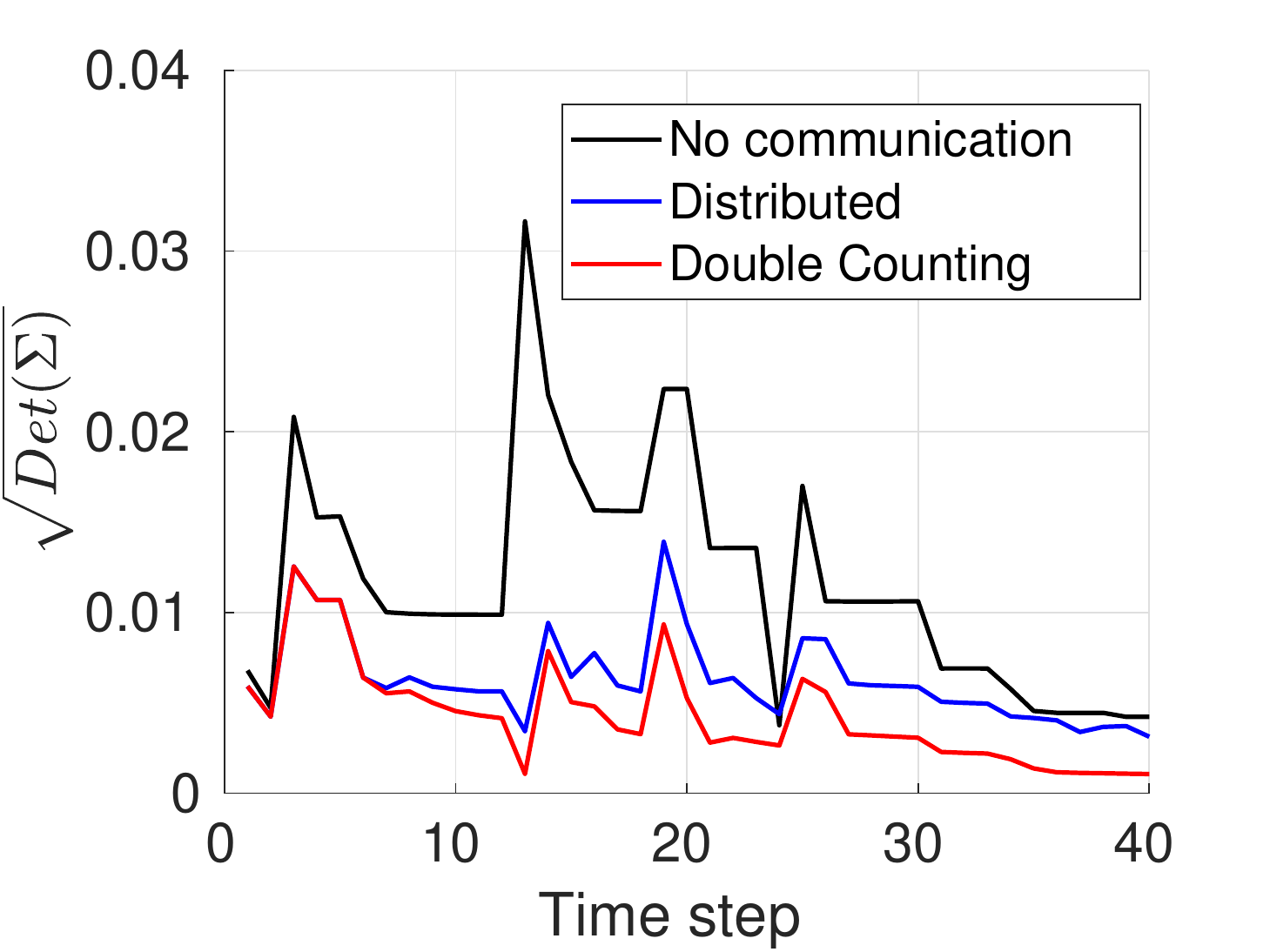}
		\caption{Object covariance}\label{fig:real_Cov_avg_obj}
	\end{subfigure}
	\vspace{-0.2cm}
	\caption{Experiment figures; \textbf{(a)} present the ground truth of the scenario. Red points represent the initial position of the robots, with different colored lines represent different robots. The green points represent the object poses. \textbf{(b)} and \textbf{(c)} represent the average $\tilde{x}^{w_{\text{avg}}}$ for robot and object positions respectively as a function of time for the experiment. \textbf{(d)} and \textbf{(e)} present the corresponding square-root of the position covariance for the robot and object average respectively.}
	\label{fig:real_x_wmax_figures}
\end{figure*}

Fig.~\ref{fig:real_x_wmax_figures} presents SLAM results for the same benchmarks as in Fig.~\ref{fig:x_wmax_figures}. Figs.~\ref{fig:real_x_wmax} and \ref{fig:real_x_wmax_obj} present an  average $\tilde{x}^{w_{\text{avg}}}$ over all robots for robot and object positions, respectively. In general, the advantage of our approach is evident with lower errors. In addition, Figs.~\ref{fig:real_Cov_avg} and \ref{fig:real_Cov_avg_obj} present a similar pattern to Figs.~\ref{fig:Cov_avg} and \ref{fig:Cov_avg_obj},  respectively, where the  covariance of our approach is smaller than the single robot case, but larger than the over-confident double counting case.

\begin{figure*}[!htbp]
	
	\begin{subfigure}[b]{0.18\textwidth}
		\includegraphics[width=\textwidth]{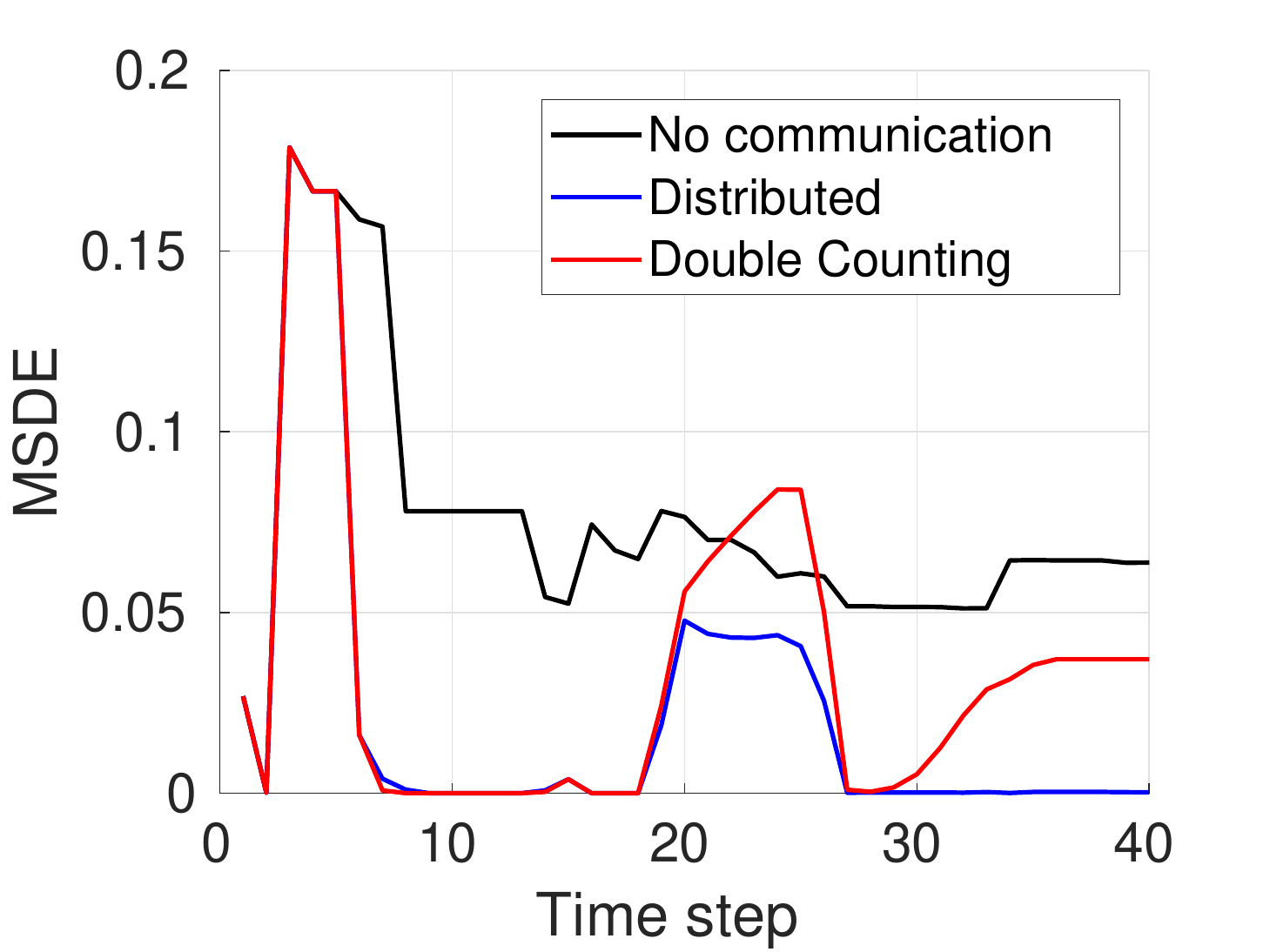}
		\caption{MSDE}\label{fig:real_MSDE}
	\end{subfigure}
	\begin{subfigure}[b]{0.18\textwidth}
		\includegraphics[width=\textwidth]{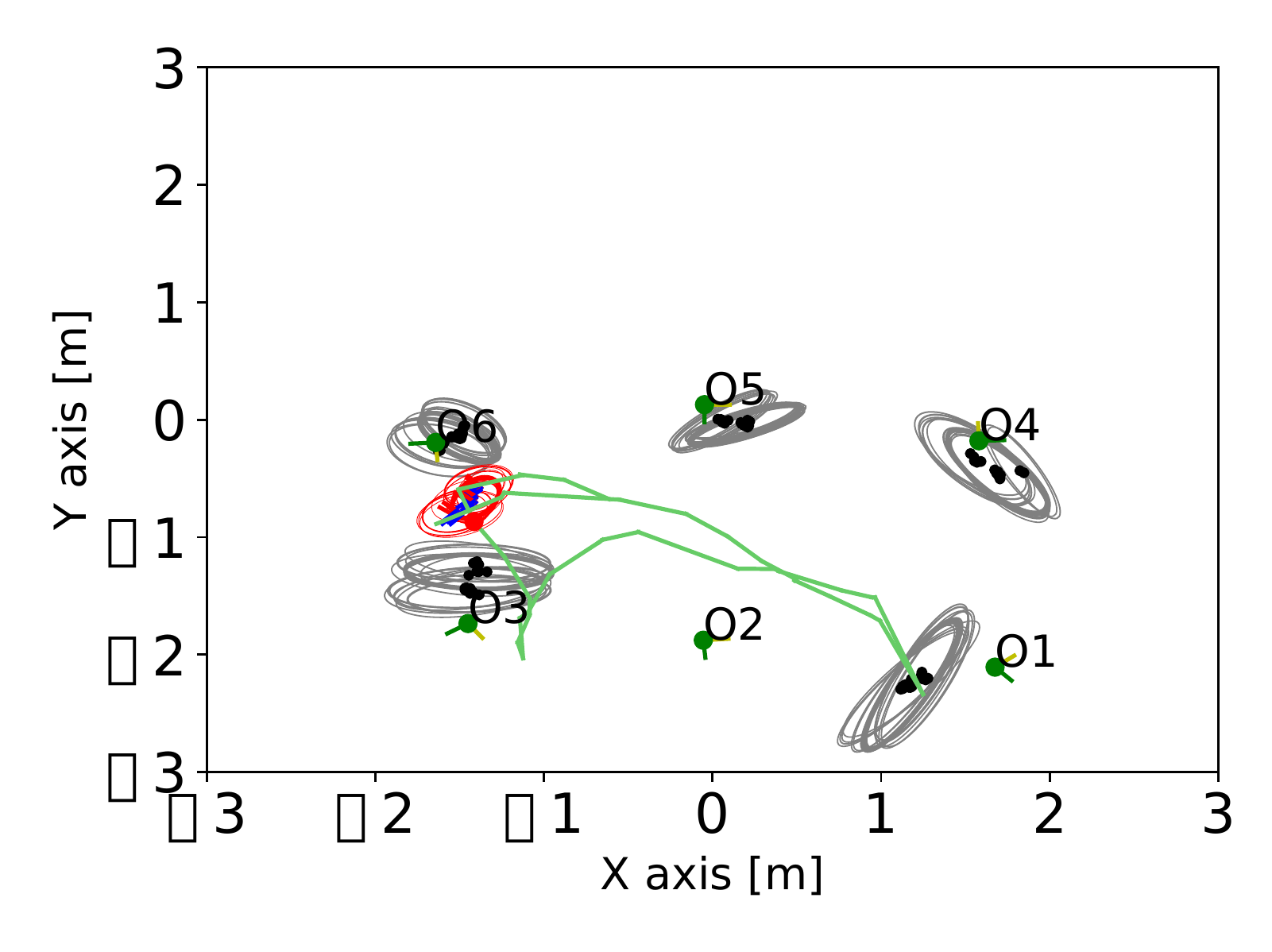}
		\caption{Local}\label{fig:real_cls_real_1}
	\end{subfigure}
	\begin{subfigure}[b]{0.18\textwidth}
		\includegraphics[width=\textwidth]{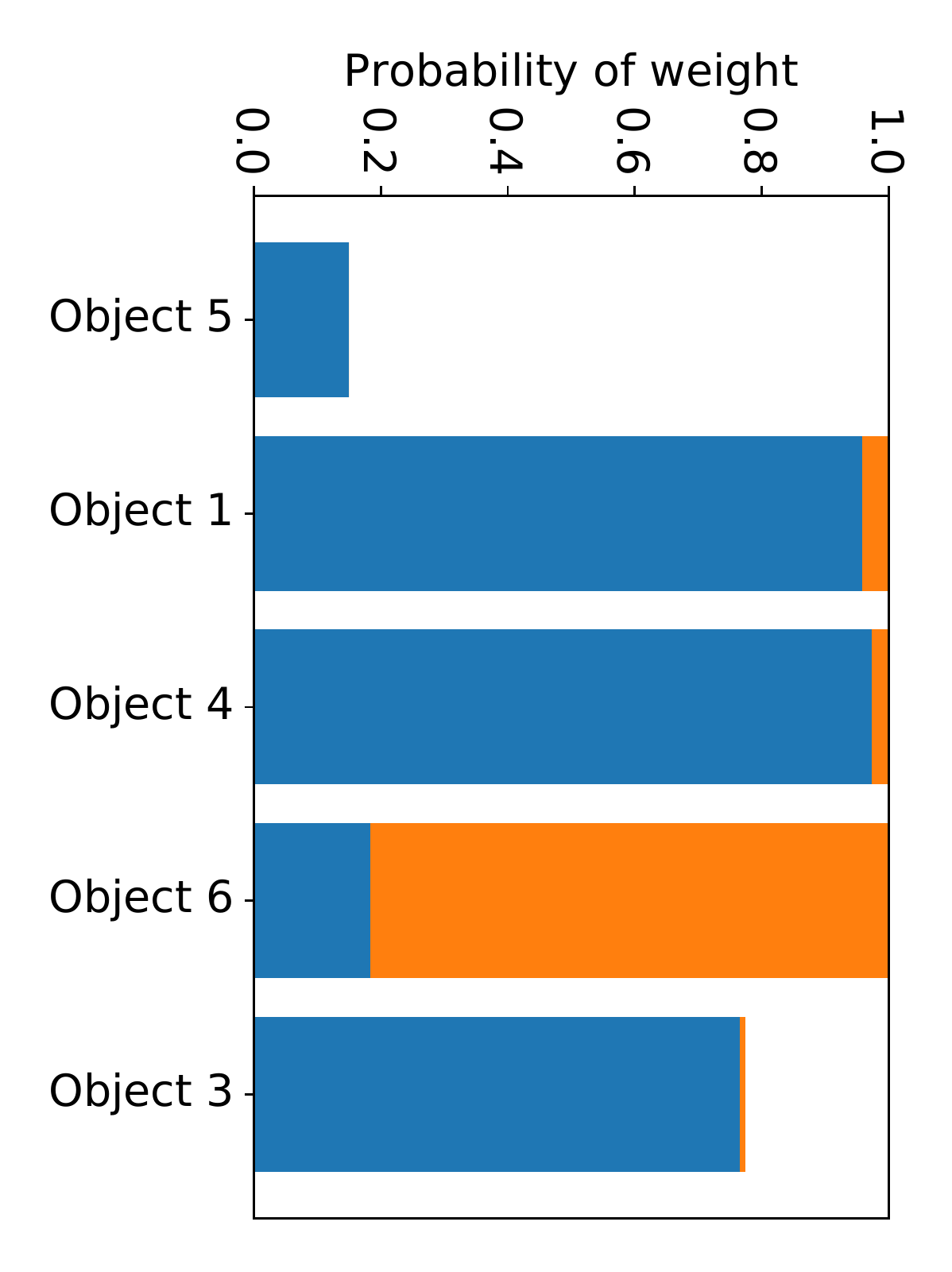}
		\caption{Local}\label{fig:real_cls_bars_1}
	\end{subfigure}
	\begin{subfigure}[b]{0.18\textwidth}
		\includegraphics[width=\textwidth]{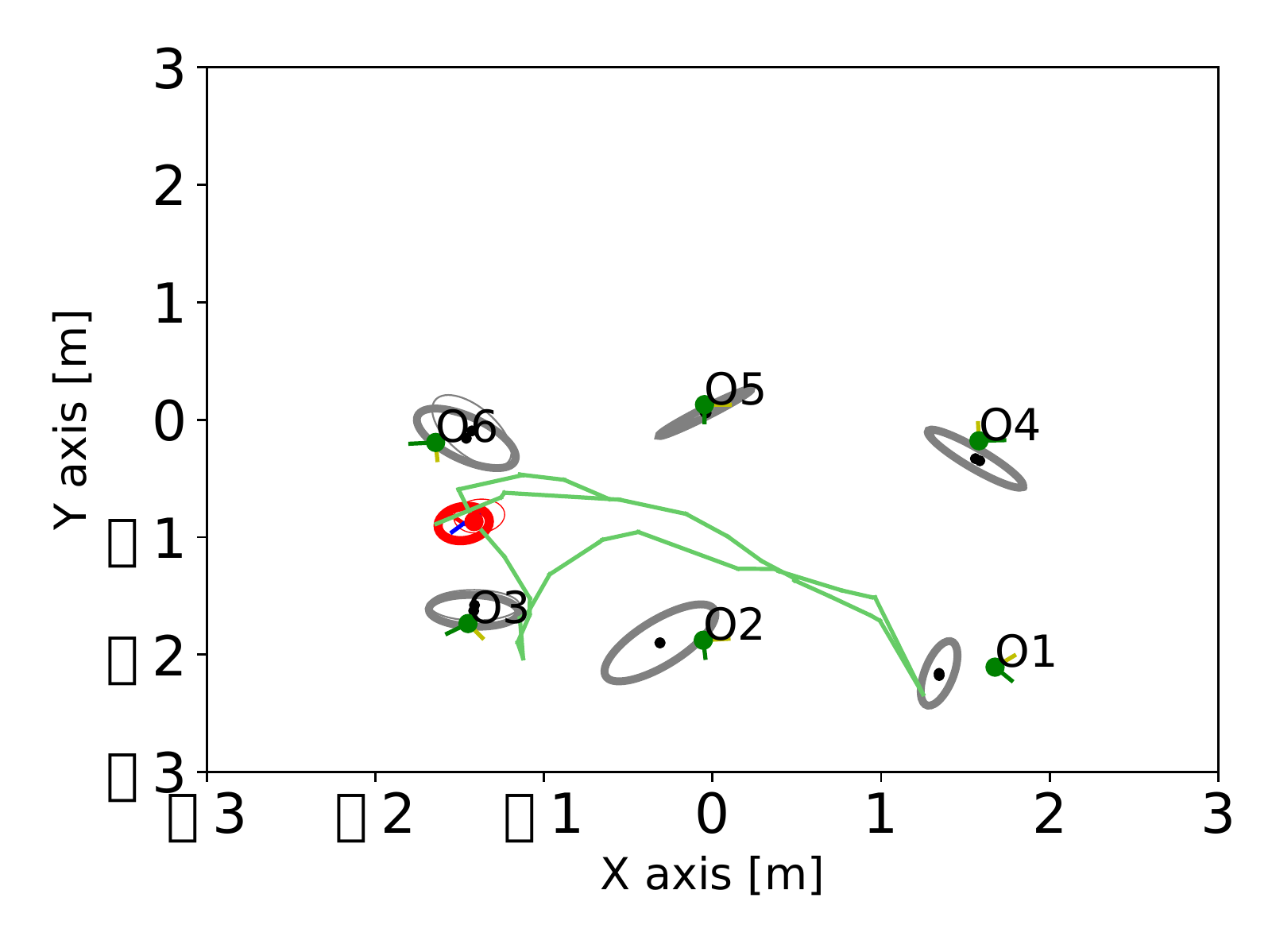}
		\caption{Distributed}\label{fig:real_cls_real_2}
	\end{subfigure}
	\begin{subfigure}[b]{0.18\textwidth}
		\includegraphics[width=\textwidth]{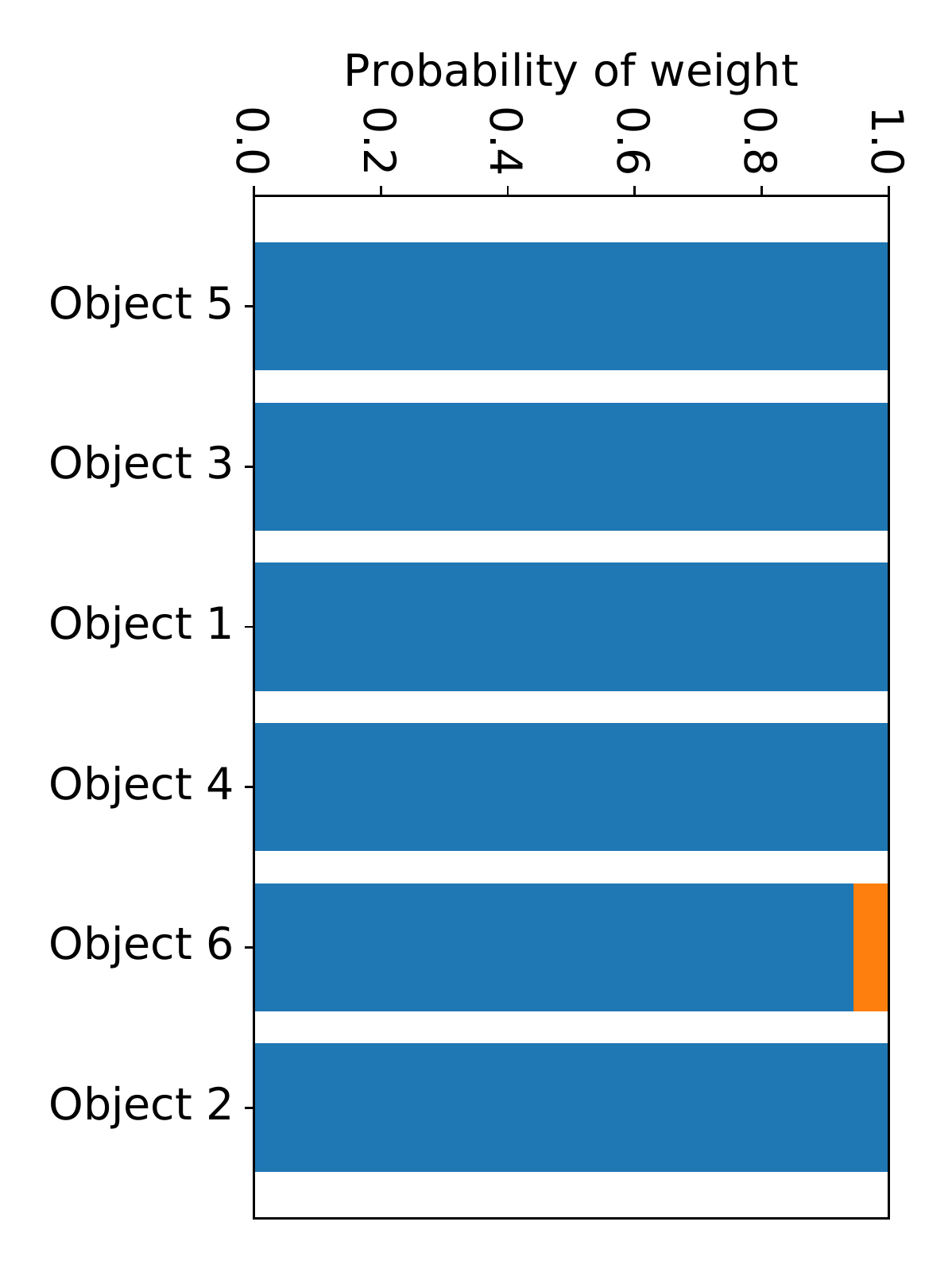}
		\caption{Distributed}\label{fig:real_cls_bars_2}
	\end{subfigure}
	
	\caption{\textbf{(a)} presents average MSDE for the robots over 100 runs with different measurements. The rest are figures for time $k=35$ of $r_2$. \textbf{(b)} and \textbf{(d)} represent multiple SLAM hypotheses for local and distributed setting respectively; Black dots with gray ellipse represent object pose estimation, red \& blue signs with red ellipse represent robot pose estimation. Green and red points represent ground truth for object and robot poses respectively. \textbf{(c)} and \textbf{(e)} represent class probabilities for $c=1$ and $c=2$ for objects observed thus far for local and distributed respectively, with {\color{blue}blue} and {\color{orange}orange} for classes 1 and 2 respectively. In this case, the ground truth class of all objects is $c=1$.}
	\label{fig:real_cls_graphs_sim}
	
\end{figure*}

For classification results, Fig.~\ref{fig:real_MSDE} shows the average MSDE per robot as a function of time step, where eventually our approach out-performs both the single robot and the double counting cases, with higher probability for the correct class realization. In Fig.~\ref{fig:real_cls_graphs_sim}, SLAM and classification results for Robot 2 at time step $k=35$ are presented, showing similar resulting trends to Fig.~\ref{fig:cls_graphs_sim}. Comparing Fig.~\ref{fig:real_cls_real_1} and Fig.~\ref{fig:real_cls_real_2}, the later shows more accurate SLAM compared to the former, with less class realizations. In addition, compared to Fig.~\ref{fig:real_cls_bars_2},  Fig.~\ref{fig:real_cls_bars_1} shows  more accurate classification with an additional object classified.

For additional results at different time steps, refer to the supplementary material Sec.~10-11 and multimedia submission.

	\section{Conclusions}
	\label{sec:conclusions}

We presented an approach for multi-robot semantic SLAM in an unknown environment. In this approach a distributed hybrid belief is maintained per robot using local information transmitted to other robots as a 'stack', designed to keep estimation consistency without complex book-keeping, both for continuous and discrete states. We utilized a viewpoint dependent classifier model to account for the coupling of relative pose between robot and object, and object's class. In simulation and real-world experiment we showed that our approach improves classification and localization performance while avoiding double counting. Future work will incorporate data association disambiguation.

	
	
	\bibliographystyle{IEEEtran}
	\bibliography{refs}

	
\end{document}


\title{Supplementary Material \\ \vspace{0.5cm} \small{Technical Report ANPL-2020-01} }

\author{\renewcommand\footnotemark{}Vladimir Tchuiev and Vadim Indelman
	\thanks{The authors are with the Department of Aerospace Engineering, Technion - Israel Institute of Technology, Haifa 32000, Israel. {\tt\{vovatch, vadim.indelman\}@technion.ac.il}.
	}	
	\vspace{-15pt}
}

\date{}

\maketitle
  

\section{Supplementary Derivation}

We present a relation that is used in equation derivation; Let $A$ be a random variable conditioned on the set $\{B_i\}$ of random variables $B_i$ that are independent from each other. By using Bayes Law, we can split the conditional probability $\prob{A|\{B_i\}}$ to a product of conditional probabilities:
\begin{equation}
\prob{A|\{B_i\}} = \frac{\prob{\{B_i\}|A} \prob{A}}{\prob{\{B_i\}}} =\! 
\frac{\prod_i \prob{B_i|A}}{\prod_i \prob{B_i}} \prob{A}.
\end{equation}
Using Bayes Law again on each element in the product, we reach the following expression:
\begin{equation}\label{eq:AB_Relations}
\prob{A|\{B_i\}} = \prod_i \left( \frac{\prob{A|B_i}}{\prob{A}} \right) \prob{A}.
\end{equation}
This allow to express a random variable as a multiplication of conditionals, which will be useful to separate local and external measurements.

\section{Derivation of $\prob{\poses^R_k \backslash x^r_k|\classes^R_k,\his^{R-}_k}$}

Recall splitting $\his^{R-}_k$ into prior history $\his^R_{k-1}$ and non-local measurements \& actions:
\begin{equation}
	\his^{R-}_k = \his^R_{k-1} \cup \hisnew^{R-}_k.
\end{equation}
We then use the above definition and relation \eqref{eq:AB_Relations} to split $\prob{\poses^R_k \backslash x^r_k|\classes^R_k,\his^{R-}_k}$ into a product of two beliefs, one that depends on prior history, and one that depends on external new measurements:
\begin{equation}\label{eq:Joint_Cont_Dev_2}
	\prob{\poses^R_k \backslash x^r_k|\classes^R_k,\his^{R-}_k} = 
	\prob{\poses^R_k \backslash x^r_k|\classes^R_k}l
	\frac{\prob{\poses^R_k \backslash x^r_k|\classes^R_k,\hisnew^{R-}_k}}
	{\prob{\poses^R_k \backslash x^r_k|\classes^R_k}}
	\frac{\prob{\poses^R_k \backslash x^r_k|\classes^R_k,\his^{R}_{k-1}}}
	{\prob{\poses^R_k \backslash x^r_k|\classes^R_k}}.
\end{equation}
This formulation allows us to isolate the new information sent by other robots at time $k$, from information already used for inference at previous times. Next, we have to address that not all known objects are present in the sent local beliefs.
Because the priors are assumed independent between poses and classes, $\prob{\poses^R_k \backslash x^r_k|\classes^R_k} = \prob{\poses^R_k \backslash x^r_k}$. From $\prob{\poses^R_k \backslash x^r_k|\classes^R_k,\his^{R}_{k-1}}$ we can split $\poses^R_k \backslash x^r_k$ into poses of objects that are involved in $\his^{R}_{k-1}$ and ones that do not as:
\begin{equation}
	\prob{\poses^R_k \backslash x^r_k|\classes^R_k,\his^{R}_{k-1}} =
	\prob{\poses^{o,R}_{\text{new},k}|\classes^R_{\text{new},k},\his^R_{k-1}} 
	\prob{\poses^R_{k-1}|\classes^R_k,\his^{R}_{k-1}}.
\end{equation}
Poses of objects that $r$ wasn't aware of at time $k-1$ are independent of $\his^R_{k-1}$, and without measurements, $\poses^{o,R}_{\text{new},k}$ are independent of $\classes^R_{\text{new},k}$ as well. In addition, $\poses^R_{k-1}$ is independent of classes of objects that are observed only at time $k$, thus we can write:
\begin{equation}\label{eq:Joint_Cont_Div_1}
	\prob{\poses^R_k \backslash x^r_k|\classes^R_k,\his^{R}_{k-1}} =
	\prob{\poses^{o,R}_{\text{new},k}} \cdot
	\prob{\poses^R_{k-1}|\classes^R_{k-1},\his^R_{k-1}},
\end{equation}
which is the prior for poses of newly known objects at time step $k$, multiplied by the conditional continuous belief for objects already known.
Similarly to Eq.~\eqref{eq:Joint_Cont_Div_1}, the prior for $\poses^{R}_k \backslash x^r_k$ is separated to previously known and new objects:
\begin{equation}\label{eq:Joint_Cont_Div_2}
	\prob{\poses^R_k \backslash x^r_k} = \prob{\poses^{o,R}_{\text{new},k}}
	\prob{\poses^R_{k-1}},
\end{equation}
therefore we can write:
\begin{equation}\label{eq:Joint_Cont_Dev_3}
	\frac{\prob{\poses^R_k \backslash x^r_k|\classes^R_k,\his^{R}_{k-1}}}
	{\prob{\poses^R_k \backslash x^r_k|\classes^R_k}} =
	\frac{b^R_{k-1}}{\prob{\poses^R_{k-1}}},
\end{equation}
and substitute it into the rightmost fracture in Eq.~\eqref{eq:Joint_Cont_Dev_2}. Then we cancel out $\prob{\poses^R_k \backslash x^r_k|\classes^R_k}$ and proceed to remove $r$'s poses from the prior by:
\begin{equation}\label{eq:Joint_Cont_Dev_4}
	\frac{\prob{\poses^R_k \backslash x^r_k|\classes^R_k,\hisnew^{R-}_k}}
	{\prob{\poses^R_{k-1}}} =
	\frac{\prob{\poses^{o,R}_k|\classes^R_k,\hisnew^{R-}_k}}
	{\prob{\poses^{o,R}_{k-1}}},
\end{equation}
as robot $r$'s poses up until time $k-1$ are independent from the new external measurements. Finally, after factoring out $\prob{\poses^R_k \backslash x^r_k|\classes^R_k}$, and Eq.~\eqref{eq:Joint_Cont_Dev_3} and \eqref{eq:Joint_Cont_Dev_4} with Eq.~\eqref{eq:Joint_Cont_Dev_2} we reach the following expression that is used in the paper:
\begin{equation}
	\prob{\poses^R_k \backslash x^r_k|\classes^R_k,\his^{R-}_k} = 
	b^R_{k-1} \cdot
	\frac{\prob{\poses^{o,R}_k|\classes^{o,R}_k,\hisnew^{R-}_k}}
	{\prob{\poses^{o,R}_{k-1}}}
\end{equation}


\section{Dependency Between Object Classes}

In this section we show that the classes of two objects are not independent. We present a simple example where $c_1$ and $c_2$ be the underlying classes of objects 1 and 2 respectively. Let $\his^R$ be the total measurement history, including semantic measurements $z^{sem}_1$ and $z^{sem}_2$ for objects 1 and 2 respectively. Recall that measurements are assumed independent from each other. Using the Bayes Law:
\begin{equation}
\prob{c_1,c_2|\his^R} \propto \prob{c_1,c_2|\his^R \backslash z^{sem}_1, z^{sem}_2}
\prob{z^{sem}_1|c_1} \prob{z^{sem}_2|c_2}.
\end{equation}
We use a viewpoint dependent classifier model, so we must marginalize $\prob{z^{sem}_1|c_1} \prob{z^{sem}_2|c_2}$ by the corresponding relative viewpoints, denoted $x^{rel}_1$ and $x^{rel}_2$ respectively:
\begin{equation}
\prob{z^{sem}_1|c_1} \prob{z^{sem}_2|c_2} = 
\int_{x^{rel}_1, x^{rel}_2} \prob{z^{sem}_1|c_1, x^{rel}_1} \prob{z^{sem}_2|c_2, x^{rel}_2}
\prob{x^{rel}_1, x^{rel}_2 | \his^R} dx^{rel}_1 dx^{rel}_2.
\end{equation}
From the above equation, the condition for $c_1$ and $c_2$ to be independent is that $x^{rel}_1$ and $x^{rel}_2$ must be independent, which is not true in the general case, thus $c_1$ and $c_2$ are dependent.


\section{Derivation of ${\color{blue}\frac{\prob{\poses^{o,R}_k|\classes^R_k,\hisnew^{R-}_k}}{\prob{\poses^{o,R}_{k}}}}$
(Continuous Belief Update)}

Using the relation \eqref{eq:AB_Relations} we can split the blue part by separating the new measurements and actions per robot, excluding $r$ itself as it is not present in $\hisnew^{R-}_k$:
\begin{equation}\label{eq:Stack_Init_Cont}
	\prob{\poses^{o,R}_k|\classes^R_k,\hisnew^{R-}_k} =
	\prod_{k',r' \in R \backslash r}
	\left( \frac{\prob{\poses^{o,R}_{k'}|\classes^{r'}_{k'},\hisnew^{r'}_{k'}}}
	{\prob{\poses^{o,R}_{k'}}} \right) \prob{\poses^{o,R}_{k}}.
\end{equation}
From that, we will address every element in the product. Poses of objects that $r'$ doesn't observe locally can be canceled out as follows, leaving only the object poses that $r'$ observed:
\begin{equation}
	\frac{\prob{\poses^{o,R}_{k'}|\classes^{r'}_{k'},\hisnew^{r'}_{k'}}}
	{\prob{\poses^{o,R}_{k'}}} = 
	\frac{\prob{\poses^{o,r'}_{k'}|\classes^{r'}_{k'},\hisnew^{r'}_{k'}}}
	{\prob{\poses^{o,r'}_{k'}}}.
\end{equation}
Then, using relation \eqref{eq:AB_Relations} again, we can expand $\prob{\poses^{o,r'}_{k'}|\classes^{r'}_{k'},\his^{r'}_{k'}}$ to separate between known prior and new measurements:
\begin{equation}\label{eq:Stack_Cont_Dev_1}
	\prob{\poses^{o,r'}_{k'}|\classes^{r'}_{k'},\his^{r'}_{k'}} =
	\prob{\poses^{o,r'}_{k'}}
	\frac{\prob{\poses^{o,r'}_{k'}|\classes^{r'}_{k'},\his^{r'}_{k'-l'}}}{\prob{\poses^{o,r'}_{k'}}}
	\frac{\prob{\poses^{o,r'}_{k'}|\classes^{r'}_{k'},\hisnew^{r'}_{k'}}}{\prob{\poses^{o,r'}_{k'}}},
\end{equation}
with $l'$ being the time difference between subsequent slots of $r$ (that can be 0 if the slot isn't updated). Then we take out all the poses that aren't dependent on the prior information, and we reach the definition of $\xi^{r,r'}_{k-1}$, i.e. the marginal object poses at the previous time.
\begin{equation}\label{eq:Stack_Cont_Dev_2}
	\frac{\prob{\poses^{o,r'}_{k'}|\classes^{r'}_{k'},\his^{r'}_{k'-l'}}}{\prob{\poses^{o,r'}_{k'}}} = 
	\frac{ \prob{\poses^{o,r'}_{k'-l'}|\classes^{r'}_{k'-l'},\his^{r'}_{k'-l'}}}
	{ \prob{\poses^{o,r'}_{k'-l'}}} \doteq \xi^{r,r'}_{k-1}.
\end{equation}
With the definition of $\xi^{r,r'}_k$, and by substituting Eq.~\eqref{eq:Stack_Cont_Dev_2} into Eq.~\eqref{eq:Stack_Cont_Dev_1} we reach the expression for a single element of the product in Eq.~\eqref{eq:Stack_Init_Cont}:
\begin{equation}\label{eq:Stack_Update_Cont}
	\frac{\prob{\poses^{o,r'}_{k'}|\classes^{r'}_{k'},\hisnew^{r'}_{k'}}}{\prob{\poses^{o,r'}_{k'}}} = 
	\frac{\xi^{r,r'}_k}{\xi^{r,r'}_{k-1}}.
\end{equation}
Finally, substituting Eq.~\eqref{eq:Stack_Update_Cont} we reach the expression for the external continuous update belief:
\begin{equation}\label{eq:Stack_Cont_Final}
	{\color{blue}\frac{\prob{\poses^{o,R}_k|\classes^R_k,\hisnew^{R-}_k}}
		{\prob{\poses^{o,R}_{k}}}} = 
	\prod_{r' \in R}
	\frac{\xi^{r,r'}_{k}}{\xi^{r,r'}_{k-1}}.
\end{equation}


\section{Derivation of {\color{red}$\frac{\prob{\classes^R_k|\hisnew^{R-}_k}}{\prob{\classes^R_k}}$}
(Discrete Belief Update)}

The discrete belief update is similar to the continuous in its derivation, with probability over class realization, rather than object poses. Again, using the relation \eqref{eq:AB_Relations} we can split the red part by separating the new measurements and actions per robot, excluding $r$ itself as it is not present in $\hisnew^{R-}_k$:
\begin{equation}\label{eq:Stack_Init_Disc}
	\prob{\classes^R_k|\hisnew^{R-}_k} =
	\prod_{k',r' \in R \backslash r}
	\left( \frac{\prob{\classes^R_{k'}|\hisnew^{r'}_{k'}}}
	{\prob{\classes^R_{k'}}} \right) \prob{\classes^R_{k'}}
\end{equation}
From that, we will address every element in the product. Classes of objects that $r'$ doesn't observe locally can be canceled out as follows, leaving only the classes of object that $r'$ observed:
\begin{equation}
	\frac{\prob{\classes^R_{k'}|\hisnew^{r'}_{k'}}}
	{\prob{\classes^R_{k'}}} = 
	\frac{\prob{\classes^{r'}_{k'}|\hisnew^{r'}_{k'}}}
	{\prob{\classes^{r'}_{k'}}}.
\end{equation}
Then, using relation \eqref{eq:AB_Relations} again, we can expand $\prob{\classes^{r'}_{k'}|\his^{r'}_{k'}}$ to separate between known prior and new measurements:
\begin{equation}\label{eq:Stack_Disc_Dev_1}
	\prob{\classes^{r'}_{k'}|\his^{r'}_{k'}} =
	\prob{\classes^{r'}_{k'}}
	\frac{\prob{\classes^{r'}_{k'}|\his^{r'}_{k'-l'}}}{\prob{\classes^{r'}_{k'}}}
	\frac{\prob{\classes^{r'}_{k'}|\hisnew^{r'}_{k'}}}{\prob{\classes^{r'}_{k'}}},
\end{equation}
with $l'$ being the time difference between subsequent slots of $r$ (that can be 0 if the slot isn't updated). Then we take out all the classes of objects that not appear in prior information, and we reach the definition of $\phi^{r,r'}_{k-1}$, i.e. the marginal object poses at the previous time.
\begin{equation}\label{eq:Stack_Disc_Dev_2}
\frac{\prob{\classes^{r'}_{k'}|\his^{r'}_{k'-l'}}}{\prob{\classes^{r'}_{k'}}} = 
\frac{ \prob{\classes^{r'}_{k'-l'}|\his^{r'}_{k'-l'}}}
{ \prob{\classes^{r'}_{k'-l'}}} \doteq \phi^{r,r'}_{k-1}
\end{equation}
With the definition of $\phi^{r,r'}_k$, and by substituting Eq.~\eqref{eq:Stack_Disc_Dev_2} into Eq.~\eqref{eq:Stack_Disc_Dev_1} we reach the expression for a single element of the product in Eq.~\eqref{eq:Stack_Init_Disc}:
\begin{equation}\label{eq:Stack_Update_Disc}
\frac{\prob{\classes^{r'}_{k'}|\hisnew^{r'}_{k'}}}{\prob{\classes^{r'}_{k'}}} = 
\frac{\phi^{r,r'}_k}{\phi^{r,r'}_{k-1}}
\end{equation}
Finally, substituting Eq.~\eqref{eq:Stack_Update_Disc} we reach the expression for the external continuous update belief:
\begin{equation}\label{eq:Stack_Disc_Final}
{\color{red}\frac{\prob{\classes^R_k|\hisnew^{R-}_k}}{\prob{\classes^R_k}}} = 
\prod_{r' \in R}
\frac{\phi^{r,r'}_{k}}{\phi^{r,r'}_{k-1}}.
\end{equation}


\section{Simulation: Parameters}

We consider a motion model with noise covariance $\Sigma_w = diag(0.003,0.003, 0.001)$, and geometric model with noise  covariance $\Sigma^{geo}_v=diag(0.1,0.1,0.01)$, both corresponding to position coordinates in meters and orientation in radians.

Our semantic model parameters are defined as:

\textsc{\begin{eqnarray*}
	h_c(c=1,\psi) \!\!&\doteq& \!\! \left[   0.25 \cdot \sin(\psi) + 0.75 ,
	0.25 ( 1 - \sin(\psi) ) \right]^T
	\\
	h_c(c=2,\psi) \!\! &\doteq& \!\! \left[   0.25 ( 1 - \sin(\psi) ) , 
	0.25 \cdot \sin(\psi) + 0.75  \right]^T,
\end{eqnarray*}}

where $h_c(c=i,\psi) \in \mathbb{R}^M$ is the predicted probability vector given object class $c$ is $i$. Recall that  our semantic measurements $z^{sem,r}_k$  are probability vectors as well. $\psi$ is the relative orientation between robot and object, computed from the relative pose $x^{rel}_k \doteq x^{o} \ominus x_k$. The measurement covariance is defined via the square root information matrix, such that $\Sigma_c \doteq (R^T R)^{-1}$, and $R = \left[ \begin{array}{cc} 1.5 & -0.75 \\ 0 & 1.5 \end{array} \right]$. Both the geometric and semantic measurements are limited to 10 meters from the robot's pose. The highest probability for ambiguous class measurements is at $\psi = -90^\circ$, where $h_c = [0.5, 0.5]^T$ for both classes.


\newpage
\section{Simulation: Table of Stack Time Stamps}

In this section we present a table of stack time stamps that indicates direct and indirect communication between robots in our scenario. Recall that the maximal communication radius is 10 meters, thus robots $r_2$ and $r_3$ communicate from time $k=6$, robots $r_1$ starts communicating to others from time $k=13$.

\begin{multicols}{2}
\begin{tabular}{|l|l|l|l|}
	\hline
	Time step & Stack of $r_1$ & Stack of $r_2$ & Stack of $r_3$ \\
	\hline

	 $k = 1$ &
	\makecell{t($r_1$): 1 \\ 	t($r_2$): 0 \\ 	t($r_3$): 0} &
	\makecell{t($r_1$): 0 \\ 	t($r_2$): 0 \\ 	t($r_3$): 0} &
	\makecell{t($r_1$): 0 \\ 	t($r_2$): 0 \\ 	t($r_3$): 1} \\ 
	\hline 
	
	$k = 2$ &
	\makecell{t($r_1$): 2 \\ 	t($r_2$): 0 \\ 	t($r_3$): 0} &
	\makecell{t($r_1$): 0 \\ 	t($r_2$): 0 \\ 	t($r_3$): 0} &
	\makecell{t($r_1$): 0 \\ 	t($r_2$): 0 \\ 	t($r_3$): 2} \\ 
	\hline 
	
	$k = 3$ &
	\makecell{t($r_1$): 3 \\ 	t($r_2$): 0 \\ 	t($r_3$): 0} &
	\makecell{t($r_1$): 0 \\ 	t($r_2$): 3 \\ 	t($r_3$): 0} &
	\makecell{t($r_1$): 0 \\ 	t($r_2$): 0 \\ 	t($r_3$): 3} \\ 
	\hline 
	
	$k = 4$ &
	\makecell{t($r_1$): 4 \\ 	t($r_2$): 0 \\ 	t($r_3$): 0} &
	\makecell{t($r_1$): 0 \\ 	t($r_2$): 4 \\ 	t($r_3$): 0} &
	\makecell{t($r_1$): 0 \\ 	t($r_2$): 0 \\ 	t($r_3$): 4} \\ 
	\hline 
	
	$k = 5$ &
	\makecell{t($r_1$): 5 \\ 	t($r_2$): 0 \\ 	t($r_3$): 0} &
	\makecell{t($r_1$): 0 \\ 	t($r_2$): 5 \\ 	t($r_3$): 0} &
	\makecell{t($r_1$): 0 \\ 	t($r_2$): 0 \\ 	t($r_3$): 5} \\ 
	\hline 
	
	$k = 6$ &
	\makecell{t($r_1$): 6 \\ 	t($r_2$): 0 \\ 	t($r_3$): 0} &
	\makecell{t($r_1$): 0 \\ 	t($r_2$): 6 \\ 	t($r_3$): 5} &
	\makecell{t($r_1$): 0 \\ 	t($r_2$): 5 \\ 	t($r_3$): 6} \\ 
	\hline 
	
	$k = 7$ &
	\makecell{t($r_1$): 7 \\ 	t($r_2$): 0 \\ 	t($r_3$): 0} &
	\makecell{t($r_1$): 0 \\ 	t($r_2$): 7 \\ 	t($r_3$): 6} &
	\makecell{t($r_1$): 0 \\ 	t($r_2$): 6 \\ 	t($r_3$): 7} \\ 
	\hline 
	
	$k = 8$ &
	\makecell{t($r_1$): 8 \\ 	t($r_2$): 0 \\ 	t($r_3$): 0} &
	\makecell{t($r_1$): 0 \\ 	t($r_2$): 8 \\ 	t($r_3$): 7} &
	\makecell{t($r_1$): 0 \\ 	t($r_2$): 7 \\ 	t($r_3$): 8} \\ 
	\hline 
	
	$k = 9$ &
	\makecell{t($r_1$): 9 \\ 	t($r_2$): 0 \\ 	t($r_3$): 0} &
	\makecell{t($r_1$): 0 \\ 	t($r_2$): 9 \\ 	t($r_3$): 8} &
	\makecell{t($r_1$): 0 \\ 	t($r_2$): 8 \\ 	t($r_3$): 9} \\ 
	\hline 
	
	$k = 10$ &
	\makecell{t($r_1$): 10 \\ 	t($r_2$): 0 \\ 	t($r_3$): 0} &
	\makecell{t($r_1$): 0 \\ 	t($r_2$): 10 \\ 	t($r_3$): 9} &
	\makecell{t($r_1$): 0 \\ 	t($r_2$): 9 \\ 	t($r_3$): 10} \\ 
	\hline 
	
	$k = 11$ &
	\makecell{t($r_1$): 11 \\ 	t($r_2$): 0 \\ 	t($r_3$): 0} &
	\makecell{t($r_1$): 0 \\ 	t($r_2$): 11 \\ 	t($r_3$): 10} &
	\makecell{t($r_1$): 0 \\ 	t($r_2$): 10 \\ 	t($r_3$): 11} \\ 
	\hline 
	
	$k = 12$ &
	\makecell{t($r_1$): 12 \\ 	t($r_2$): 0 \\ 	t($r_3$): 0} &
	\makecell{t($r_1$): 0 \\ 	t($r_2$): 12 \\ 	t($r_3$): 11} &
	\makecell{t($r_1$): 0 \\ 	t($r_2$): 11 \\ 	t($r_3$): 12} \\ 
	\hline 
	
	$k = 13$ &
	\makecell{t($r_1$): 13 \\ 	t($r_2$): 12 \\ 	t($r_3$): 12} &
	\makecell{t($r_1$): 12 \\ 	t($r_2$): 13 \\ 	t($r_3$): 12} &
	\makecell{t($r_1$): 12 \\ 	t($r_2$): 12 \\ 	t($r_3$): 13} \\ 
	\hline 
	
	$k = 14$ &
	\makecell{t($r_1$): 14 \\ 	t($r_2$): 13 \\ 	t($r_3$): 13} &
	\makecell{t($r_1$): 13 \\ 	t($r_2$): 14 \\ 	t($r_3$): 13} &
	\makecell{t($r_1$): 13 \\ 	t($r_2$): 13 \\ 	t($r_3$): 14} \\ 
	\hline 
	
	$k = 15$ &
	\makecell{t($r_1$): 15 \\ 	t($r_2$): 14 \\ 	t($r_3$): 14} &
	\makecell{t($r_1$): 14 \\ 	t($r_2$): 15 \\ 	t($r_3$): 14} &
	\makecell{t($r_1$): 14 \\ 	t($r_2$): 14 \\ 	t($r_3$): 15} \\ 
	\hline 
	
\end{tabular}

\begin{tabular}{|l|l|l|l|}
	\hline
	Time step & Stack of $r_1$ & Stack of $r_2$ & Stack of $r_3$ \\
	\hline
	
	 $k = 16$ &
	\makecell{t($r_1$): 16 \\ 	t($r_2$): 15 \\ 	t($r_3$): 15} &
	\makecell{t($r_1$): 15 \\ 	t($r_2$): 16 \\ 	t($r_3$): 15} &
	\makecell{t($r_1$): 15 \\ 	t($r_2$): 15 \\ 	t($r_3$): 16} \\ 
	\hline 
	
	$k = 17$ &
	\makecell{t($r_1$): 17 \\ 	t($r_2$): 16 \\ 	t($r_3$): 16} &
	\makecell{t($r_1$): 16 \\ 	t($r_2$): 17 \\ 	t($r_3$): 16} &
	\makecell{t($r_1$): 16 \\ 	t($r_2$): 16 \\ 	t($r_3$): 17} \\ 
	\hline 
	
	$k = 18$ &
	\makecell{t($r_1$): 18 \\ 	t($r_2$): 17 \\ 	t($r_3$): 17} &
	\makecell{t($r_1$): 17 \\ 	t($r_2$): 18 \\ 	t($r_3$): 17} &
	\makecell{t($r_1$): 17 \\ 	t($r_2$): 17 \\ 	t($r_3$): 18} \\ 
	\hline 
	
	$k = 19$ &
	\makecell{t($r_1$): 19 \\ 	t($r_2$): 18 \\ 	t($r_3$): 18} &
	\makecell{t($r_1$): 18 \\ 	t($r_2$): 19 \\ 	t($r_3$): 18} &
	\makecell{t($r_1$): 18 \\ 	t($r_2$): 18 \\ 	t($r_3$): 19} \\ 
	\hline 
	
	$k = 20$ &
	\makecell{t($r_1$): 20 \\ 	t($r_2$): 19 \\ 	t($r_3$): 19} &
	\makecell{t($r_1$): 19 \\ 	t($r_2$): 20 \\ 	t($r_3$): 19} &
	\makecell{t($r_1$): 19 \\ 	t($r_2$): 19 \\ 	t($r_3$): 20} \\ 
	\hline 
	
	$k = 21$ &
	\makecell{t($r_1$): 21 \\ 	t($r_2$): 20 \\ 	t($r_3$): 20} &
	\makecell{t($r_1$): 20 \\ 	t($r_2$): 21 \\ 	t($r_3$): 20} &
	\makecell{t($r_1$): 20 \\ 	t($r_2$): 20 \\ 	t($r_3$): 21} \\ 
	\hline 
	
	$k = 22$ &
	\makecell{t($r_1$): 22 \\ 	t($r_2$): 21 \\ 	t($r_3$): 21} &
	\makecell{t($r_1$): 21 \\ 	t($r_2$): 22 \\ 	t($r_3$): 21} &
	\makecell{t($r_1$): 21 \\ 	t($r_2$): 21 \\ 	t($r_3$): 22} \\ 
	\hline 
	
	$k = 23$ &
	\makecell{t($r_1$): 23 \\ 	t($r_2$): 22 \\ 	t($r_3$): 22} &
	\makecell{t($r_1$): 22 \\ 	t($r_2$): 23 \\ 	t($r_3$): 22} &
	\makecell{t($r_1$): 22 \\ 	t($r_2$): 22 \\ 	t($r_3$): 23} \\ 
	\hline 
	
	$k = 24$ &
	\makecell{t($r_1$): 24 \\ 	t($r_2$): 23 \\ 	t($r_3$): 23} &
	\makecell{t($r_1$): 23 \\ 	t($r_2$): 24 \\ 	t($r_3$): 23} &
	\makecell{t($r_1$): 23 \\ 	t($r_2$): 23 \\ 	t($r_3$): 24} \\ 
	\hline 
	
	$k = 25$ &
	\makecell{t($r_1$): 25 \\ 	t($r_2$): 24 \\ 	t($r_3$): 24} &
	\makecell{t($r_1$): 24 \\ 	t($r_2$): 25 \\ 	t($r_3$): 24} &
	\makecell{t($r_1$): 24 \\ 	t($r_2$): 24 \\ 	t($r_3$): 25} \\ 
	\hline 
	
	$k = 26$ &
	\makecell{t($r_1$): 26 \\ 	t($r_2$): 25 \\ 	t($r_3$): 25} &
	\makecell{t($r_1$): 25 \\ 	t($r_2$): 26 \\ 	t($r_3$): 25} &
	\makecell{t($r_1$): 25 \\ 	t($r_2$): 25 \\ 	t($r_3$): 26} \\ 
	\hline 
	
	$k = 27$ &
	\makecell{t($r_1$): 27 \\ 	t($r_2$): 26 \\ 	t($r_3$): 26} &
	\makecell{t($r_1$): 26 \\ 	t($r_2$): 27 \\ 	t($r_3$): 26} &
	\makecell{t($r_1$): 26 \\ 	t($r_2$): 26 \\ 	t($r_3$): 27} \\ 
	\hline 
	
	$k = 28$ &
	\makecell{t($r_1$): 28 \\ 	t($r_2$): 27 \\ 	t($r_3$): 27} &
	\makecell{t($r_1$): 27 \\ 	t($r_2$): 28 \\ 	t($r_3$): 27} &
	\makecell{t($r_1$): 27 \\ 	t($r_2$): 27 \\ 	t($r_3$): 28} \\ 
	\hline 
	
	$k = 29$ &
	\makecell{t($r_1$): 29 \\ 	t($r_2$): 28 \\ 	t($r_3$): 28} &
	\makecell{t($r_1$): 28 \\ 	t($r_2$): 29 \\ 	t($r_3$): 28} &
	\makecell{t($r_1$): 28 \\ 	t($r_2$): 28 \\ 	t($r_3$): 29} \\ 
	\hline 
	
	$k = 30$ &
	\makecell{t($r_1$): 30 \\ 	t($r_2$): 29 \\ 	t($r_3$): 29} &
	\makecell{t($r_1$): 29 \\ 	t($r_2$): 30 \\ 	t($r_3$): 29} &
	\makecell{t($r_1$): 29 \\ 	t($r_2$): 29 \\ 	t($r_3$): 30} \\ 
	\hline 
\end{tabular}
\end{multicols}

\newpage

\begin{multicols}{2}
	\begin{tabular}{|l|l|l|l|}
		\hline
		Time step & Stack of $r_1$ & Stack of $r_2$ & Stack of $r_3$ \\
		\hline

		$k = 31$ &
		\makecell{t($r_1$): 31 \\ 	t($r_2$): 30 \\ 	t($r_3$): 30} &
		\makecell{t($r_1$): 30 \\ 	t($r_2$): 31 \\ 	t($r_3$): 30} &
		\makecell{t($r_1$): 30 \\ 	t($r_2$): 30 \\ 	t($r_3$): 31} \\ 
		\hline 
		
		$k = 32$ &
		\makecell{t($r_1$): 32 \\ 	t($r_2$): 31 \\ 	t($r_3$): 31} &
		\makecell{t($r_1$): 31 \\ 	t($r_2$): 32 \\ 	t($r_3$): 31} &
		\makecell{t($r_1$): 31 \\ 	t($r_2$): 31 \\ 	t($r_3$): 32} \\ 
		\hline 
		
		$k = 33$ &
		\makecell{t($r_1$): 33 \\ 	t($r_2$): 32 \\ 	t($r_3$): 32} &
		\makecell{t($r_1$): 32 \\ 	t($r_2$): 33 \\ 	t($r_3$): 32} &
		\makecell{t($r_1$): 32 \\ 	t($r_2$): 32 \\ 	t($r_3$): 33} \\ 
		\hline 
		
		$k = 34$ &
		\makecell{t($r_1$): 34 \\ 	t($r_2$): 33 \\ 	t($r_3$): 33} &
		\makecell{t($r_1$): 33 \\ 	t($r_2$): 34 \\ 	t($r_3$): 33} &
		\makecell{t($r_1$): 33 \\ 	t($r_2$): 33 \\ 	t($r_3$): 34} \\ 
		\hline 
		
		$k = 35$ &
		\makecell{t($r_1$): 35 \\ 	t($r_2$): 34 \\ 	t($r_3$): 34} &
		\makecell{t($r_1$): 34 \\ 	t($r_2$): 35 \\ 	t($r_3$): 34} &
		\makecell{t($r_1$): 34 \\ 	t($r_2$): 34 \\ 	t($r_3$): 35} \\ 
		\hline 
		
		$k = 36$ &
		\makecell{t($r_1$): 36 \\ 	t($r_2$): 35 \\ 	t($r_3$): 35} &
		\makecell{t($r_1$): 35 \\ 	t($r_2$): 36 \\ 	t($r_3$): 35} &
		\makecell{t($r_1$): 35 \\ 	t($r_2$): 35 \\ 	t($r_3$): 36} \\ 
		\hline 
		
		$k = 37$ &
		\makecell{t($r_1$): 37 \\ 	t($r_2$): 36 \\ 	t($r_3$): 36} &
		\makecell{t($r_1$): 36 \\ 	t($r_2$): 37 \\ 	t($r_3$): 36} &
		\makecell{t($r_1$): 36 \\ 	t($r_2$): 36 \\ 	t($r_3$): 37} \\ 
		\hline 
		
		$k = 38$ &
		\makecell{t($r_1$): 38 \\ 	t($r_2$): 37 \\ 	t($r_3$): 37} &
		\makecell{t($r_1$): 37 \\ 	t($r_2$): 38 \\ 	t($r_3$): 37} &
		\makecell{t($r_1$): 37 \\ 	t($r_2$): 37 \\ 	t($r_3$): 38} \\ 
		\hline 
		
		$k = 39$ &
		\makecell{t($r_1$): 39 \\ 	t($r_2$): 38 \\ 	t($r_3$): 38} &
		\makecell{t($r_1$): 38 \\ 	t($r_2$): 39 \\ 	t($r_3$): 38} &
		\makecell{t($r_1$): 38 \\ 	t($r_2$): 38 \\ 	t($r_3$): 39} \\ 
		\hline 
		
		$k = 40$ &
		\makecell{t($r_1$): 40 \\ 	t($r_2$): 39 \\ 	t($r_3$): 39} &
		\makecell{t($r_1$): 39 \\ 	t($r_2$): 40 \\ 	t($r_3$): 39} &
		\makecell{t($r_1$): 39 \\ 	t($r_2$): 39 \\ 	t($r_3$): 40} \\ 
		\hline 
		
		$k = 41$ &
		\makecell{t($r_1$): 41 \\ 	t($r_2$): 40 \\ 	t($r_3$): 40} &
		\makecell{t($r_1$): 40 \\ 	t($r_2$): 41 \\ 	t($r_3$): 40} &
		\makecell{t($r_1$): 40 \\ 	t($r_2$): 40 \\ 	t($r_3$): 41} \\ 
		\hline 
		
		$k = 42$ &
		\makecell{t($r_1$): 42 \\ 	t($r_2$): 41 \\ 	t($r_3$): 41} &
		\makecell{t($r_1$): 41 \\ 	t($r_2$): 42 \\ 	t($r_3$): 41} &
		\makecell{t($r_1$): 41 \\ 	t($r_2$): 41 \\ 	t($r_3$): 42} \\ 
		\hline 
		
		$k = 43$ &
		\makecell{t($r_1$): 43 \\ 	t($r_2$): 42 \\ 	t($r_3$): 42} &
		\makecell{t($r_1$): 42 \\ 	t($r_2$): 43 \\ 	t($r_3$): 42} &
		\makecell{t($r_1$): 42 \\ 	t($r_2$): 42 \\ 	t($r_3$): 43} \\ 
		\hline 
		
		$k = 44$ &
		\makecell{t($r_1$): 44 \\ 	t($r_2$): 43 \\ 	t($r_3$): 43} &
		\makecell{t($r_1$): 43 \\ 	t($r_2$): 44 \\ 	t($r_3$): 43} &
		\makecell{t($r_1$): 43 \\ 	t($r_2$): 43 \\ 	t($r_3$): 44} \\ 
		\hline 
		
		$k = 45$ &
		\makecell{t($r_1$): 45 \\ 	t($r_2$): 44 \\ 	t($r_3$): 44} &
		\makecell{t($r_1$): 44 \\ 	t($r_2$): 45 \\ 	t($r_3$): 44} &
		\makecell{t($r_1$): 44 \\ 	t($r_2$): 44 \\ 	t($r_3$): 45} \\ 
		\hline 
		
	\end{tabular}

	\begin{tabular}{|l|l|l|l|}
		\hline
		Time step & Stack of $r_1$ & Stack of $r_2$ & Stack of $r_3$ \\
		\hline
		
		 $k = 46$ &
		\makecell{t($r_1$): 46 \\ 	t($r_2$): 45 \\ 	t($r_3$): 45} &
		\makecell{t($r_1$): 45 \\ 	t($r_2$): 46 \\ 	t($r_3$): 45} &
		\makecell{t($r_1$): 45 \\ 	t($r_2$): 45 \\ 	t($r_3$): 46} \\ 
		\hline 
		
		$k = 47$ &
		\makecell{t($r_1$): 47 \\ 	t($r_2$): 46 \\ 	t($r_3$): 46} &
		\makecell{t($r_1$): 46 \\ 	t($r_2$): 47 \\ 	t($r_3$): 46} &
		\makecell{t($r_1$): 46 \\ 	t($r_2$): 46 \\ 	t($r_3$): 47} \\ 
		\hline 
		
		$k = 48$ &
		\makecell{t($r_1$): 48 \\ 	t($r_2$): 47 \\ 	t($r_3$): 47} &
		\makecell{t($r_1$): 47 \\ 	t($r_2$): 48 \\ 	t($r_3$): 47} &
		\makecell{t($r_1$): 47 \\ 	t($r_2$): 47 \\ 	t($r_3$): 48} \\ 
		\hline 
		
		$k = 49$ &
		\makecell{t($r_1$): 49 \\ 	t($r_2$): 48 \\ 	t($r_3$): 48} &
		\makecell{t($r_1$): 48 \\ 	t($r_2$): 49 \\ 	t($r_3$): 48} &
		\makecell{t($r_1$): 48 \\ 	t($r_2$): 48 \\ 	t($r_3$): 49} \\ 
		\hline 
		
		$k = 50$ &
		\makecell{t($r_1$): 50 \\ 	t($r_2$): 49 \\ 	t($r_3$): 49} &
		\makecell{t($r_1$): 49 \\ 	t($r_2$): 50 \\ 	t($r_3$): 49} &
		\makecell{t($r_1$): 49 \\ 	t($r_2$): 49 \\ 	t($r_3$): 50} \\ 
		\hline 
		
		$k = 51$ &
		\makecell{t($r_1$): 51 \\ 	t($r_2$): 50 \\ 	t($r_3$): 50} &
		\makecell{t($r_1$): 50 \\ 	t($r_2$): 51 \\ 	t($r_3$): 50} &
		\makecell{t($r_1$): 50 \\ 	t($r_2$): 50 \\ 	t($r_3$): 51} \\ 
		\hline 
		
		$k = 52$ &
		\makecell{t($r_1$): 52 \\ 	t($r_2$): 51 \\ 	t($r_3$): 51} &
		\makecell{t($r_1$): 51 \\ 	t($r_2$): 52 \\ 	t($r_3$): 51} &
		\makecell{t($r_1$): 51 \\ 	t($r_2$): 51 \\ 	t($r_3$): 52} \\ 
		\hline 
		
		$k = 53$ &
		\makecell{t($r_1$): 53 \\ 	t($r_2$): 52 \\ 	t($r_3$): 52} &
		\makecell{t($r_1$): 52 \\ 	t($r_2$): 53 \\ 	t($r_3$): 52} &
		\makecell{t($r_1$): 52 \\ 	t($r_2$): 52 \\ 	t($r_3$): 53} \\ 
		\hline 
		
		$k = 54$ &
		\makecell{t($r_1$): 54 \\ 	t($r_2$): 53 \\ 	t($r_3$): 53} &
		\makecell{t($r_1$): 53 \\ 	t($r_2$): 54 \\ 	t($r_3$): 53} &
		\makecell{t($r_1$): 53 \\ 	t($r_2$): 53 \\ 	t($r_3$): 54} \\ 
		\hline 
		
		$k = 55$ &
		\makecell{t($r_1$): 55 \\ 	t($r_2$): 54 \\ 	t($r_3$): 54} &
		\makecell{t($r_1$): 54 \\ 	t($r_2$): 55 \\ 	t($r_3$): 54} &
		\makecell{t($r_1$): 54 \\ 	t($r_2$): 54 \\ 	t($r_3$): 55} \\ 
		\hline 
		
		$k = 56$ &
		\makecell{t($r_1$): 56 \\ 	t($r_2$): 55 \\ 	t($r_3$): 55} &
		\makecell{t($r_1$): 55 \\ 	t($r_2$): 56 \\ 	t($r_3$): 55} &
		\makecell{t($r_1$): 55 \\ 	t($r_2$): 55 \\ 	t($r_3$): 56} \\ 
		\hline 
		
		$k = 57$ &
		\makecell{t($r_1$): 57 \\ 	t($r_2$): 56 \\ 	t($r_3$): 56} &
		\makecell{t($r_1$): 56 \\ 	t($r_2$): 57 \\ 	t($r_3$): 56} &
		\makecell{t($r_1$): 56 \\ 	t($r_2$): 56 \\ 	t($r_3$): 57} \\ 
		\hline 
		
		$k = 58$ &
		\makecell{t($r_1$): 58 \\ 	t($r_2$): 57 \\ 	t($r_3$): 57} &
		\makecell{t($r_1$): 57 \\ 	t($r_2$): 58 \\ 	t($r_3$): 57} &
		\makecell{t($r_1$): 57 \\ 	t($r_2$): 57 \\ 	t($r_3$): 58} \\ 
		\hline 
		
		$k = 59$ &
		\makecell{t($r_1$): 59 \\ 	t($r_2$): 58 \\ 	t($r_3$): 58} &
		\makecell{t($r_1$): 58 \\ 	t($r_2$): 59 \\ 	t($r_3$): 58} &
		\makecell{t($r_1$): 58 \\ 	t($r_2$): 58 \\ 	t($r_3$): 59} \\ 
		\hline 
		
		$k = 60$ &
		\makecell{t($r_1$): 60 \\ 	t($r_2$): 59 \\ 	t($r_3$): 59} &
		\makecell{t($r_1$): 59 \\ 	t($r_2$): 60 \\ 	t($r_3$): 59} &
		\makecell{t($r_1$): 59 \\ 	t($r_2$): 59 \\ 	t($r_3$): 60} \\ 
		\hline 
	\end{tabular}
\end{multicols}

\clearpage
\section{Simulation: Additional Results}

\begin{figure}[!htbp]
	
	\begin{subfigure}[b]{0.24\textwidth}
		\includegraphics[width=\textwidth]{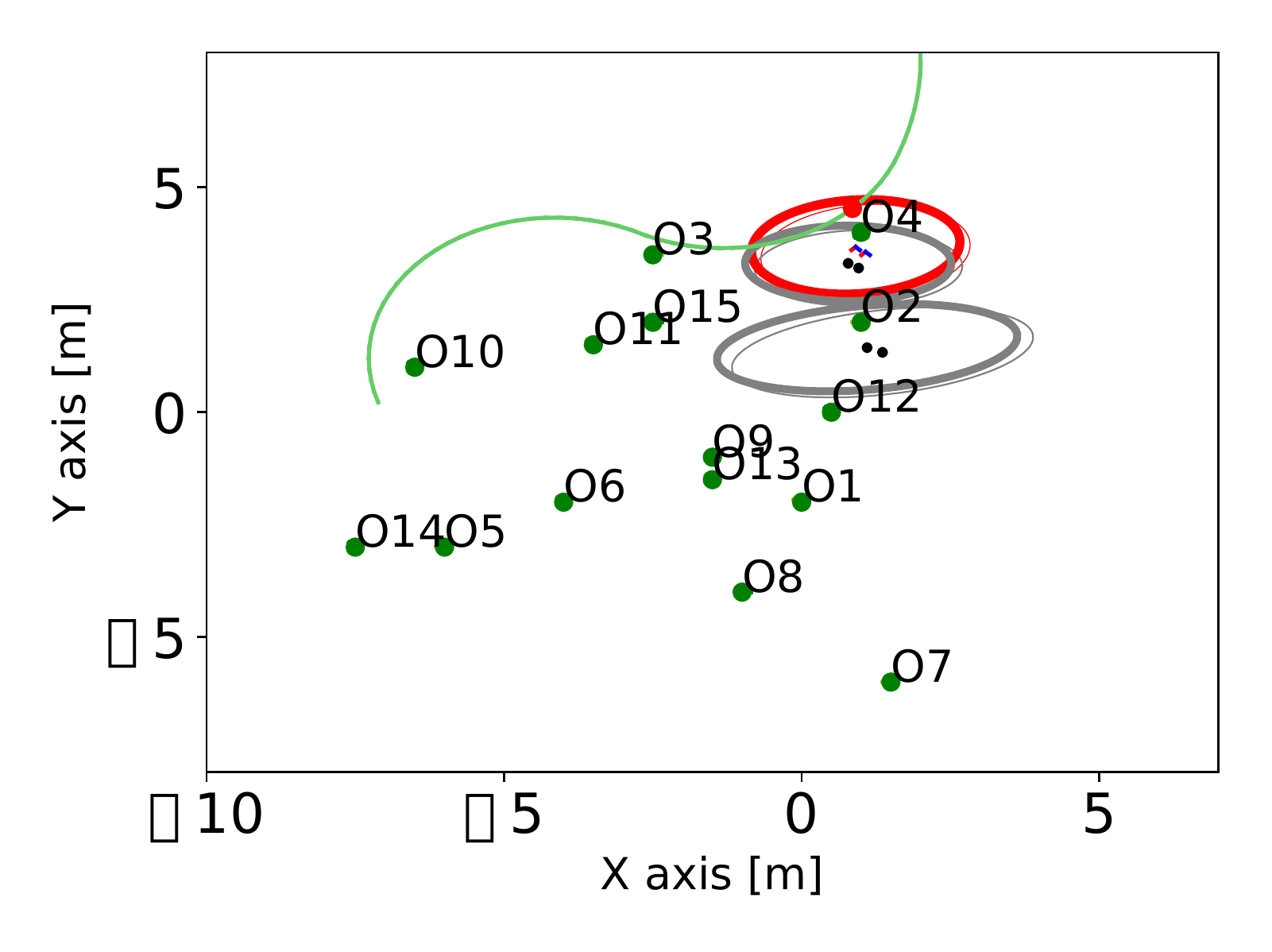}
		\caption{Local SLAM $r_2$}\label{fig:Robot_2_x15}
	\end{subfigure}
	\begin{subfigure}[b]{0.24\textwidth}
		\includegraphics[width=\textwidth]{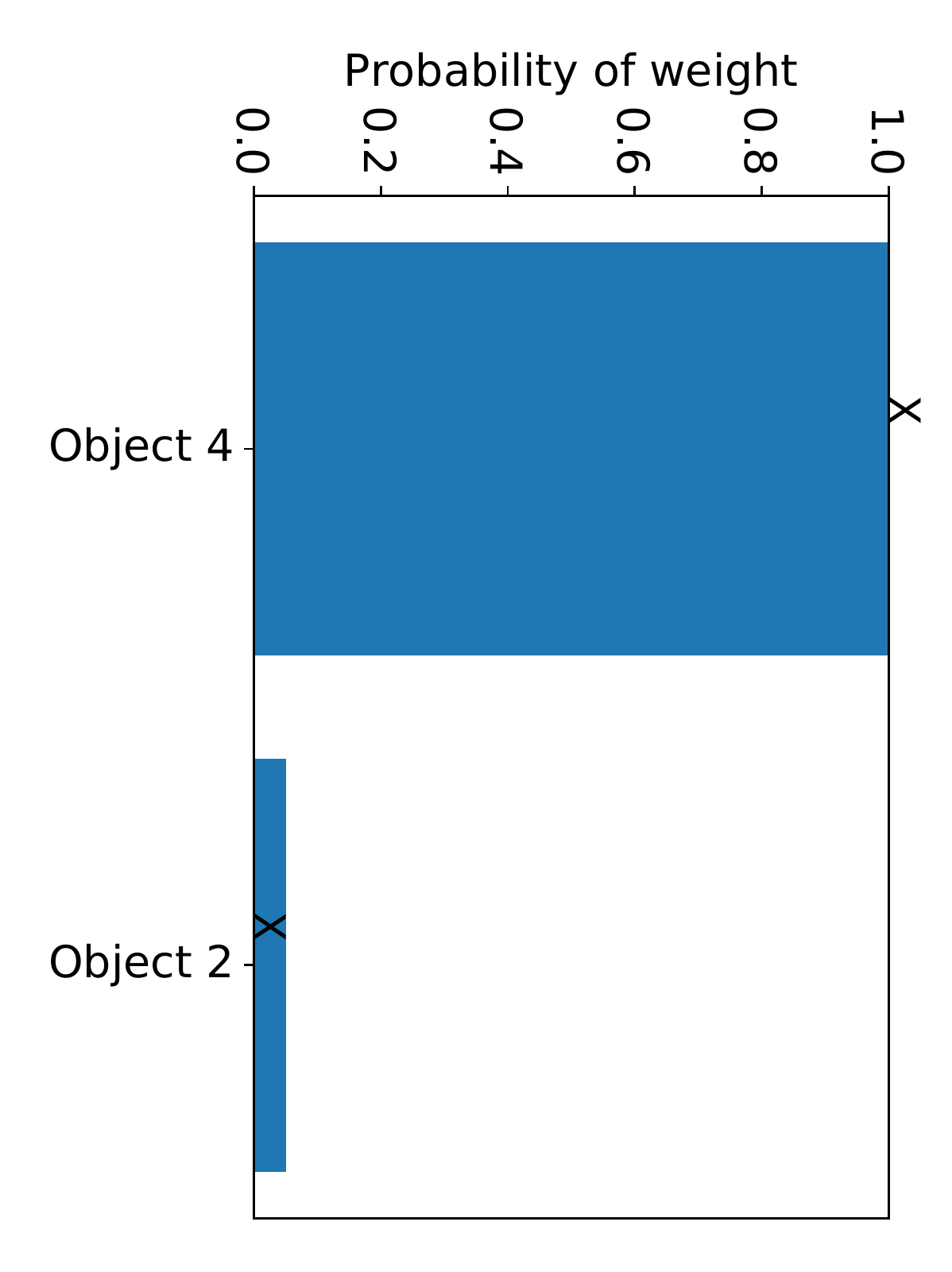}
		\caption{Local classification $r_2$}\label{fig:Robot_2_x15_cls}
	\end{subfigure}
	\begin{subfigure}[b]{0.24\textwidth}
		\includegraphics[width=\textwidth]{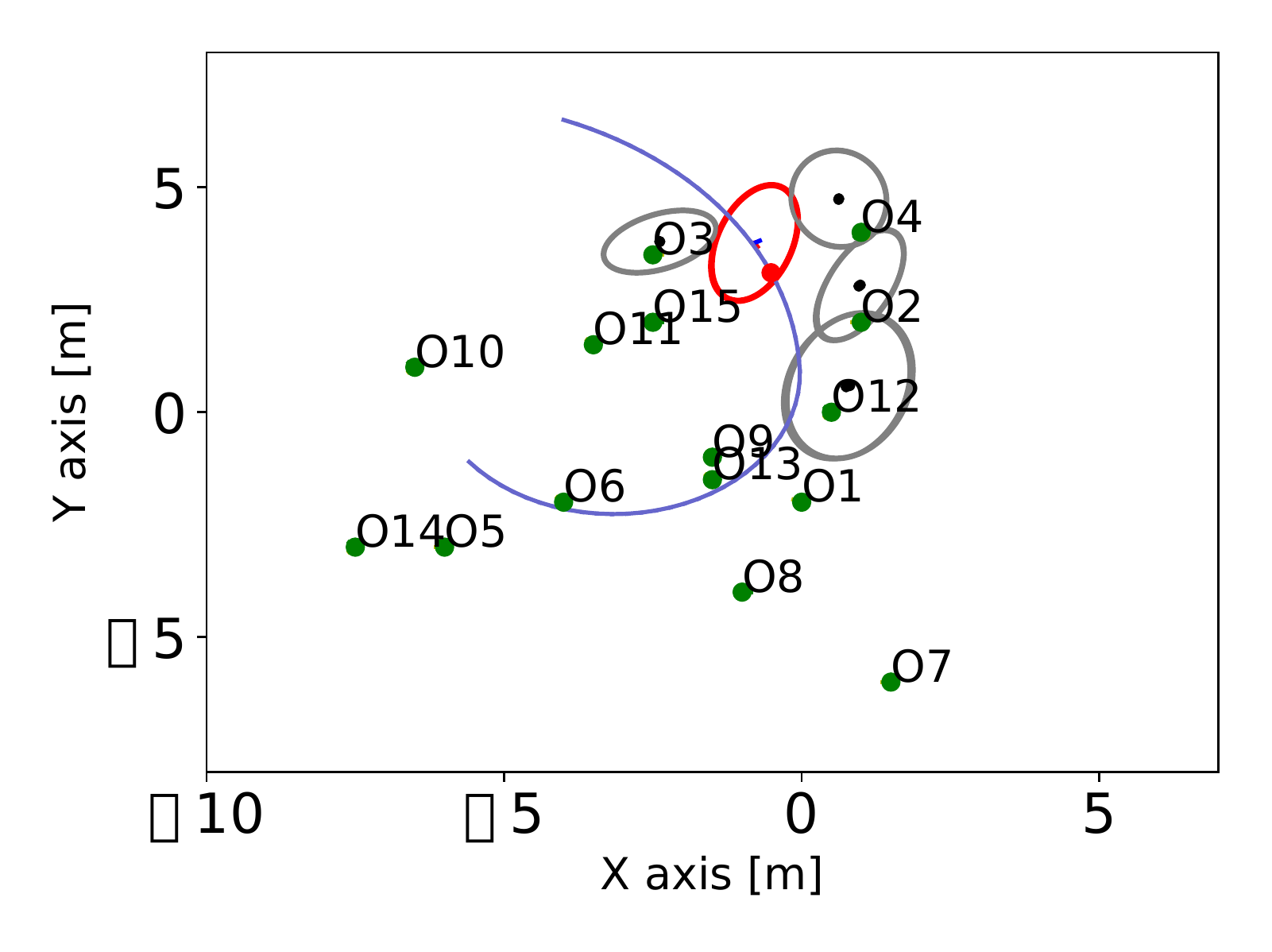}
		\caption{Local SLAM $r_3$}\label{fig:Robot_3_x20}
	\end{subfigure}
	\begin{subfigure}[b]{0.24\textwidth}
		\includegraphics[width=\textwidth]{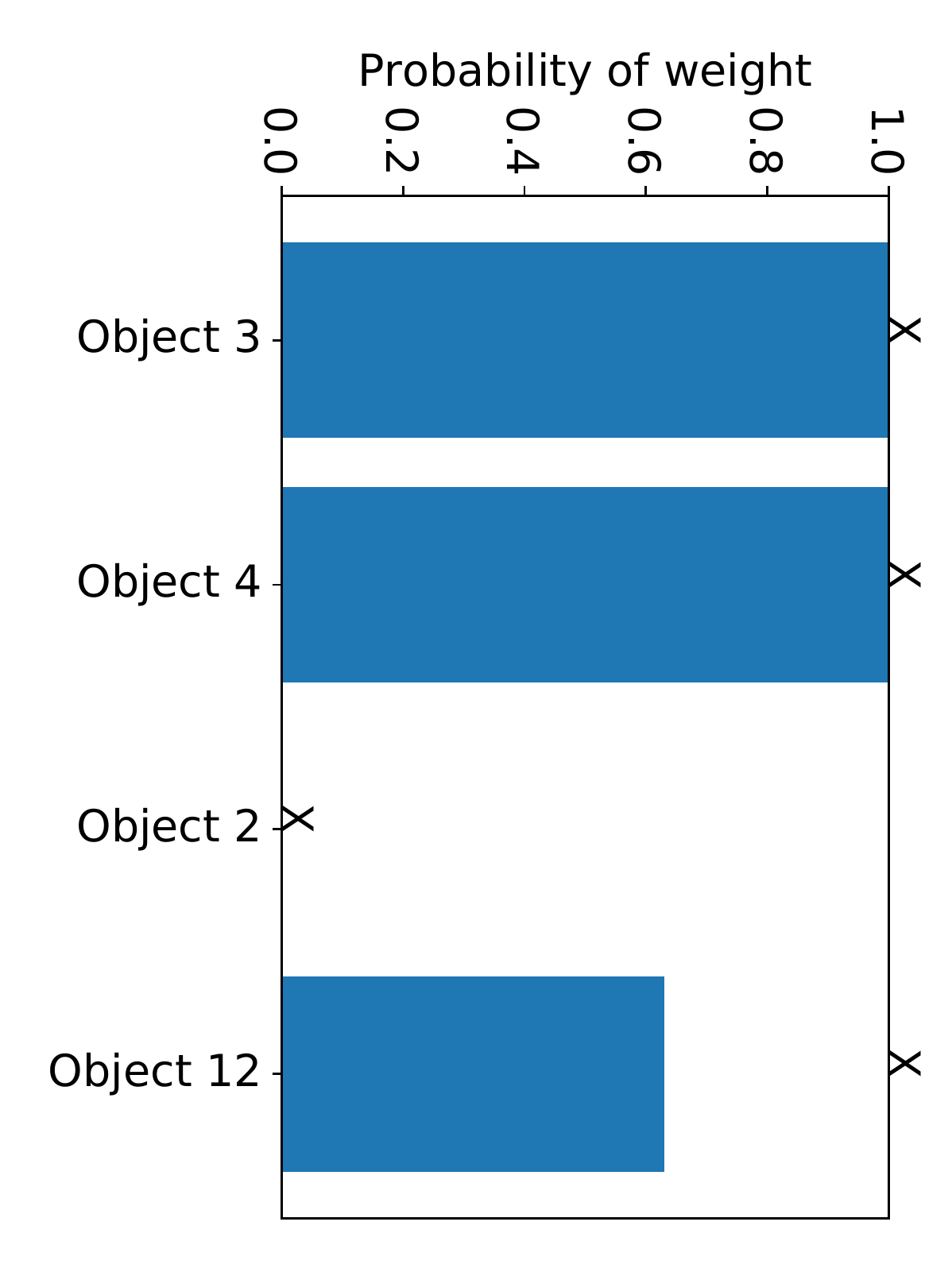}
		\caption{Local classification $r_3$}\label{fig:Robot_3_x20_cls}
	\end{subfigure}
	
	\caption{Figures for robot $r_2$ and $r_3$, local beliefs for time $k=15$ and $k=20$ respectively. \textbf{(a)} and \textbf{(b)} show results for $r_2$, \textbf{(c)} and \textbf{(d)} for $r_3$. \textbf{(a)} and \textbf{(c)} present SLAM results, \textbf{(b)} and \textbf{(d)} present classification results.}
	\label{fig:Figures_x4_local}
\end{figure}

\begin{figure}[!htbp]
	
	\begin{subfigure}[b]{0.24\textwidth}
		\includegraphics[width=\textwidth]{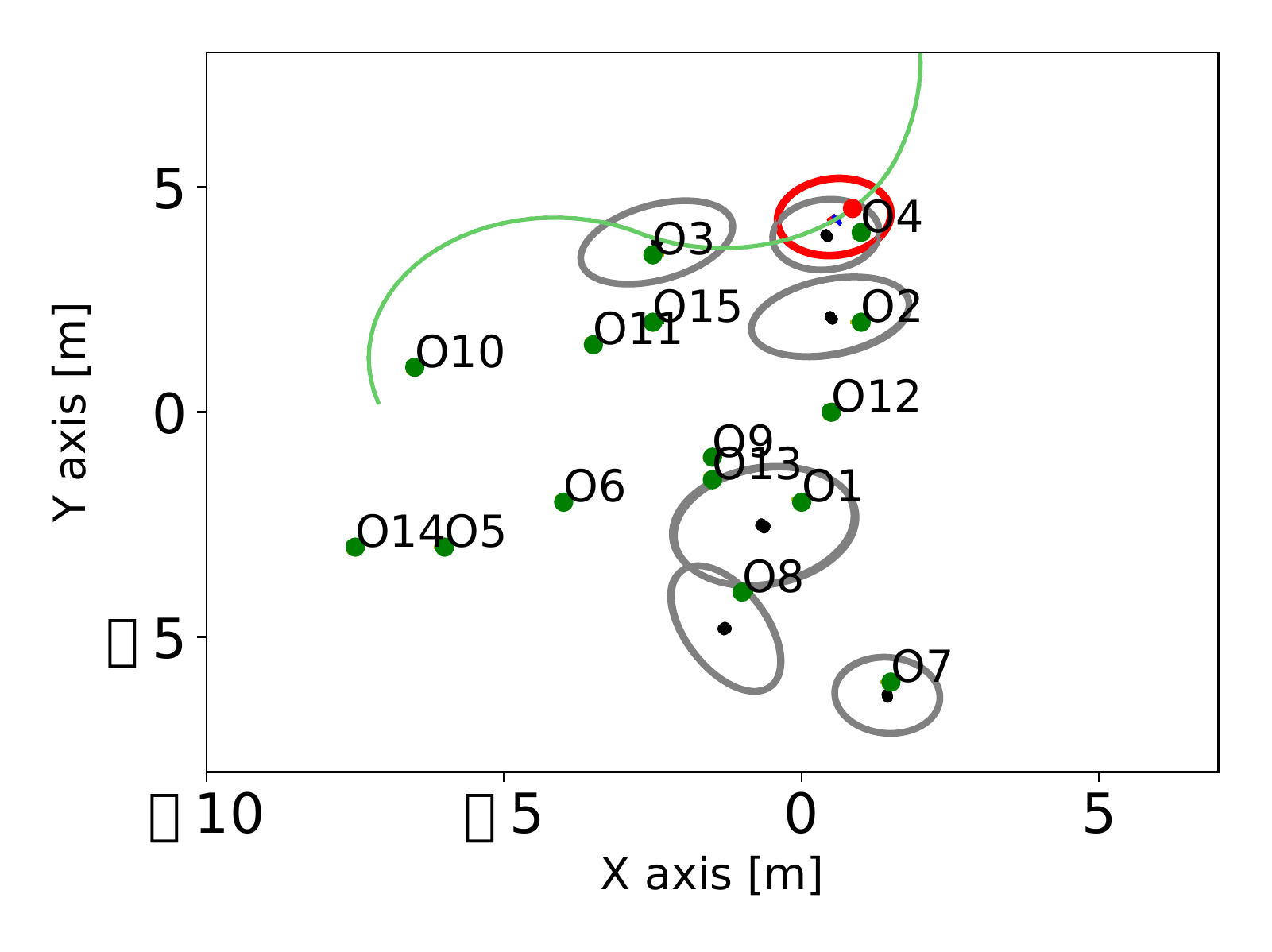}
		\caption{Dis. SLAM $r_2$}\label{fig:d_Robot_2_x15}
	\end{subfigure}
	\begin{subfigure}[b]{0.24\textwidth}
		\includegraphics[width=\textwidth]{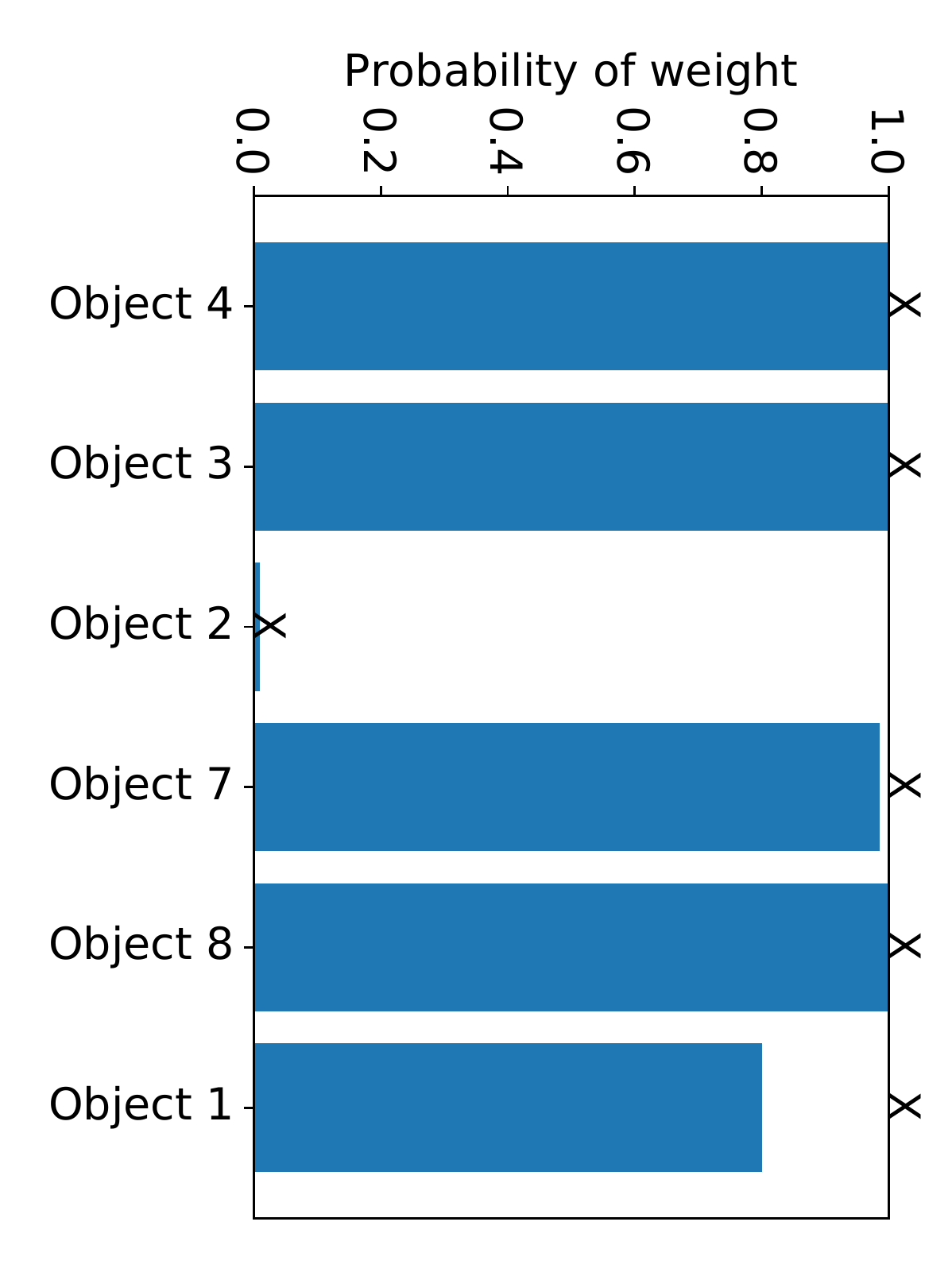}
		\caption{Dis. classification $r_2$}\label{fig:d_Robot_2_x15_cls}
	\end{subfigure}
	\begin{subfigure}[b]{0.24\textwidth}
		\includegraphics[width=\textwidth]{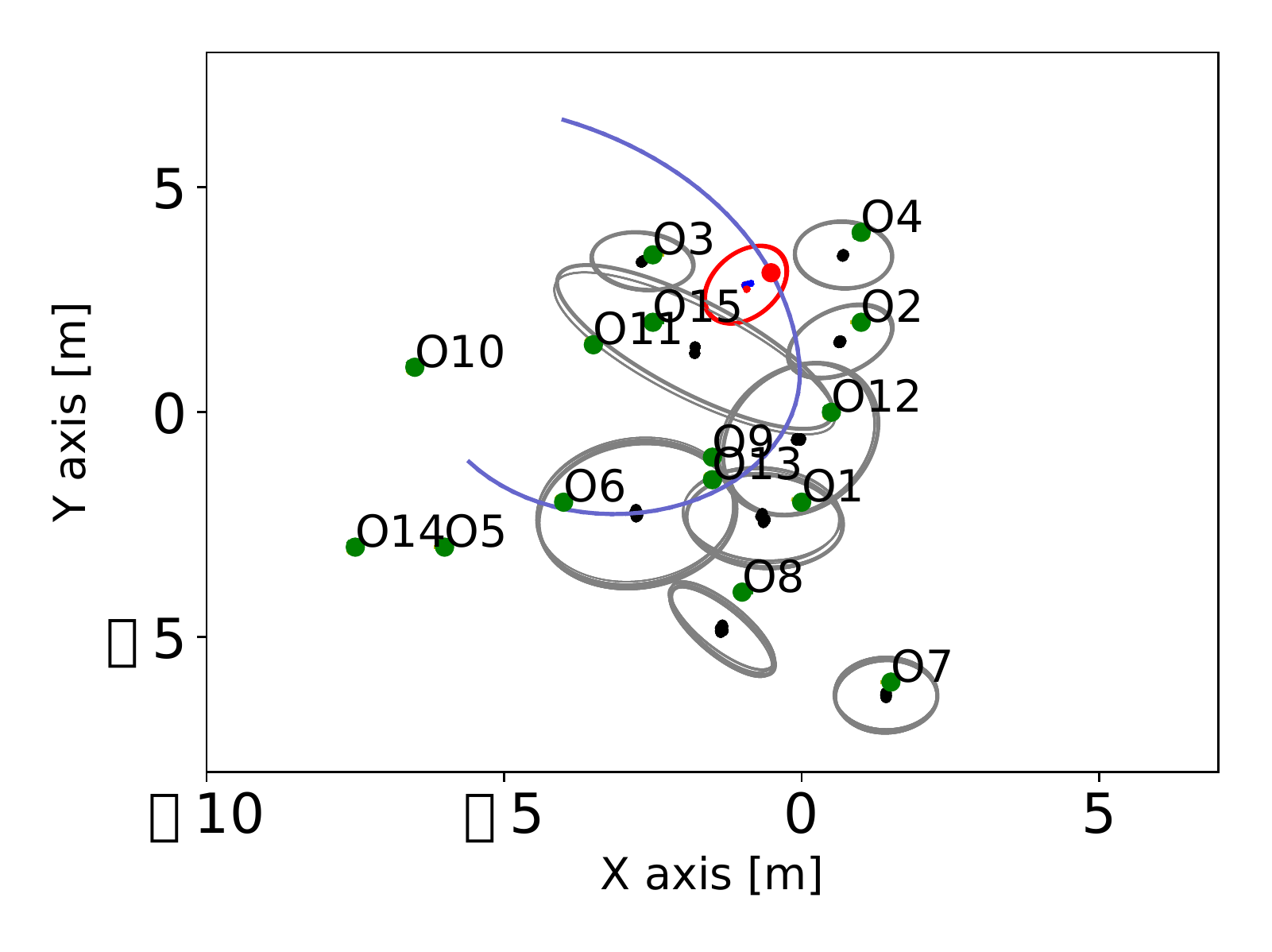}
		\caption{Dis. SLAM $r_3$}\label{fig:d_Robot_3_x20}
	\end{subfigure}
	\begin{subfigure}[b]{0.24\textwidth}
		\includegraphics[width=\textwidth]{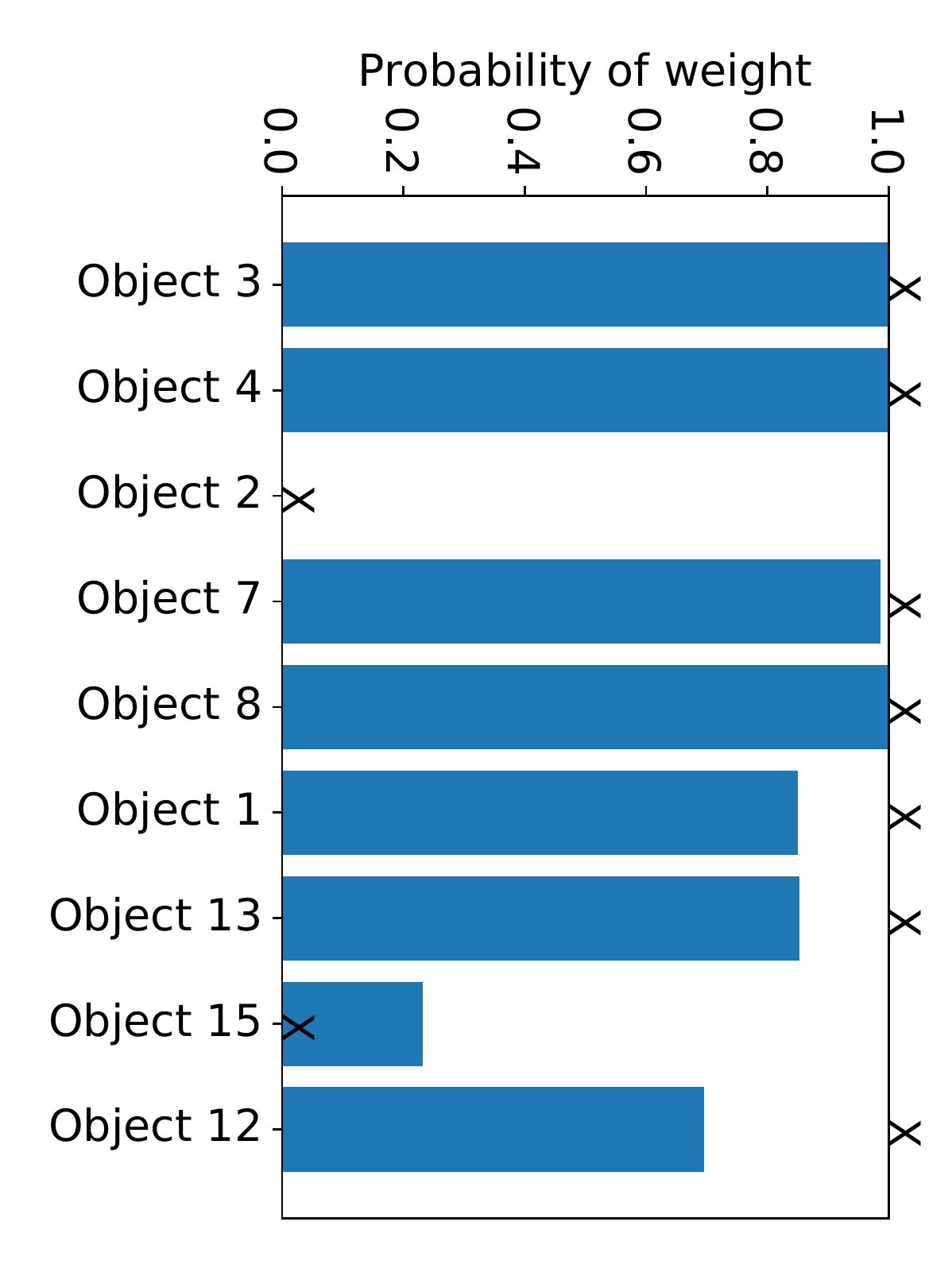}
		\caption{Dis. classification $r_3$}\label{fig:d_Robot_3_x20_cls}
	\end{subfigure}
	
	\caption{Figures for robot $r_2$ and $r_3$, distributed beliefs for time $k=15$ and $k=20$ respectively. \textbf{(a)} and \textbf{(b)} show results for $r_2$, \textbf{(c)} and \textbf{(d)} for $r_3$. \textbf{(a)} and \textbf{(c)} present SLAM results, \textbf{(b)} and \textbf{(d)} present classification results.}
	\label{fig:Figures_x4_dis}
\end{figure}

\begin{figure}[!htbp]
	
	\begin{subfigure}[b]{0.24\textwidth}
		\includegraphics[width=\textwidth]{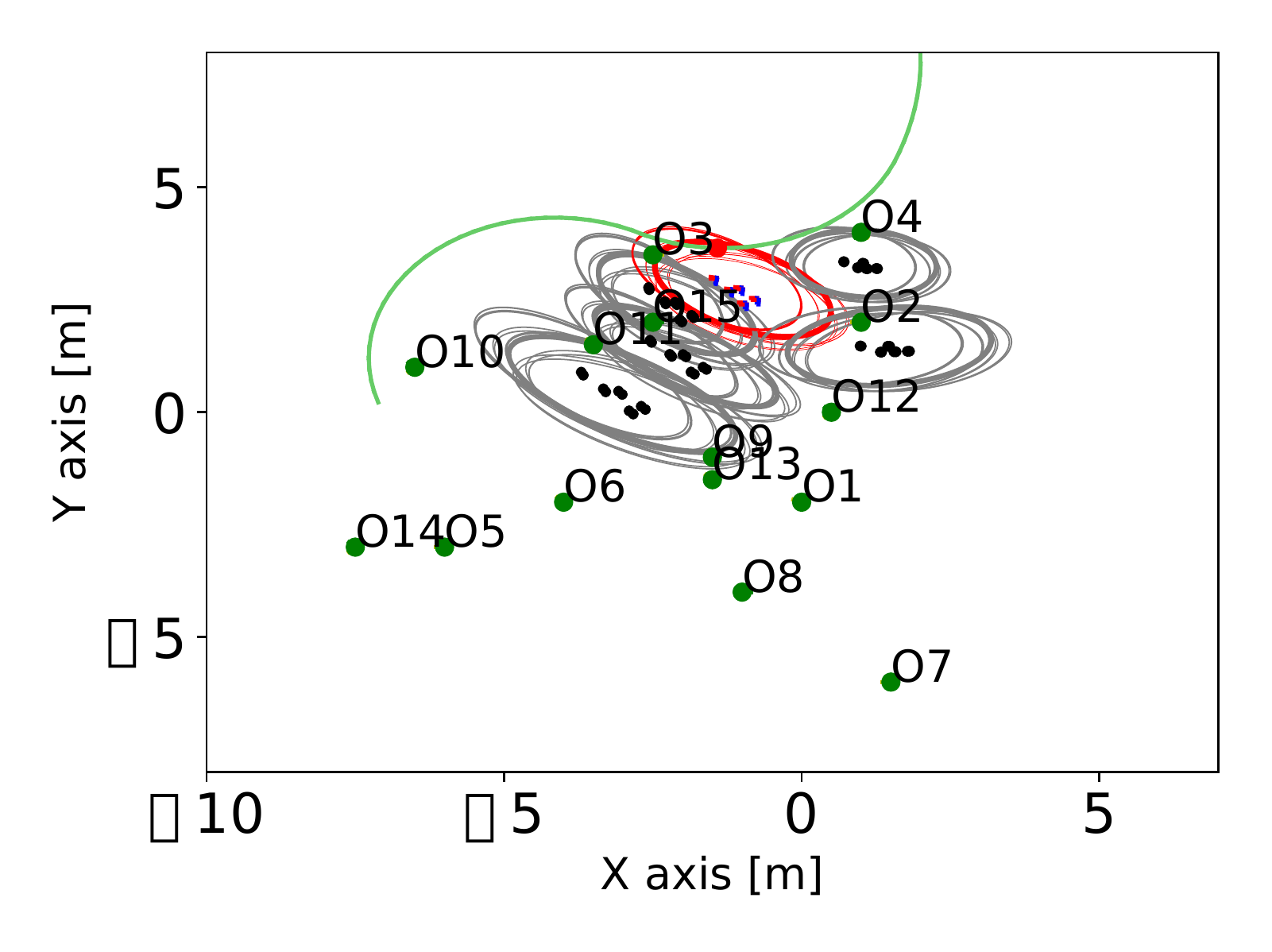}
		\caption{Local SLAM $r_2$}\label{fig:Robot_2_x25}
	\end{subfigure}
	\begin{subfigure}[b]{0.24\textwidth}
		\includegraphics[width=\textwidth]{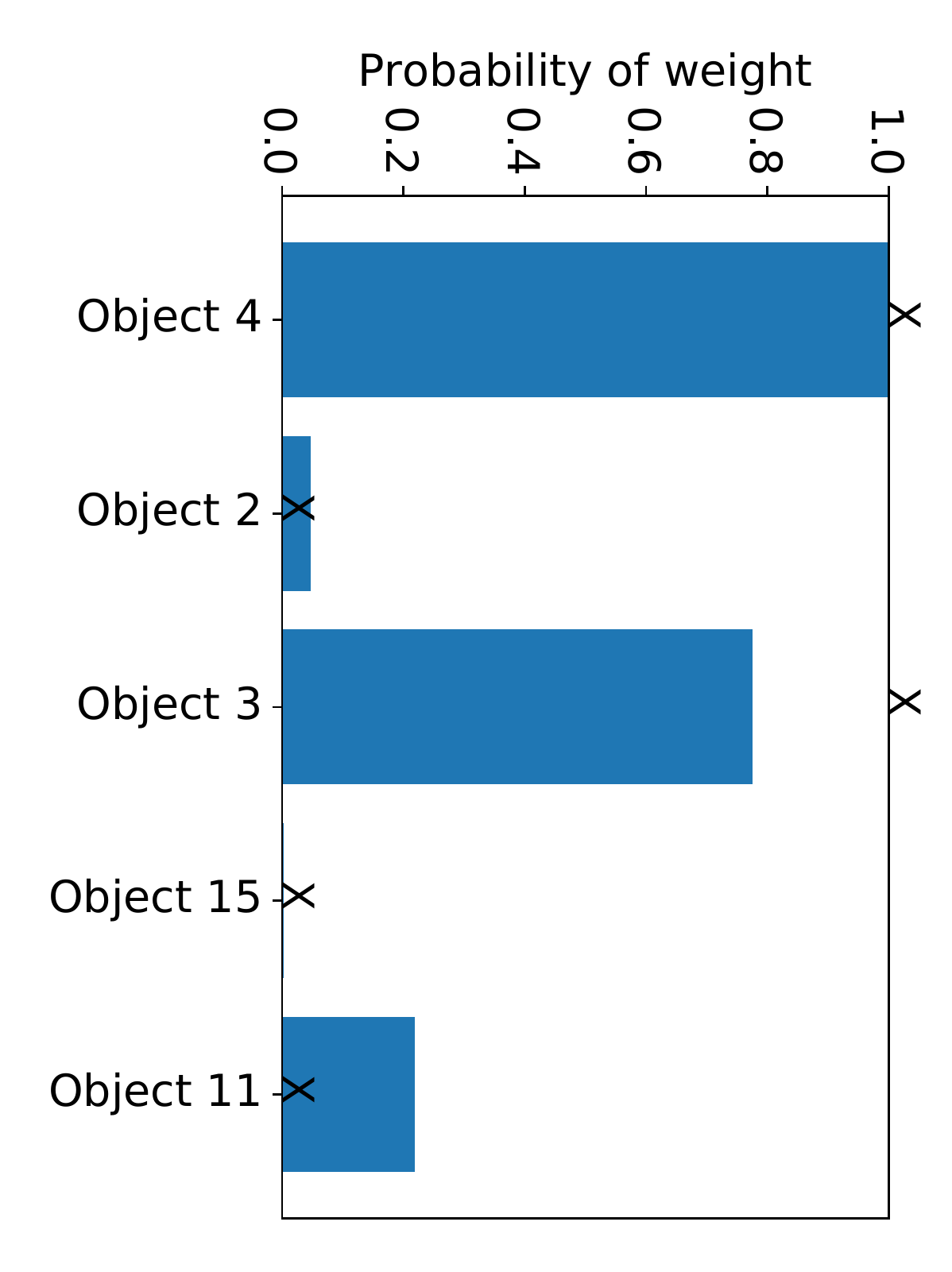}
		\caption{Local classification $r_2$}\label{fig:Robot_2_x25_cls}
	\end{subfigure}
	\begin{subfigure}[b]{0.24\textwidth}
		\includegraphics[width=\textwidth]{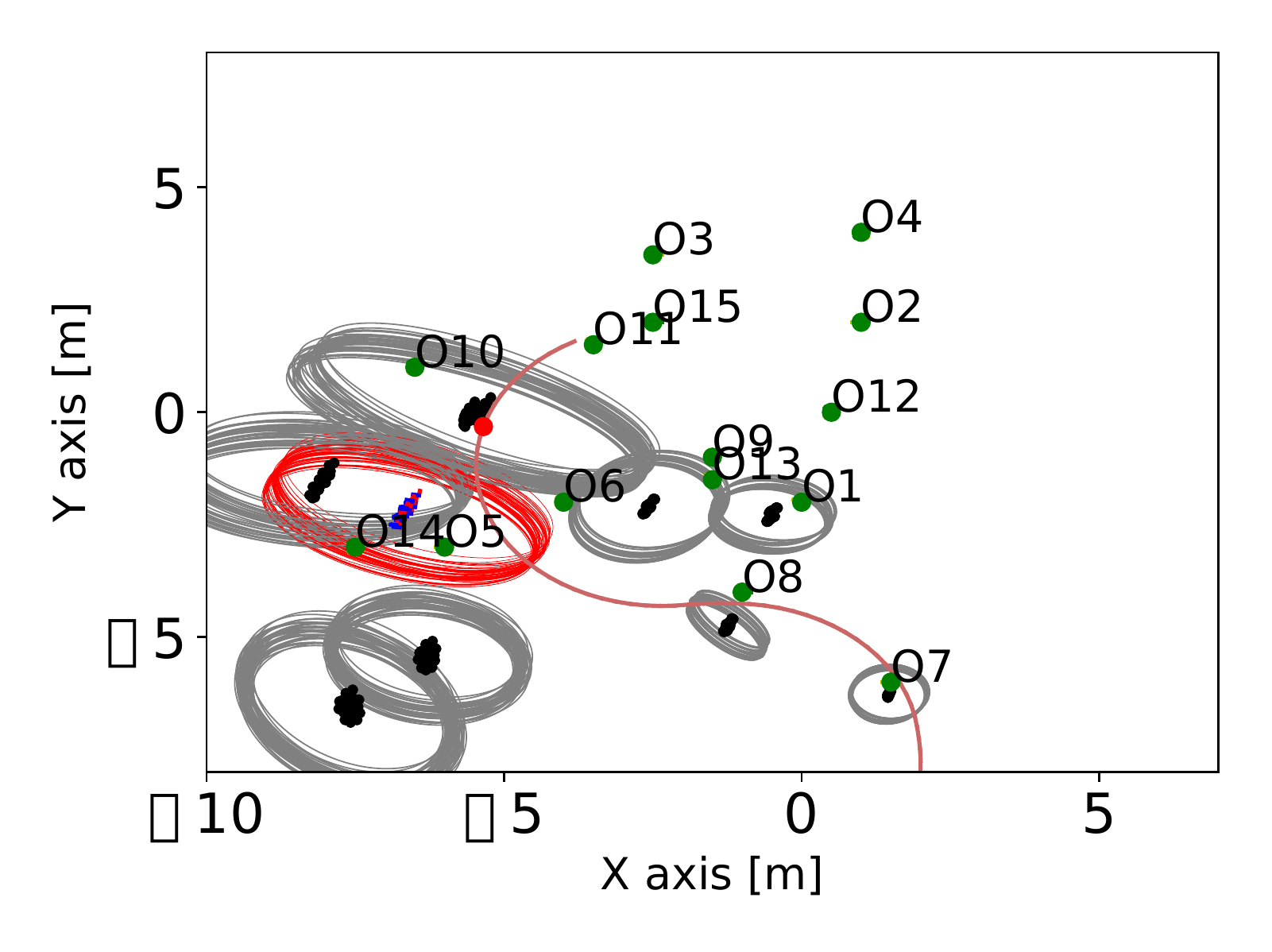}
		\caption{Local SLAM $r_1$}\label{fig:Robot_1_x50}
	\end{subfigure}
	\begin{subfigure}[b]{0.24\textwidth}
		\includegraphics[width=\textwidth]{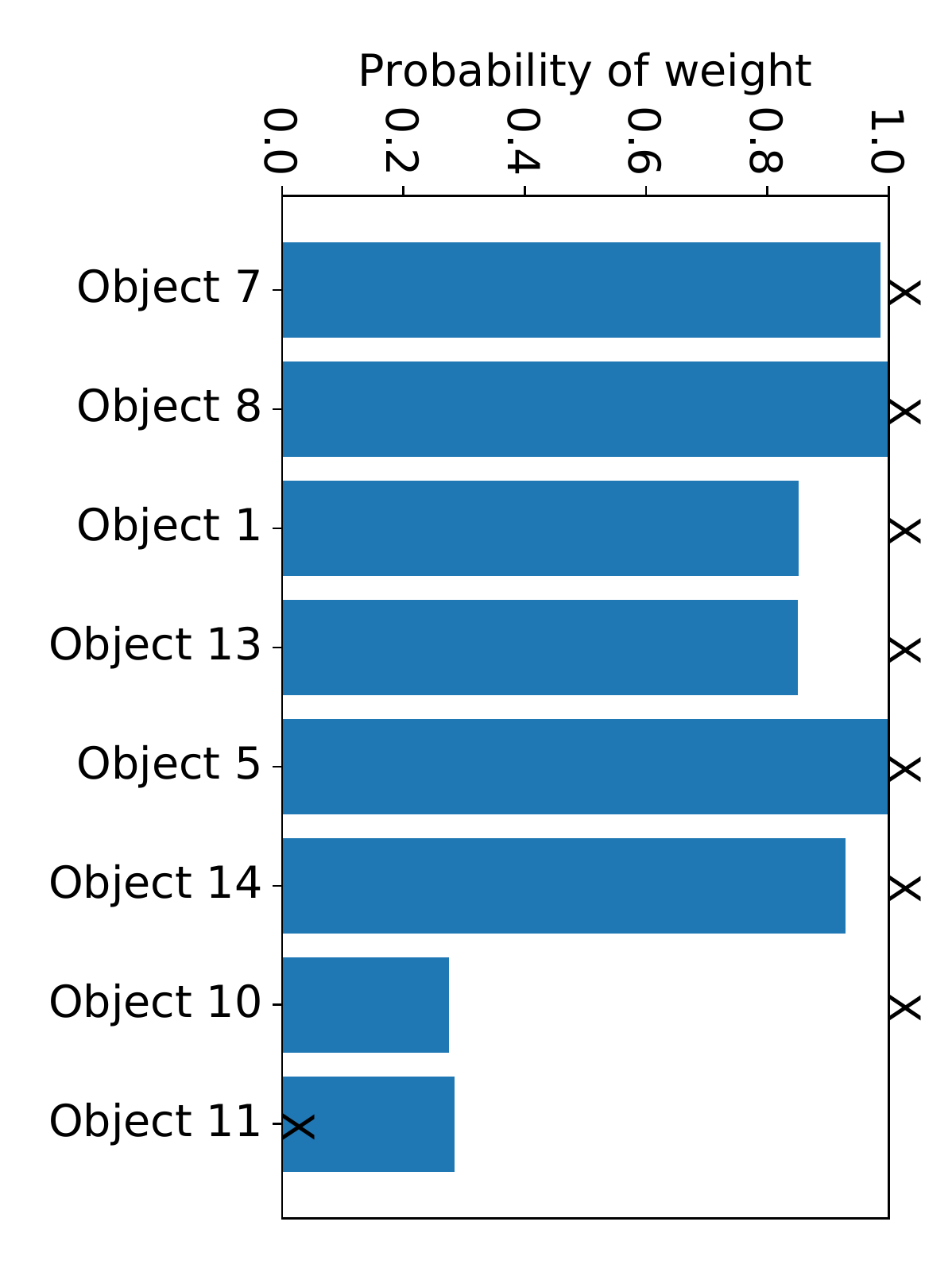}
		\caption{Local classification $r_1$}\label{fig:Robot_1_x50_cls}
	\end{subfigure}
	
	\caption{Figures for robot $r_2$ and $r_1$, local beliefs for time $k=25$ and $k=50$ respectively. \textbf{(a)} and \textbf{(b)} show results for $r_2$, \textbf{(c)} and \textbf{(d)} for $r_1$. \textbf{(a)} and \textbf{(c)} present SLAM results, \textbf{(b)} and \textbf{(d)} present classification results.}
	\label{fig:Figures_x21_local}
\end{figure}

\begin{figure}[!htbp]
	
	\begin{subfigure}[b]{0.24\textwidth}
		\includegraphics[width=\textwidth]{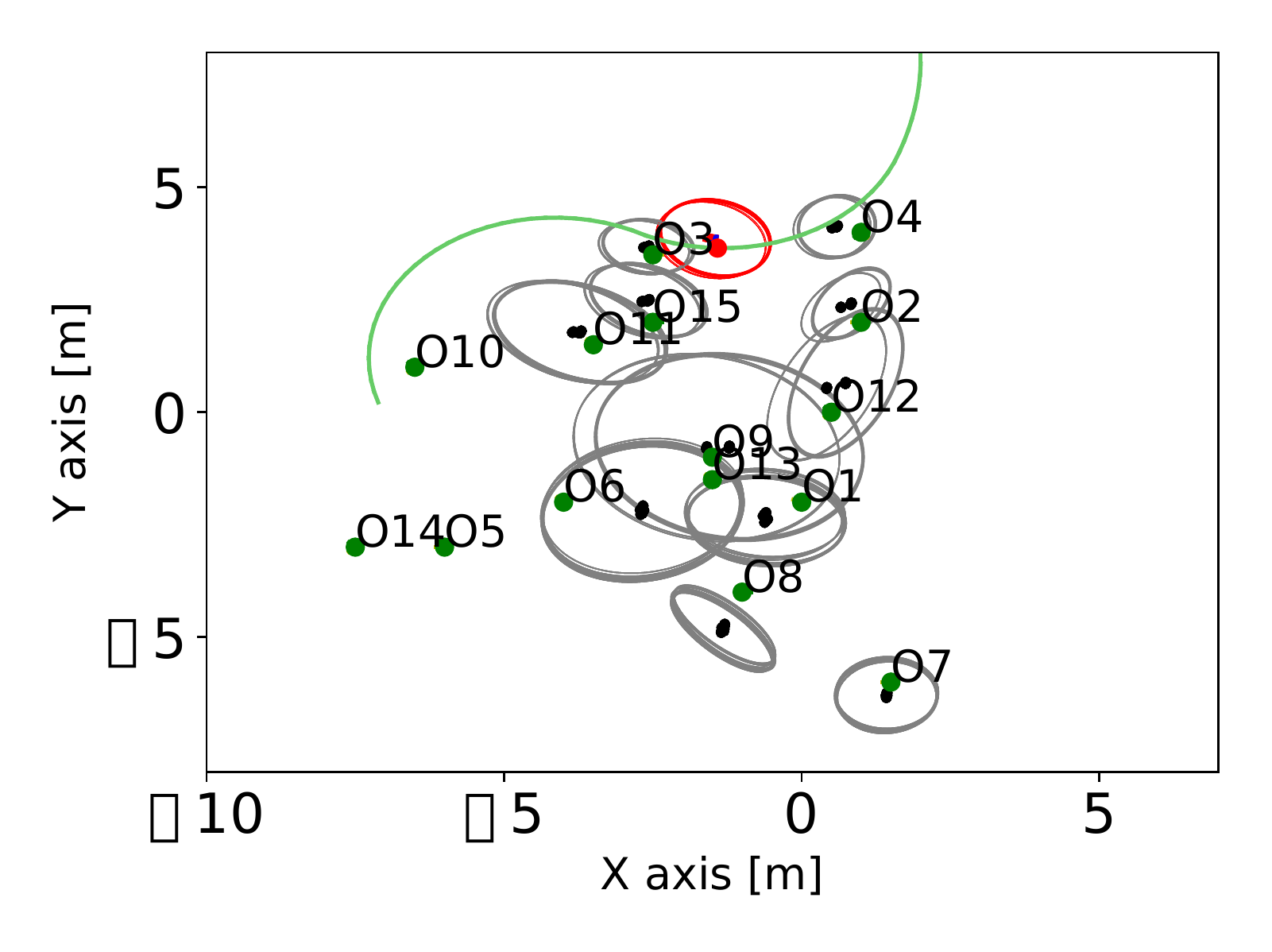}
		\caption{Dis. SLAM $r_2$}\label{fig:d_Robot_2_x25}
	\end{subfigure}
	\begin{subfigure}[b]{0.24\textwidth}
		\includegraphics[width=\textwidth]{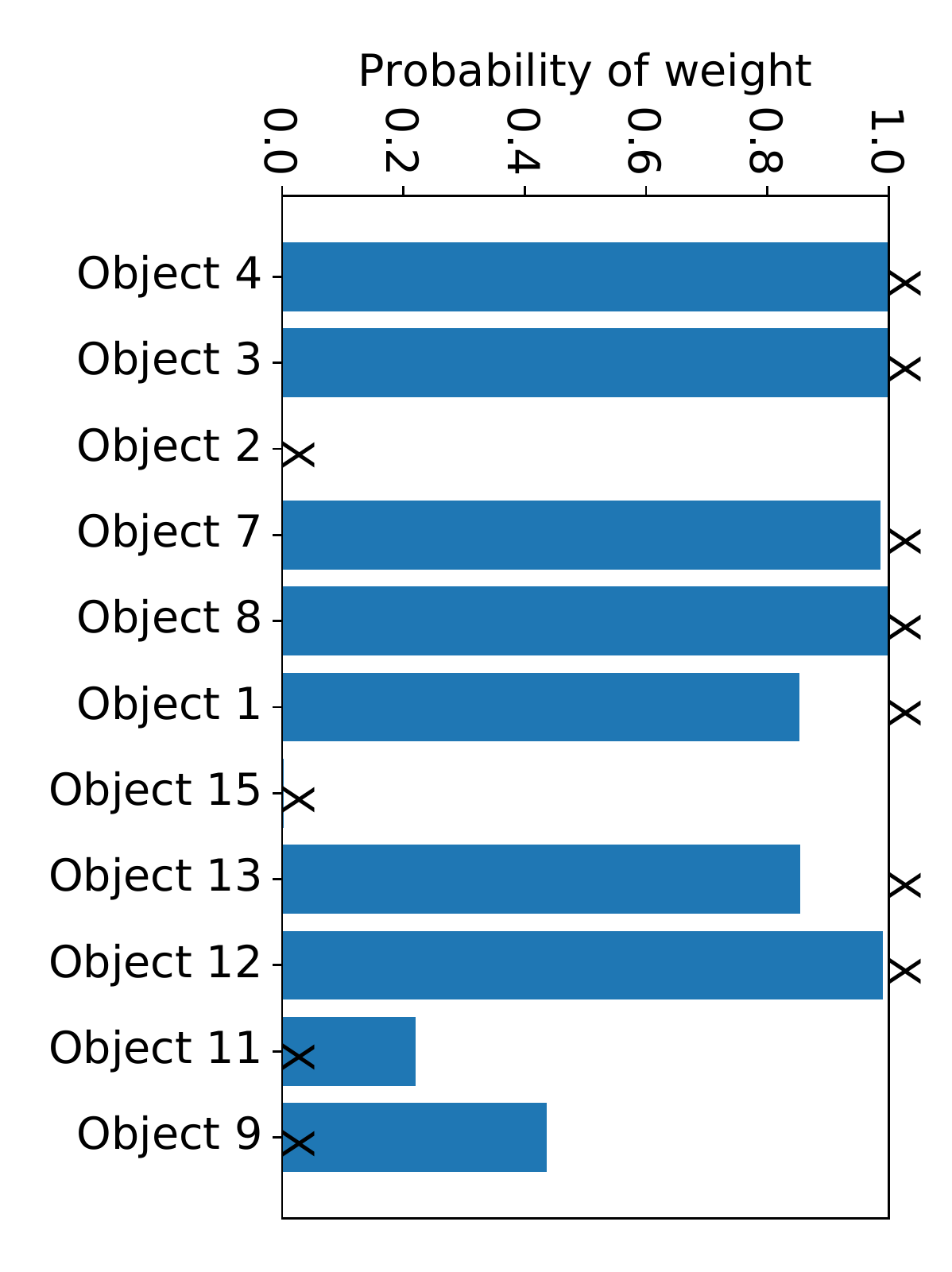}
		\caption{Dis. classification $r_2$}\label{fig:d_Robot_2_x25_cls}
	\end{subfigure}
	\begin{subfigure}[b]{0.24\textwidth}
		\includegraphics[width=\textwidth]{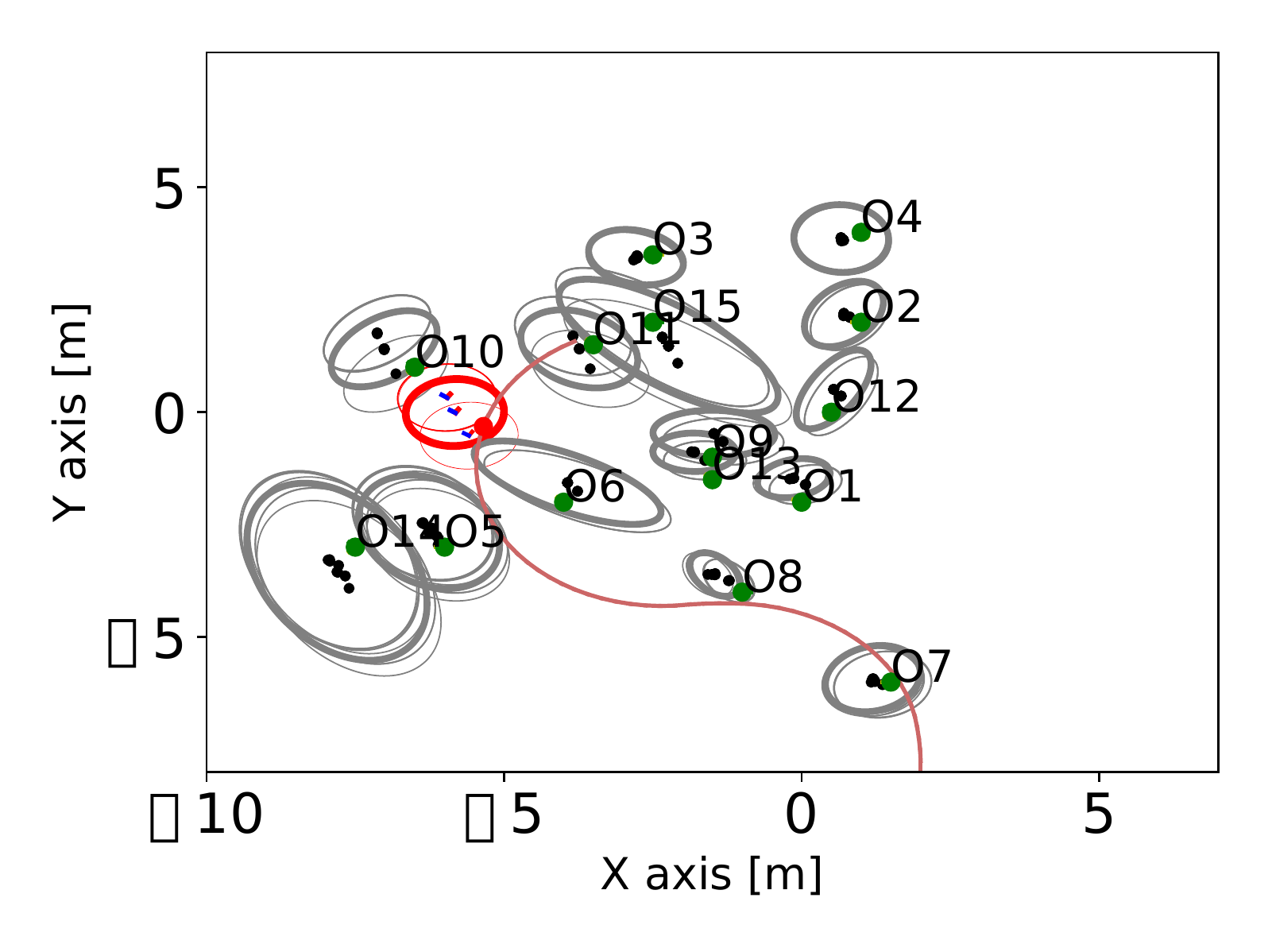}
		\caption{Dis. SLAM $r_1$}\label{fig:d_Robot_1_x50}
	\end{subfigure}
	\begin{subfigure}[b]{0.24\textwidth}
		\includegraphics[width=\textwidth]{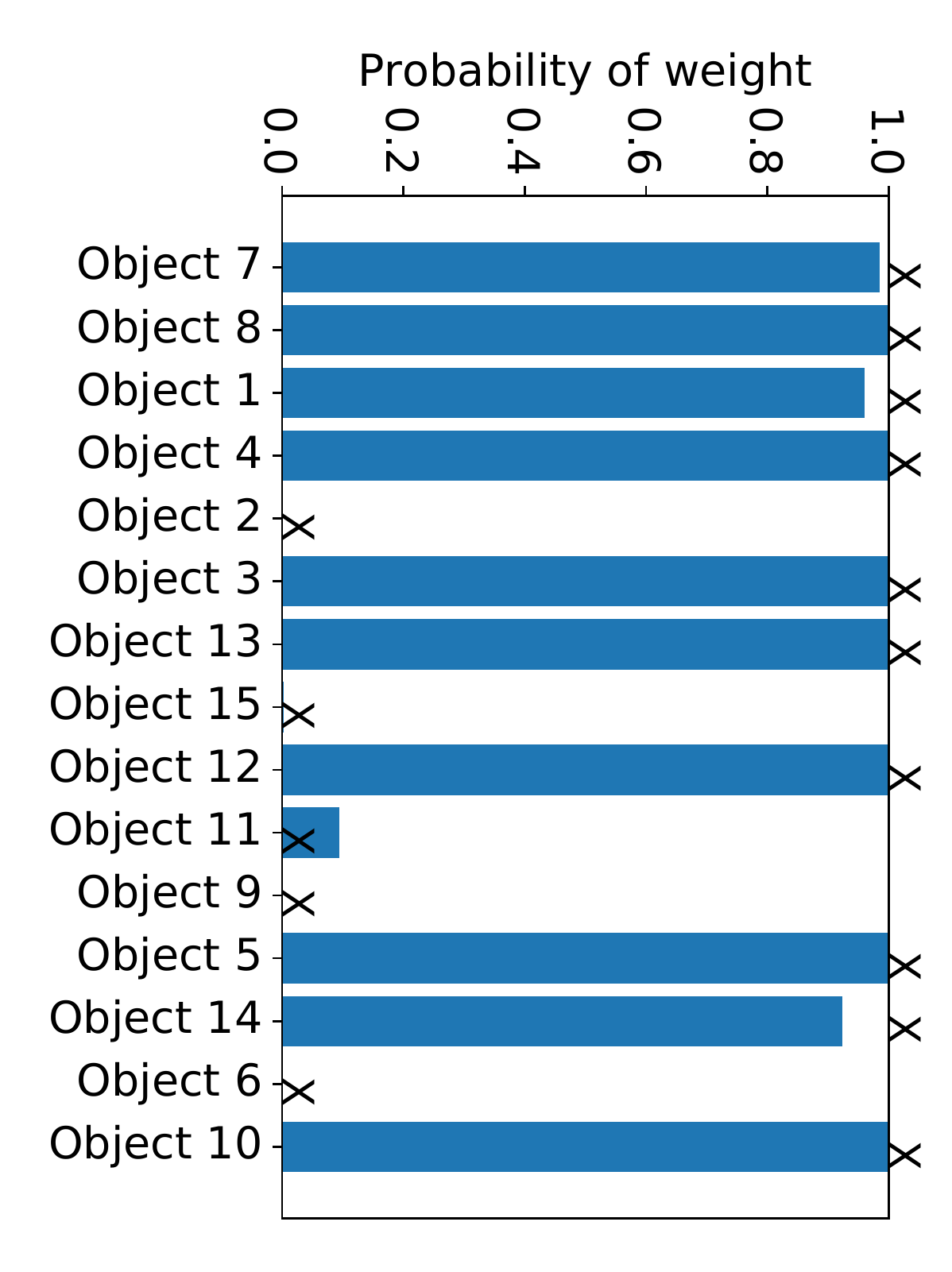}
		\caption{Dis. classification $r_1$}\label{fig:d_Robot_1_x50_cls}
	\end{subfigure}
	
	\caption{Figures for robot $r_2$ and $r_1$, distributed beliefs for time $k=25$ and $k=50$ respectively. \textbf{(a)} and \textbf{(b)} show results for $r_2$, \textbf{(c)} and \textbf{(d)} for $r_1$. \textbf{(a)} and \textbf{(c)} present SLAM results, \textbf{(b)} and \textbf{(d)} present classification results.}
	\label{fig:Figures_x21_dis}
\end{figure}

In this section we present additional results for the simulation. In Fig.~\ref{fig:Figures_x4_local}, \ref{fig:Figures_x4_dis}, \ref{fig:Figures_x21_local}, and \ref{fig:Figures_x21_dis} we show the beliefs at various stages of the path. 
	
The results of all the graphs support the paper results, where both classification and SLAM in general are more accurate for the distributed belief. In addition, the robots inferring the distributed belief take into account objects that they didn't observe directly.

\begin{figure}[!htbp]
	
	\includegraphics[width=0.5\textwidth]{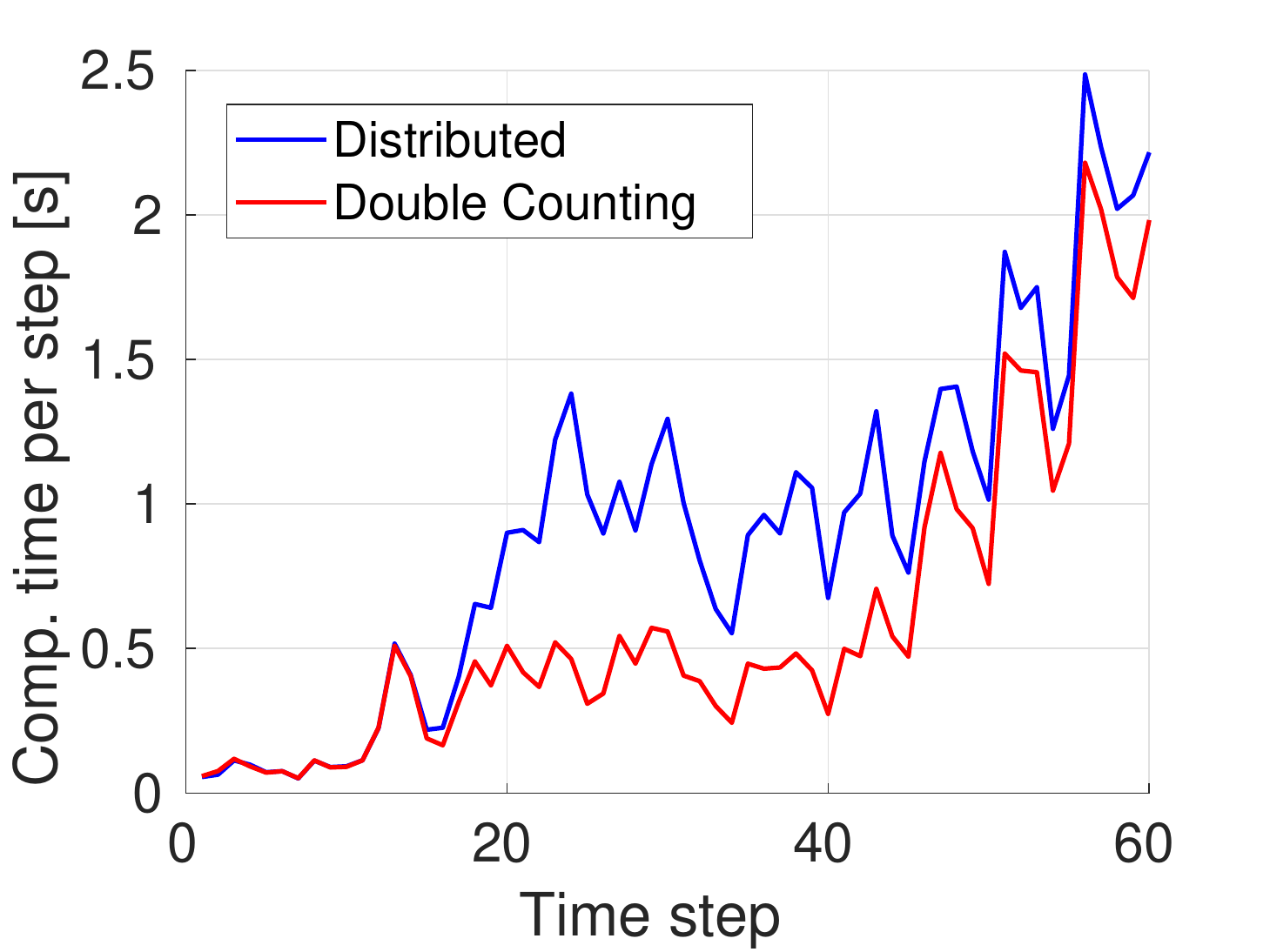}
	\caption{Calculation time as a function of the time step in seconds.}\label{fig:Time_Sim}

\end{figure}

In Fig.~\ref{fig:Time_Sim} we show the time each inference time-step takes to compute for the distributed case, without and with double-counting. In general, computation time is influenced by the number of class realizations that aren't pruned, and is higher when robots communicate between each other. For each newly observed object the algorithm must consider all realizations for the said object, thus the computation time "spikes" at the first step the new object is observed.

\clearpage
\section{Experiment: Parameters}

We consider a motion model with noise covariance $\Sigma_w = diag(0.0003,0.0003, 0.0001)$, and geometric model with noise  covariance $\Sigma^{geo}_v=diag(0.04,0.04,0.005)$, both corresponding to position coordinates in meters and orientation in radians. We simulated noisy odometry and geometric measurements, while using YOLO3 to create object proposals and a classifier to classify them. The communication radius in this scenario is 3 meters. The robot's and chair ground truth was measured via motion capture cameras with OptiTrack. The chairs' center of mass is used as a frame of reference for relative poses.

The classifier used in our experiment is the Pytorch implementation of ResNet 50, pre-trained on ImageNet dataset \cite{Deng2009cvpr}.
We trained three classifier models, one per each class. Class $c=1$ is 'Barber Chair' and is considered our ground truth. Class $c=2$ is 'Punching Bag' and class $c=3$ is 'Traffic Light'. We trained the classifiers using pairs of relative pose and probability vectors; for $c=1$, we used images of a chair used in the experiment, while for $c=2$ and $c=3$, we sampled measurements from Dirichlet Distribution with parameters $\alpha=[5,15,3]$ and $\alpha = [5,3,15]$ respectively. Each relative pose was parametrized by the relative yaw angle $\psi$, and the relative $\theta$, with the camera being viewed from the object's frame of reference.

\begin{figure}[!htbp]
	
	\begin{subfigure}[b]{0.24\textwidth}
		\includegraphics[width=\textwidth]{Chair1.jpg}
	\end{subfigure}
	\begin{subfigure}[b]{0.24\textwidth}
		\includegraphics[width=\textwidth]{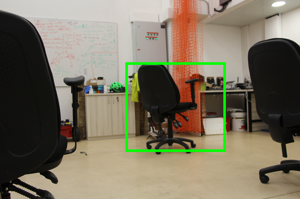}
	\end{subfigure}
	\begin{subfigure}[b]{0.24\textwidth}
		\includegraphics[width=\textwidth]{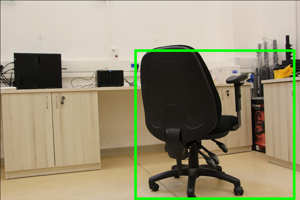}
	\end{subfigure}
	\begin{subfigure}[b]{0.24\textwidth}
		\includegraphics[width=\textwidth]{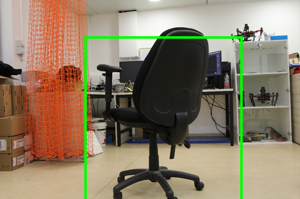}
	\end{subfigure}
	
	\caption{Four of the experiment images shown with corresponding bounding boxes.}
	\label{fig:Chairs}
\end{figure}

\begin{figure}[!htbp]
	
	\begin{subfigure}[b]{0.48\textwidth}
		\includegraphics[width=\textwidth]{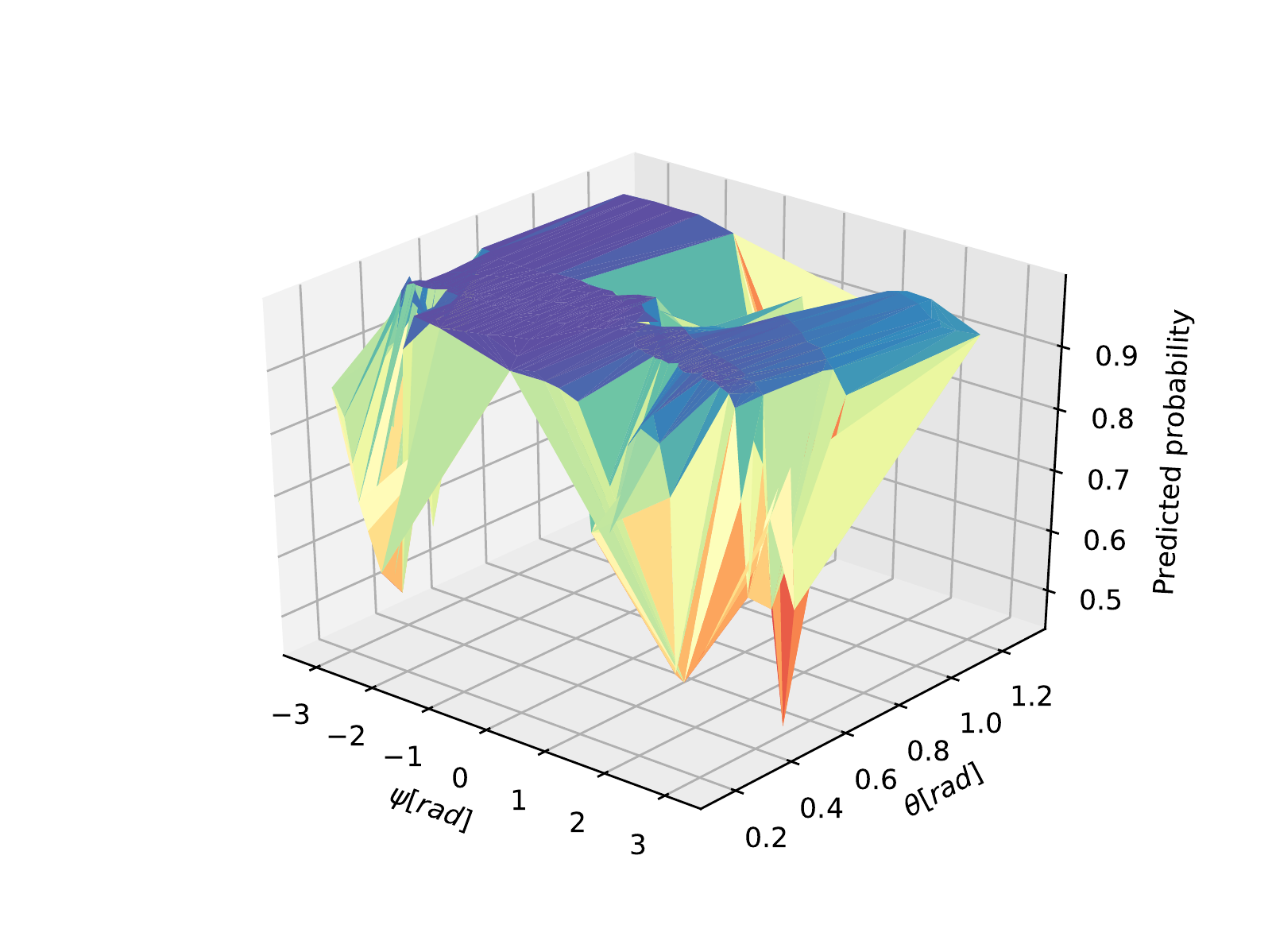}
		\caption{$c=1$ model, $c=1$ prob.}\label{fig:C1_1}
	\end{subfigure}
	\begin{subfigure}[b]{0.48\textwidth}
		\includegraphics[width=\textwidth]{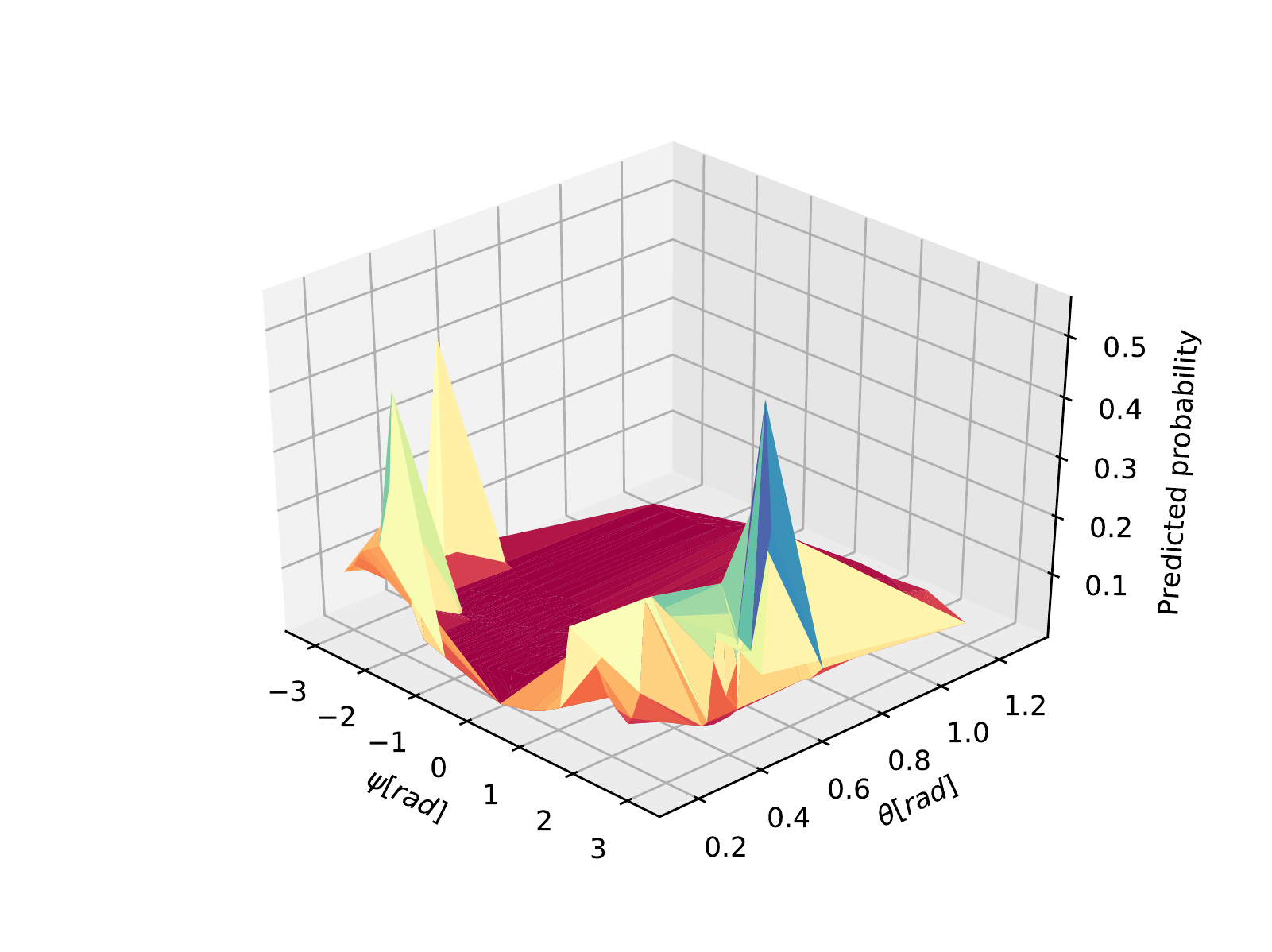}
		\caption{$c=1$ model, $c=2$ prob.}\label{fig:C1_2}
	\end{subfigure}
	
	\begin{subfigure}[b]{0.5\textwidth}
		\includegraphics[width=\textwidth]{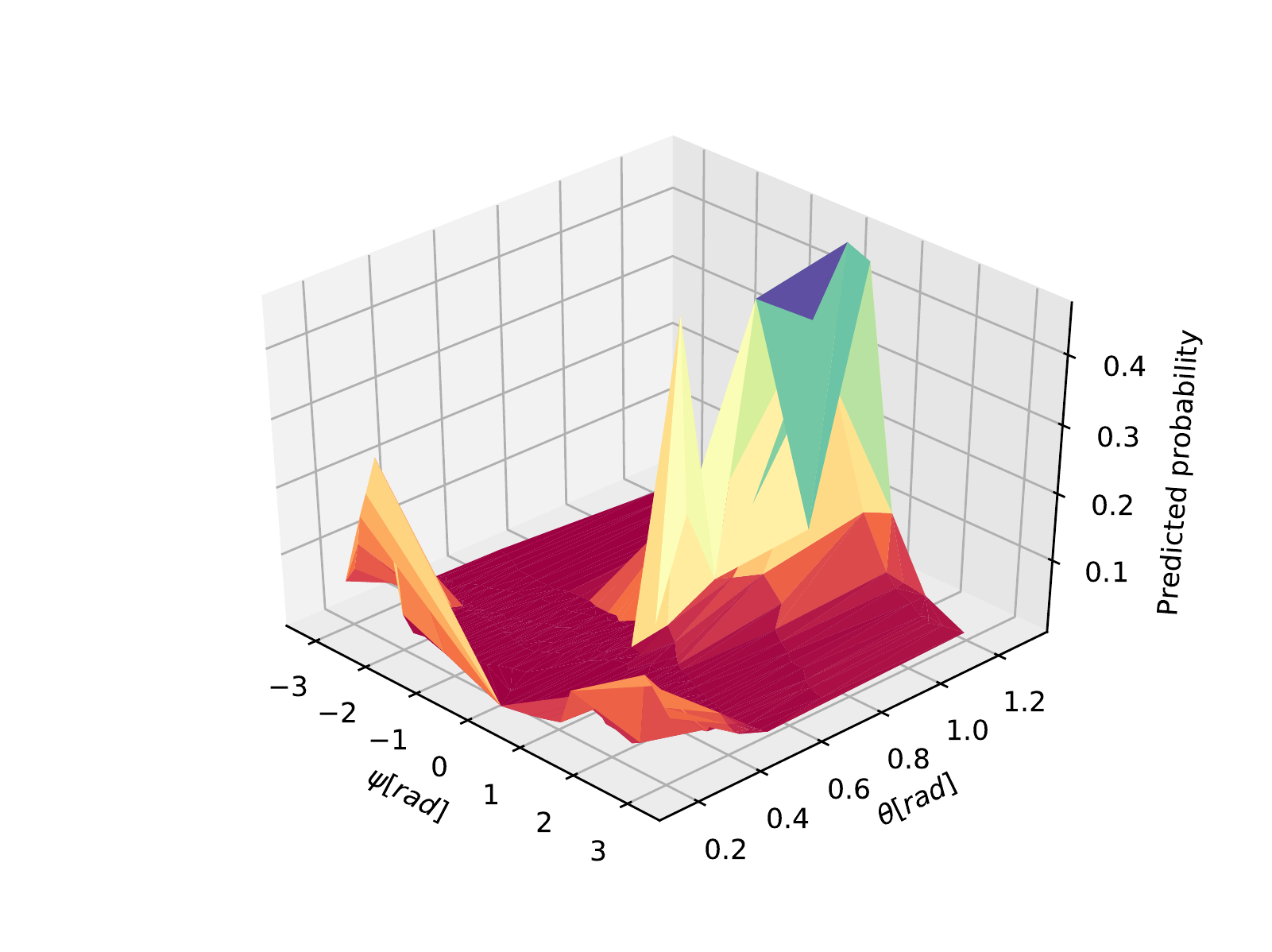}
		\caption{$c=1$ model, $c=3$ prob.}\label{fig:C1_3}
	\end{subfigure}
	\caption{Classifier model for $c=1$, trained on real images: probabilities of classes 1 to 3 depending on relative yaw and pitch angles presented i \textbf{(a)} to \textbf{(c)} respectively. Higher surfaces go have bluer color.}
	\label{fig:C1}
\end{figure}

\begin{figure}[!htbp]
	
	\begin{subfigure}[b]{0.48\textwidth}
		\includegraphics[width=\textwidth]{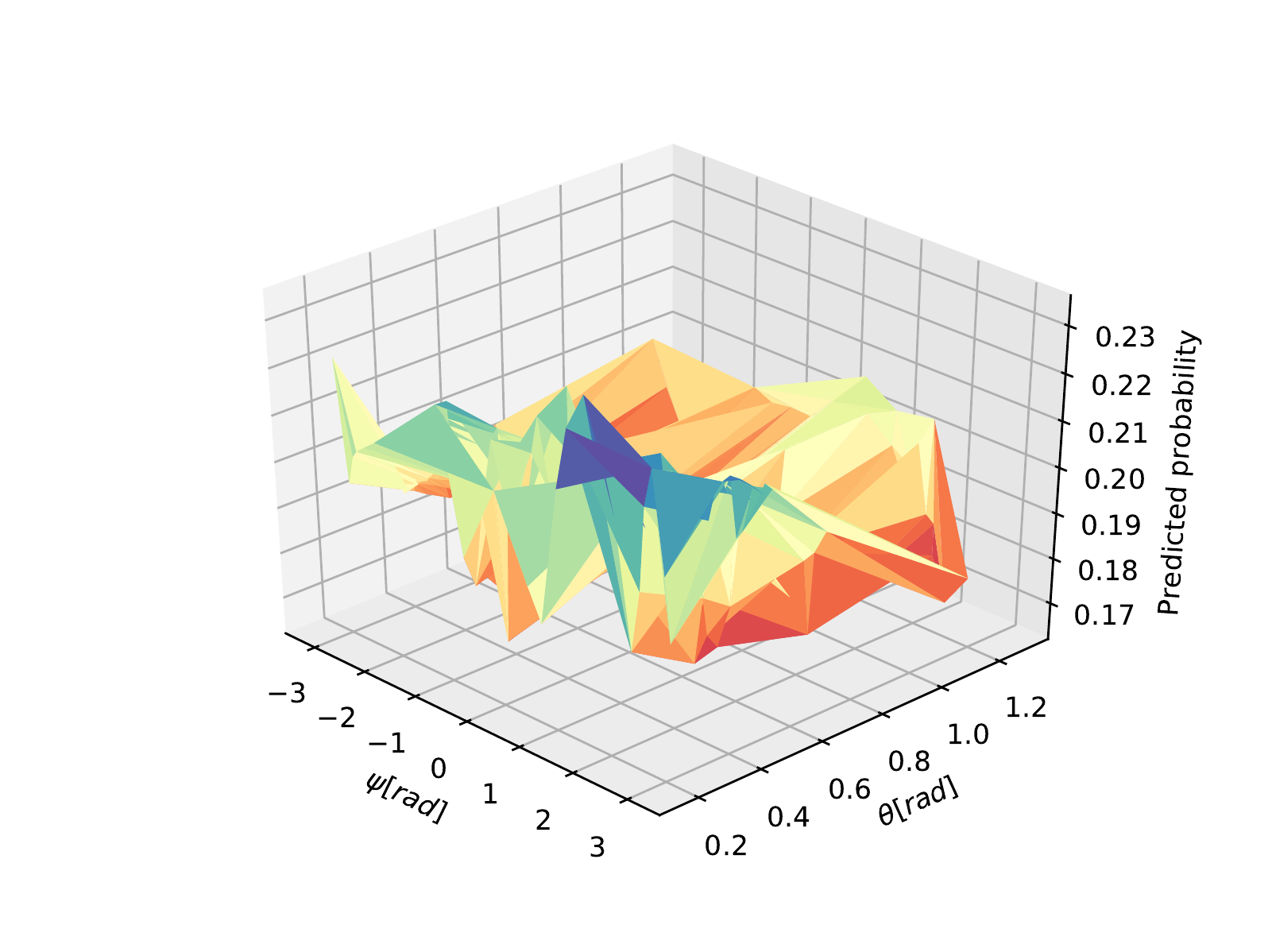}
		\caption{$c=2$ model, $c=1$ prob.}\label{fig:C2_1}
	\end{subfigure}
	\begin{subfigure}[b]{0.48\textwidth}
		\includegraphics[width=\textwidth]{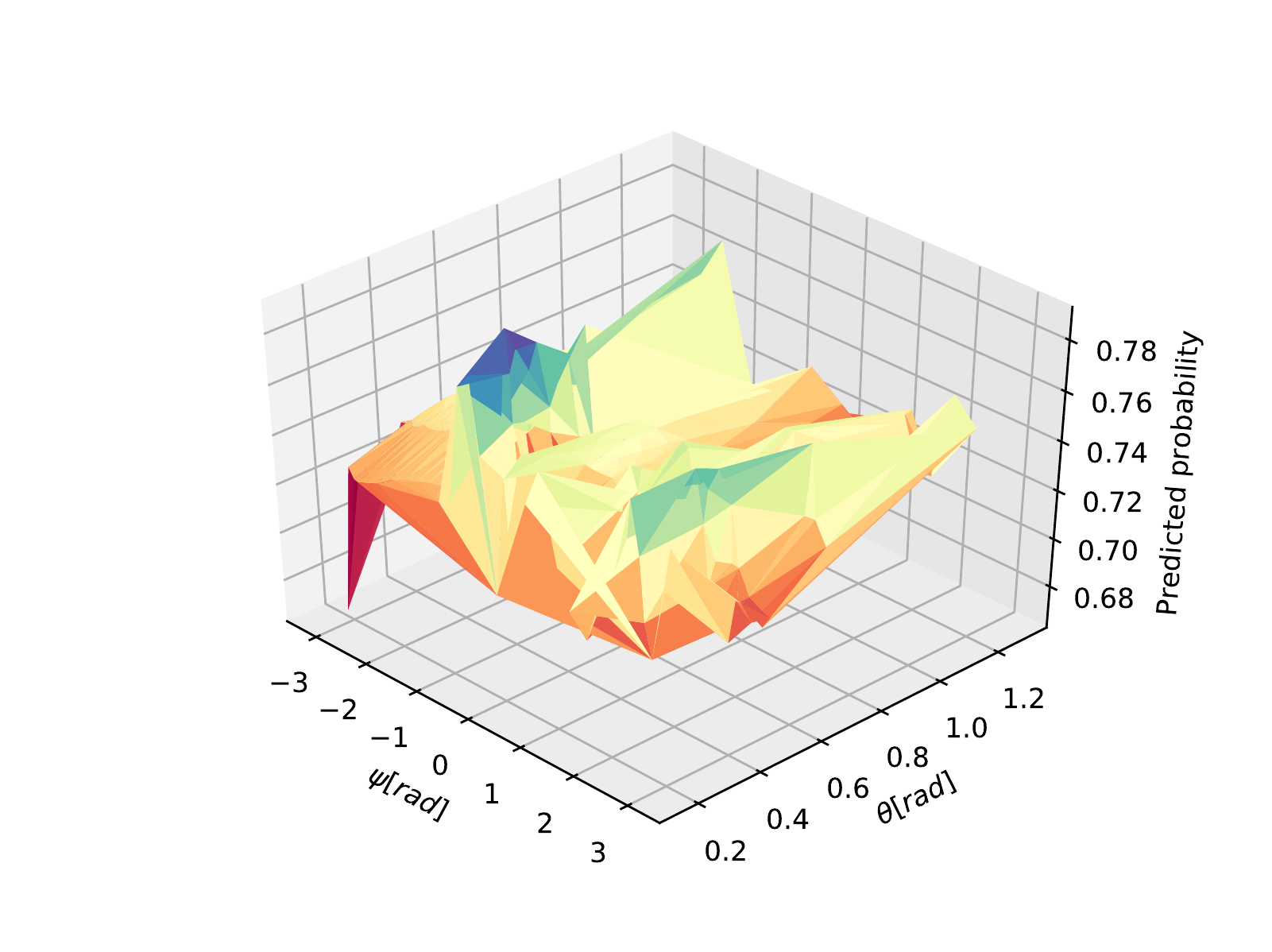}
		\caption{$c=2$ model, $c=2$ prob.}\label{fig:C2_2}
	\end{subfigure}
	
	\begin{subfigure}[b]{0.5\textwidth}
		\includegraphics[width=\textwidth]{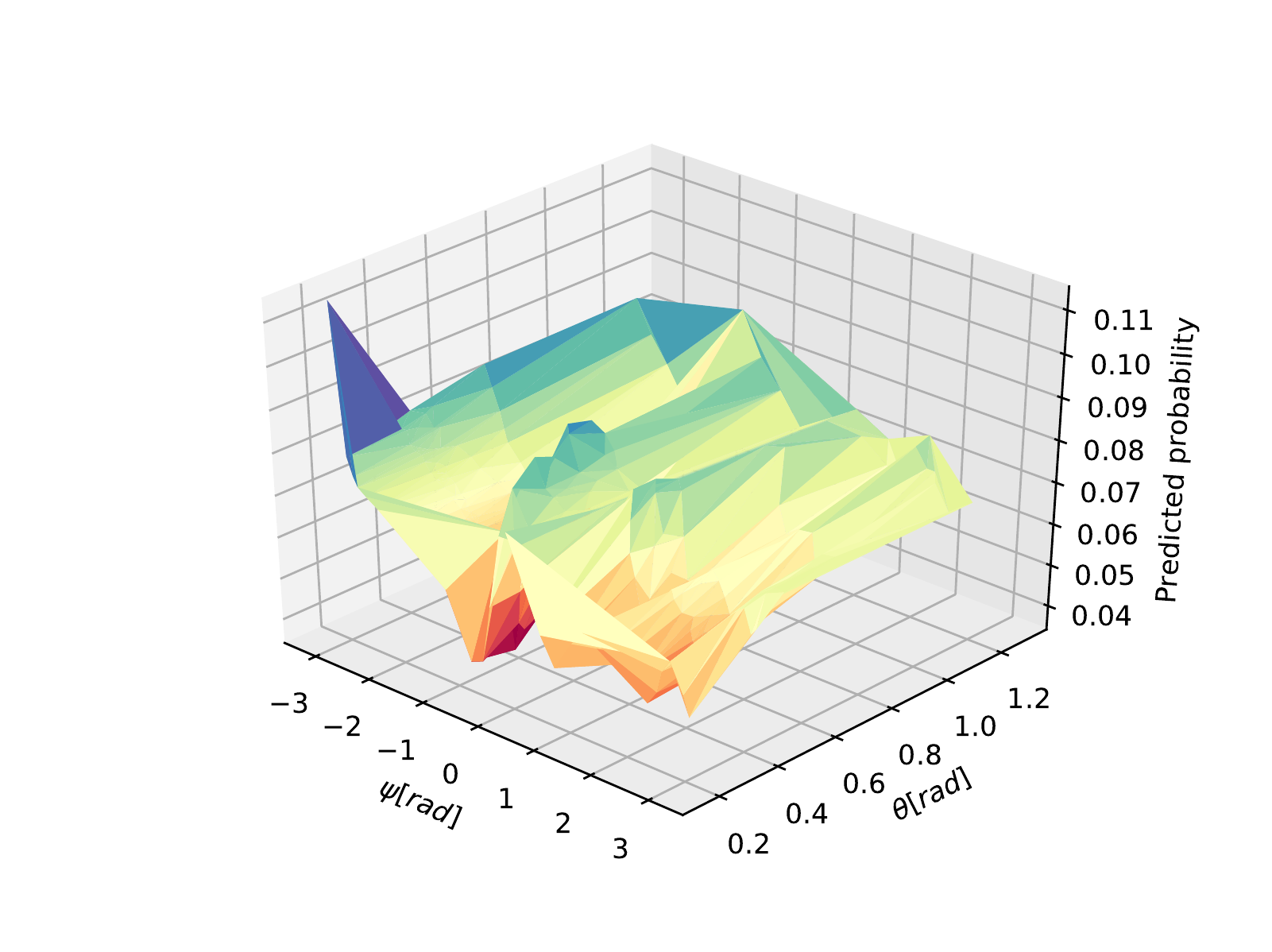}
		\caption{$c=2$ model, $c=3$ prob.}\label{fig:C2_3}
	\end{subfigure}
	\caption{Classifier model for $c=2$, trained on real images: probabilities of classes 1 to 3 depending on relative yaw and pitch angles presented i \textbf{(a)} to \textbf{(c)} respectively. Higher surfaces go have bluer color.}
	\label{fig:C2}
\end{figure}

\begin{figure}[!htbp]
	
	\begin{subfigure}[b]{0.48\textwidth}
		\includegraphics[width=\textwidth]{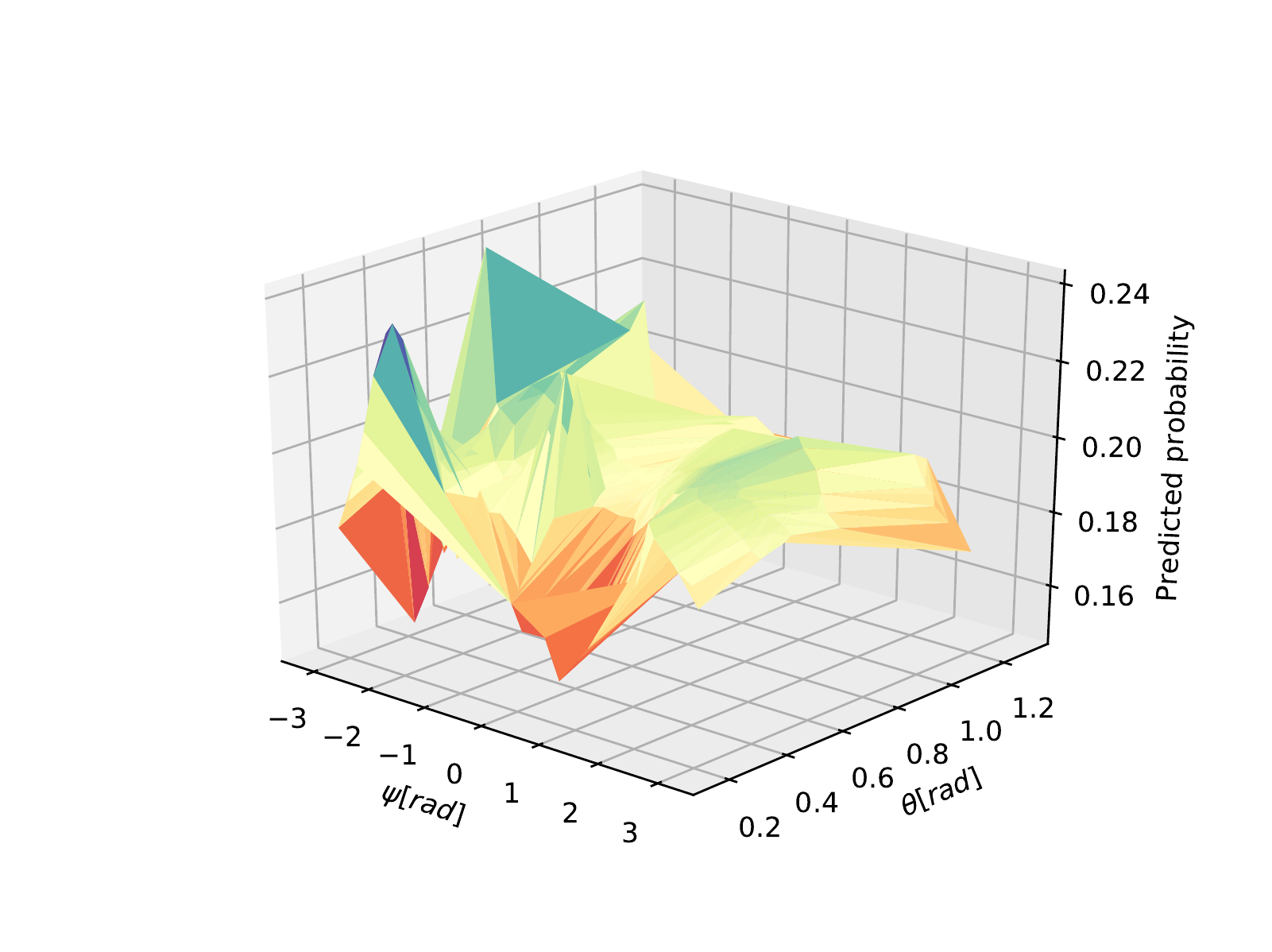}
		\caption{$c=3$ model, $c=1$ prob.}\label{fig:C3_1}
	\end{subfigure}
	\begin{subfigure}[b]{0.48\textwidth}
		\includegraphics[width=\textwidth]{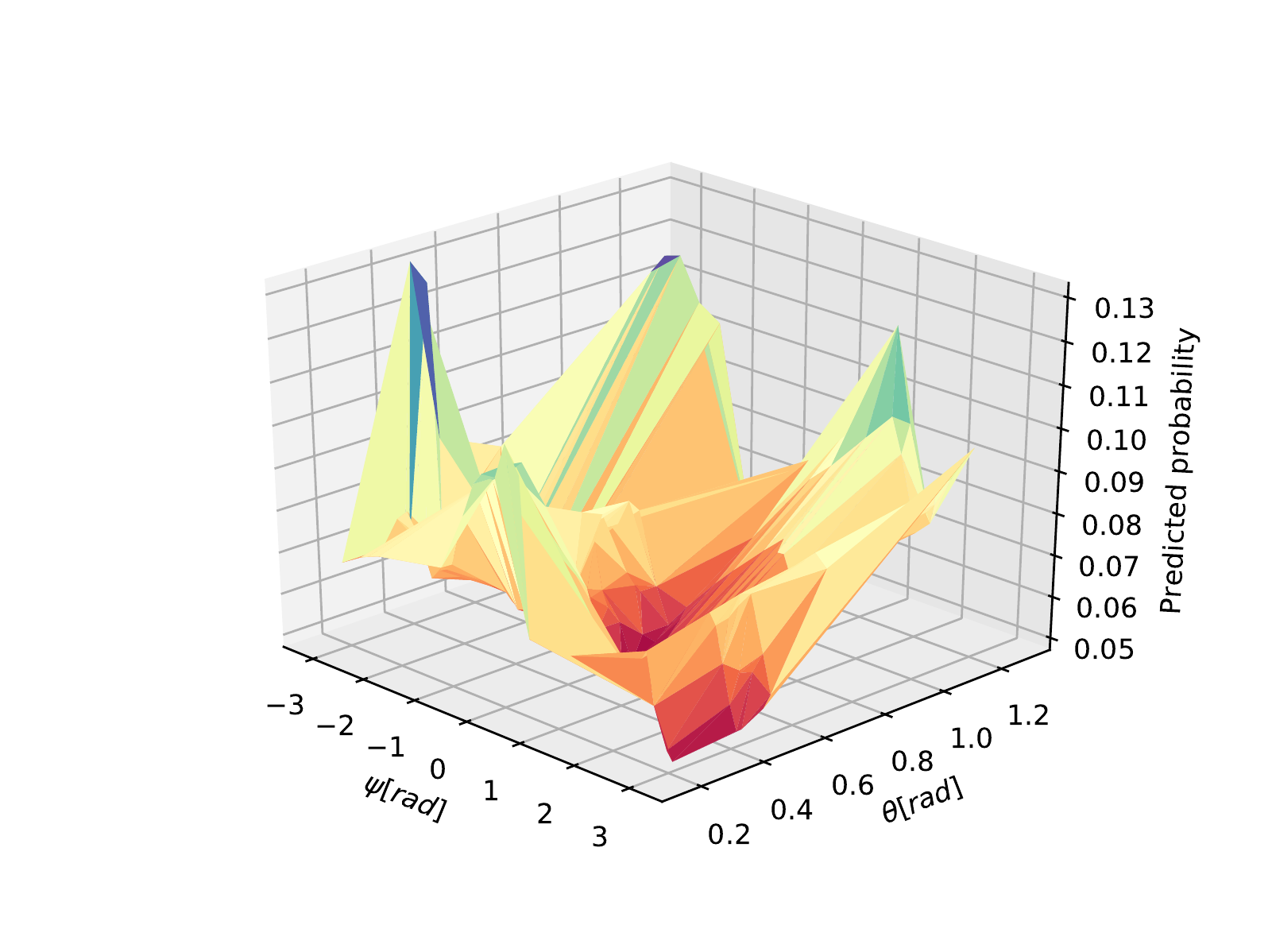}
		\caption{$c=3$ model, $c=2$ prob.}\label{fig:C3_2}
	\end{subfigure}
	
	\begin{subfigure}[b]{0.5\textwidth}
		\includegraphics[width=\textwidth]{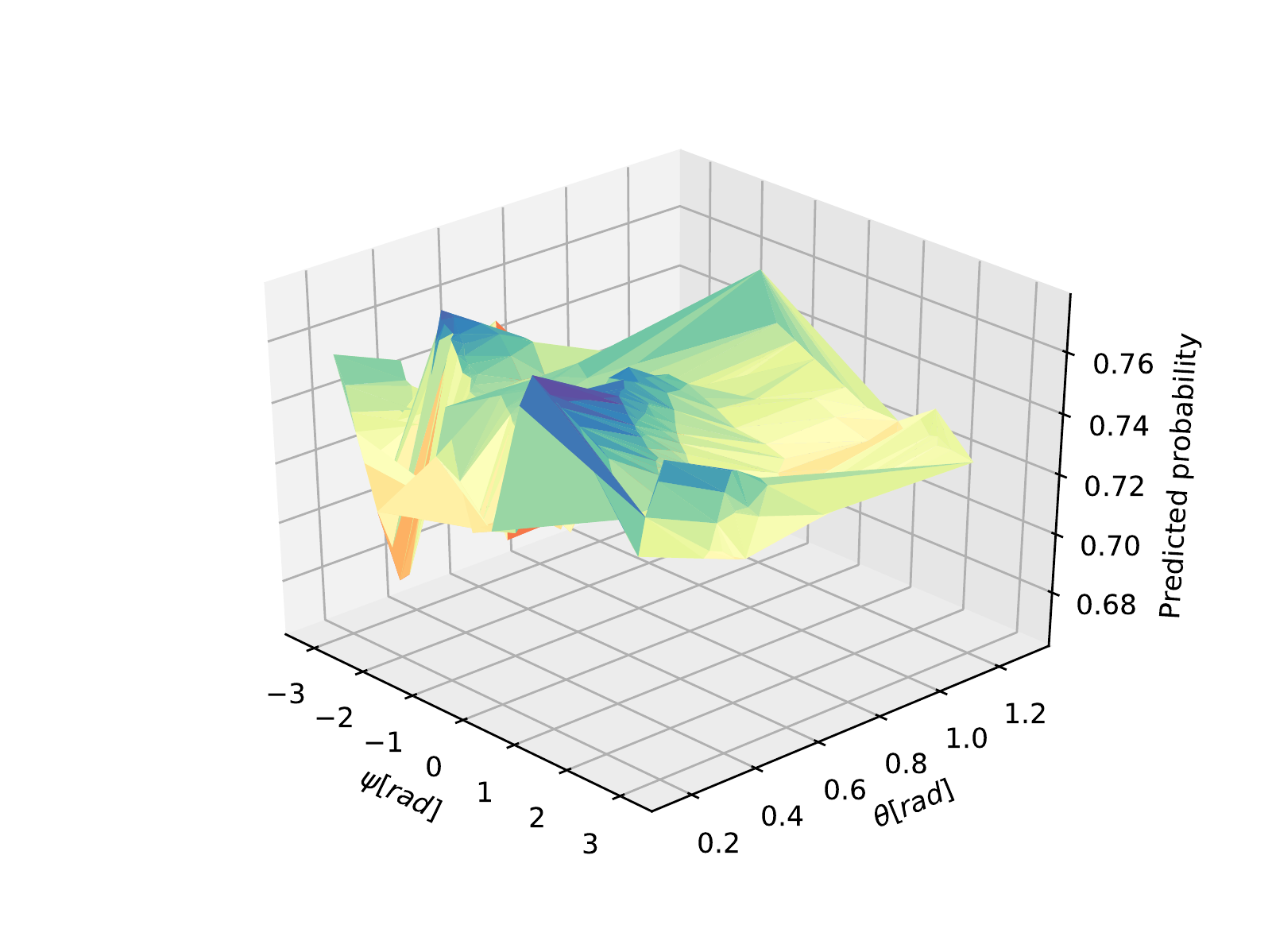}
		\caption{$c=3$ model, $c=3$ prob.}\label{fig:C3_3}
	\end{subfigure}
	\caption{Classifier model for $c=3$, trained on real images: probabilities of classes 1 to 3 depending on relative yaw and pitch angles presented i \textbf{(a)} to \textbf{(c)} respectively. Higher surfaces go have bluer color.}
	\label{fig:C3}
\end{figure}

Fig.~\ref{fig:Chairs} presents 4 of the images used in the experiment, with bounding boxes for the chairs.
Fig.~\ref{fig:C1}, Fig~\ref{fig:C2}, and~\ref{fig:C3} present the trained expected probability values for each relative $\psi$ and $\theta$ values, i.e. $\prob{z^{sem}|c=i,\psi,\theta}$ for each figure with different $i$. Each subfigure \textbf{(a)} to \textbf{(c)} representing measurement probability of class $c=1$ to $c=3$ respectively.

\clearpage

\section{Experiment: Table of Stack Time Stamps}

In this section we present a table of stack time stamps that indicates direct and indirect communication between robots in our scenario. Recall that the maximal communication radius is 3 meters, thus all robots start to communicate between them at step $k=6$.

\begin{multicols}{2}
\begin{tabular}{|l|l|l|l|}
	\hline
	Time step & Stack of $r_1$ & Stack of $r_2$ & Stack of $r_3$ \\
	\hline

	$k = 1$ &
	\makecell{t($r_1$): 1 \\ 	t($r_2$): 0 \\ 	t($r_3$): 0} &
	\makecell{t($r_1$): 0 \\ 	t($r_2$): 0 \\ 	t($r_3$): 0} &
	\makecell{t($r_1$): 0 \\ 	t($r_2$): 0 \\ 	t($r_3$): 1} \\ 
	\hline 
	
	$k = 2$ &
	\makecell{t($r_1$): 2 \\ 	t($r_2$): 0 \\ 	t($r_3$): 0} &
	\makecell{t($r_1$): 0 \\ 	t($r_2$): 0 \\ 	t($r_3$): 0} &
	\makecell{t($r_1$): 0 \\ 	t($r_2$): 0 \\ 	t($r_3$): 2} \\ 
	\hline 
	
	$k = 3$ &
	\makecell{t($r_1$): 3 \\ 	t($r_2$): 0 \\ 	t($r_3$): 0} &
	\makecell{t($r_1$): 0 \\ 	t($r_2$): 3 \\ 	t($r_3$): 0} &
	\makecell{t($r_1$): 0 \\ 	t($r_2$): 0 \\ 	t($r_3$): 3} \\ 
	\hline 
	
	$k = 4$ &
	\makecell{t($r_1$): 4 \\ 	t($r_2$): 0 \\ 	t($r_3$): 0} &
	\makecell{t($r_1$): 0 \\ 	t($r_2$): 4 \\ 	t($r_3$): 0} &
	\makecell{t($r_1$): 0 \\ 	t($r_2$): 0 \\ 	t($r_3$): 4} \\ 
	\hline 
	
	$k = 5$ &
	\makecell{t($r_1$): 5 \\ 	t($r_2$): 0 \\ 	t($r_3$): 0} &
	\makecell{t($r_1$): 0 \\ 	t($r_2$): 5 \\ 	t($r_3$): 0} &
	\makecell{t($r_1$): 0 \\ 	t($r_2$): 0 \\ 	t($r_3$): 5} \\ 
	\hline 
	
	$k = 6$ &
	\makecell{t($r_1$): 6 \\ 	t($r_2$): 5 \\ 	t($r_3$): 5} &
	\makecell{t($r_1$): 5 \\ 	t($r_2$): 6 \\ 	t($r_3$): 5} &
	\makecell{t($r_1$): 5 \\ 	t($r_2$): 5 \\ 	t($r_3$): 6} \\ 
	\hline 
	
	$k = 7$ &
	\makecell{t($r_1$): 7 \\ 	t($r_2$): 6 \\ 	t($r_3$): 6} &
	\makecell{t($r_1$): 6 \\ 	t($r_2$): 7 \\ 	t($r_3$): 6} &
	\makecell{t($r_1$): 6 \\ 	t($r_2$): 6 \\ 	t($r_3$): 7} \\ 
	\hline 
	
	$k = 8$ &
	\makecell{t($r_1$): 8 \\ 	t($r_2$): 7 \\ 	t($r_3$): 7} &
	\makecell{t($r_1$): 7 \\ 	t($r_2$): 8 \\ 	t($r_3$): 7} &
	\makecell{t($r_1$): 7 \\ 	t($r_2$): 7 \\ 	t($r_3$): 8} \\ 
	\hline 
	
	$k = 9$ &
	\makecell{t($r_1$): 9 \\ 	t($r_2$): 8 \\ 	t($r_3$): 8} &
	\makecell{t($r_1$): 8 \\ 	t($r_2$): 9 \\ 	t($r_3$): 8} &
	\makecell{t($r_1$): 8 \\ 	t($r_2$): 8 \\ 	t($r_3$): 9} \\ 
	\hline 
	
	$k = 10$ &
	\makecell{t($r_1$): 10 \\ 	t($r_2$): 9 \\ 	t($r_3$): 9} &
	\makecell{t($r_1$): 9 \\ 	t($r_2$): 10 \\ 	t($r_3$): 9} &
	\makecell{t($r_1$): 9 \\ 	t($r_2$): 9 \\ 	t($r_3$): 10} \\ 
	\hline 
	
	$k = 11$ &
	\makecell{t($r_1$): 11 \\ 	t($r_2$): 10 \\ 	t($r_3$): 10} &
	\makecell{t($r_1$): 10 \\ 	t($r_2$): 11 \\ 	t($r_3$): 10} &
	\makecell{t($r_1$): 10 \\ 	t($r_2$): 10 \\ 	t($r_3$): 11} \\ 
	\hline 
	
	$k = 12$ &
	\makecell{t($r_1$): 12 \\ 	t($r_2$): 11 \\ 	t($r_3$): 11} &
	\makecell{t($r_1$): 11 \\ 	t($r_2$): 12 \\ 	t($r_3$): 11} &
	\makecell{t($r_1$): 11 \\ 	t($r_2$): 11 \\ 	t($r_3$): 12} \\ 
	\hline 
	
	$k = 13$ &
	\makecell{t($r_1$): 13 \\ 	t($r_2$): 12 \\ 	t($r_3$): 12} &
	\makecell{t($r_1$): 12 \\ 	t($r_2$): 13 \\ 	t($r_3$): 12} &
	\makecell{t($r_1$): 12 \\ 	t($r_2$): 12 \\ 	t($r_3$): 13} \\ 
	\hline 
	
	$k = 14$ &
	\makecell{t($r_1$): 14 \\ 	t($r_2$): 13 \\ 	t($r_3$): 13} &
	\makecell{t($r_1$): 13 \\ 	t($r_2$): 14 \\ 	t($r_3$): 13} &
	\makecell{t($r_1$): 13 \\ 	t($r_2$): 13 \\ 	t($r_3$): 14} \\ 
	\hline 
	
	$k = 15$ &
	\makecell{t($r_1$): 15 \\ 	t($r_2$): 14 \\ 	t($r_3$): 14} &
	\makecell{t($r_1$): 14 \\ 	t($r_2$): 15 \\ 	t($r_3$): 14} &
	\makecell{t($r_1$): 14 \\ 	t($r_2$): 14 \\ 	t($r_3$): 15} \\ 
	\hline 

\end{tabular}

\begin{tabular}{|l|l|l|l|}

	\hline
	Time step & Stack of $r_1$ & Stack of $r_2$ & Stack of $r_3$ \\
	\hline
	
	$k = 16$ &
	\makecell{t($r_1$): 16 \\ 	t($r_2$): 15 \\ 	t($r_3$): 15} &
	\makecell{t($r_1$): 15 \\ 	t($r_2$): 16 \\ 	t($r_3$): 15} &
	\makecell{t($r_1$): 15 \\ 	t($r_2$): 15 \\ 	t($r_3$): 16} \\ 
	\hline 
	
	$k = 17$ &
	\makecell{t($r_1$): 17 \\ 	t($r_2$): 16 \\ 	t($r_3$): 16} &
	\makecell{t($r_1$): 16 \\ 	t($r_2$): 17 \\ 	t($r_3$): 16} &
	\makecell{t($r_1$): 16 \\ 	t($r_2$): 16 \\ 	t($r_3$): 17} \\ 
	\hline 
	
	$k = 18$ &
	\makecell{t($r_1$): 18 \\ 	t($r_2$): 17 \\ 	t($r_3$): 17} &
	\makecell{t($r_1$): 17 \\ 	t($r_2$): 18 \\ 	t($r_3$): 17} &
	\makecell{t($r_1$): 17 \\ 	t($r_2$): 17 \\ 	t($r_3$): 18} \\ 
	\hline 
	
	$k = 19$ &
	\makecell{t($r_1$): 19 \\ 	t($r_2$): 18 \\ 	t($r_3$): 18} &
	\makecell{t($r_1$): 18 \\ 	t($r_2$): 19 \\ 	t($r_3$): 18} &
	\makecell{t($r_1$): 18 \\ 	t($r_2$): 18 \\ 	t($r_3$): 19} \\ 
	\hline 
	
	$k = 20$ &
	\makecell{t($r_1$): 20 \\ 	t($r_2$): 19 \\ 	t($r_3$): 19} &
	\makecell{t($r_1$): 19 \\ 	t($r_2$): 20 \\ 	t($r_3$): 19} &
	\makecell{t($r_1$): 19 \\ 	t($r_2$): 19 \\ 	t($r_3$): 20} \\ 
	\hline 
	
	$k = 21$ &
	\makecell{t($r_1$): 21 \\ 	t($r_2$): 19 \\ 	t($r_3$): 19} &
	\makecell{t($r_1$): 19 \\ 	t($r_2$): 21 \\ 	t($r_3$): 20} &
	\makecell{t($r_1$): 19 \\ 	t($r_2$): 20 \\ 	t($r_3$): 21} \\ 
	\hline 
	
	$k = 22$ &
	\makecell{t($r_1$): 22 \\ 	t($r_2$): 19 \\ 	t($r_3$): 19} &
	\makecell{t($r_1$): 19 \\ 	t($r_2$): 22 \\ 	t($r_3$): 21} &
	\makecell{t($r_1$): 19 \\ 	t($r_2$): 21 \\ 	t($r_3$): 22} \\ 
	\hline 
	
	$k = 23$ &
	\makecell{t($r_1$): 23 \\ 	t($r_2$): 19 \\ 	t($r_3$): 22} &
	\makecell{t($r_1$): 21 \\ 	t($r_2$): 23 \\ 	t($r_3$): 21} &
	\makecell{t($r_1$): 22 \\ 	t($r_2$): 21 \\ 	t($r_3$): 23} \\ 
	\hline 
	
	$k = 24$ &
	\makecell{t($r_1$): 24 \\ 	t($r_2$): 19 \\ 	t($r_3$): 23} &
	\makecell{t($r_1$): 21 \\ 	t($r_2$): 24 \\ 	t($r_3$): 21} &
	\makecell{t($r_1$): 23 \\ 	t($r_2$): 21 \\ 	t($r_3$): 24} \\ 
	\hline 
	
	$k = 25$ &
	\makecell{t($r_1$): 25 \\ 	t($r_2$): 23 \\ 	t($r_3$): 24} &
	\makecell{t($r_1$): 21 \\ 	t($r_2$): 25 \\ 	t($r_3$): 24} &
	\makecell{t($r_1$): 24 \\ 	t($r_2$): 24 \\ 	t($r_3$): 25} \\ 
	\hline 
	
	$k = 26$ &
	\makecell{t($r_1$): 26 \\ 	t($r_2$): 24 \\ 	t($r_3$): 25} &
	\makecell{t($r_1$): 24 \\ 	t($r_2$): 26 \\ 	t($r_3$): 25} &
	\makecell{t($r_1$): 25 \\ 	t($r_2$): 25 \\ 	t($r_3$): 26} \\ 
	\hline 
	
	$k = 27$ &
	\makecell{t($r_1$): 27 \\ 	t($r_2$): 25 \\ 	t($r_3$): 26} &
	\makecell{t($r_1$): 25 \\ 	t($r_2$): 27 \\ 	t($r_3$): 26} &
	\makecell{t($r_1$): 26 \\ 	t($r_2$): 26 \\ 	t($r_3$): 27} \\ 
	\hline 
	
	$k = 28$ &
	\makecell{t($r_1$): 28 \\ 	t($r_2$): 26 \\ 	t($r_3$): 27} &
	\makecell{t($r_1$): 26 \\ 	t($r_2$): 28 \\ 	t($r_3$): 27} &
	\makecell{t($r_1$): 27 \\ 	t($r_2$): 27 \\ 	t($r_3$): 28} \\ 
	\hline 
	
	$k = 29$ &
	\makecell{t($r_1$): 29 \\ 	t($r_2$): 27 \\ 	t($r_3$): 28} &
	\makecell{t($r_1$): 27 \\ 	t($r_2$): 29 \\ 	t($r_3$): 28} &
	\makecell{t($r_1$): 28 \\ 	t($r_2$): 28 \\ 	t($r_3$): 29} \\ 
	\hline 
	
	$k = 30$ &
	\makecell{t($r_1$): 30 \\ 	t($r_2$): 29 \\ 	t($r_3$): 29} &
	\makecell{t($r_1$): 29 \\ 	t($r_2$): 30 \\ 	t($r_3$): 29} &
	\makecell{t($r_1$): 29 \\ 	t($r_2$): 29 \\ 	t($r_3$): 30} \\ 
	\hline 
	
\end{tabular}
\end{multicols}

\begin{tabular}{|l|l|l|l|}
	
	\hline
	Time step & Stack of $r_1$ & Stack of $r_2$ & Stack of $r_3$ \\
	\hline
	
$k = 31$ &
	\makecell{t($r_1$): 31 \\ 	t($r_2$): 30 \\ 	t($r_3$): 30} &
	\makecell{t($r_1$): 30 \\ 	t($r_2$): 31 \\ 	t($r_3$): 30} &
	\makecell{t($r_1$): 30 \\ 	t($r_2$): 30 \\ 	t($r_3$): 31} \\ 
	\hline 
	
	$k = 32$ &
	\makecell{t($r_1$): 32 \\ 	t($r_2$): 31 \\ 	t($r_3$): 31} &
	\makecell{t($r_1$): 31 \\ 	t($r_2$): 32 \\ 	t($r_3$): 31} &
	\makecell{t($r_1$): 31 \\ 	t($r_2$): 31 \\ 	t($r_3$): 32} \\ 
	\hline 
	
	$k = 33$ &
	\makecell{t($r_1$): 33 \\ 	t($r_2$): 32 \\ 	t($r_3$): 32} &
	\makecell{t($r_1$): 32 \\ 	t($r_2$): 33 \\ 	t($r_3$): 32} &
	\makecell{t($r_1$): 32 \\ 	t($r_2$): 32 \\ 	t($r_3$): 33} \\ 
	\hline 
	
	$k = 34$ &
	\makecell{t($r_1$): 34 \\ 	t($r_2$): 33 \\ 	t($r_3$): 33} &
	\makecell{t($r_1$): 33 \\ 	t($r_2$): 34 \\ 	t($r_3$): 33} &
	\makecell{t($r_1$): 33 \\ 	t($r_2$): 33 \\ 	t($r_3$): 34} \\ 
	\hline 
	
	$k = 35$ &
	\makecell{t($r_1$): 35 \\ 	t($r_2$): 34 \\ 	t($r_3$): 34} &
	\makecell{t($r_1$): 34 \\ 	t($r_2$): 35 \\ 	t($r_3$): 34} &
	\makecell{t($r_1$): 34 \\ 	t($r_2$): 34 \\ 	t($r_3$): 35} \\ 
	\hline 
	
	$k = 36$ &
	\makecell{t($r_1$): 36 \\ 	t($r_2$): 35 \\ 	t($r_3$): 35} &
	\makecell{t($r_1$): 35 \\ 	t($r_2$): 36 \\ 	t($r_3$): 35} &
	\makecell{t($r_1$): 35 \\ 	t($r_2$): 35 \\ 	t($r_3$): 36} \\ 
	\hline 
	
	$k = 37$ &
	\makecell{t($r_1$): 37 \\ 	t($r_2$): 36 \\ 	t($r_3$): 36} &
	\makecell{t($r_1$): 36 \\ 	t($r_2$): 37 \\ 	t($r_3$): 36} &
	\makecell{t($r_1$): 36 \\ 	t($r_2$): 36 \\ 	t($r_3$): 37} \\ 
	\hline 
	
	$k = 38$ &
	\makecell{t($r_1$): 38 \\ 	t($r_2$): 37 \\ 	t($r_3$): 37} &
	\makecell{t($r_1$): 37 \\ 	t($r_2$): 38 \\ 	t($r_3$): 37} &
	\makecell{t($r_1$): 37 \\ 	t($r_2$): 37 \\ 	t($r_3$): 38} \\ 
	\hline 
	
	$k = 39$ &
	\makecell{t($r_1$): 39 \\ 	t($r_2$): 38 \\ 	t($r_3$): 38} &
	\makecell{t($r_1$): 38 \\ 	t($r_2$): 39 \\ 	t($r_3$): 38} &
	\makecell{t($r_1$): 38 \\ 	t($r_2$): 38 \\ 	t($r_3$): 39} \\ 
	\hline 
	
	$k = 40$ &
	\makecell{t($r_1$): 40 \\ 	t($r_2$): 39 \\ 	t($r_3$): 39} &
	\makecell{t($r_1$): 39 \\ 	t($r_2$): 40 \\ 	t($r_3$): 39} &
	\makecell{t($r_1$): 39 \\ 	t($r_2$): 39 \\ 	t($r_3$): 40} \\ 
	\hline 
\end{tabular}

\newpage
\section{Experiment: Additional Results}

In this section we present additional results for the experiment. We show the beliefs at various stages of the path.

\begin{figure}[!htbp]
	
	\begin{subfigure}[b]{0.24\textwidth}
		\includegraphics[width=\textwidth]{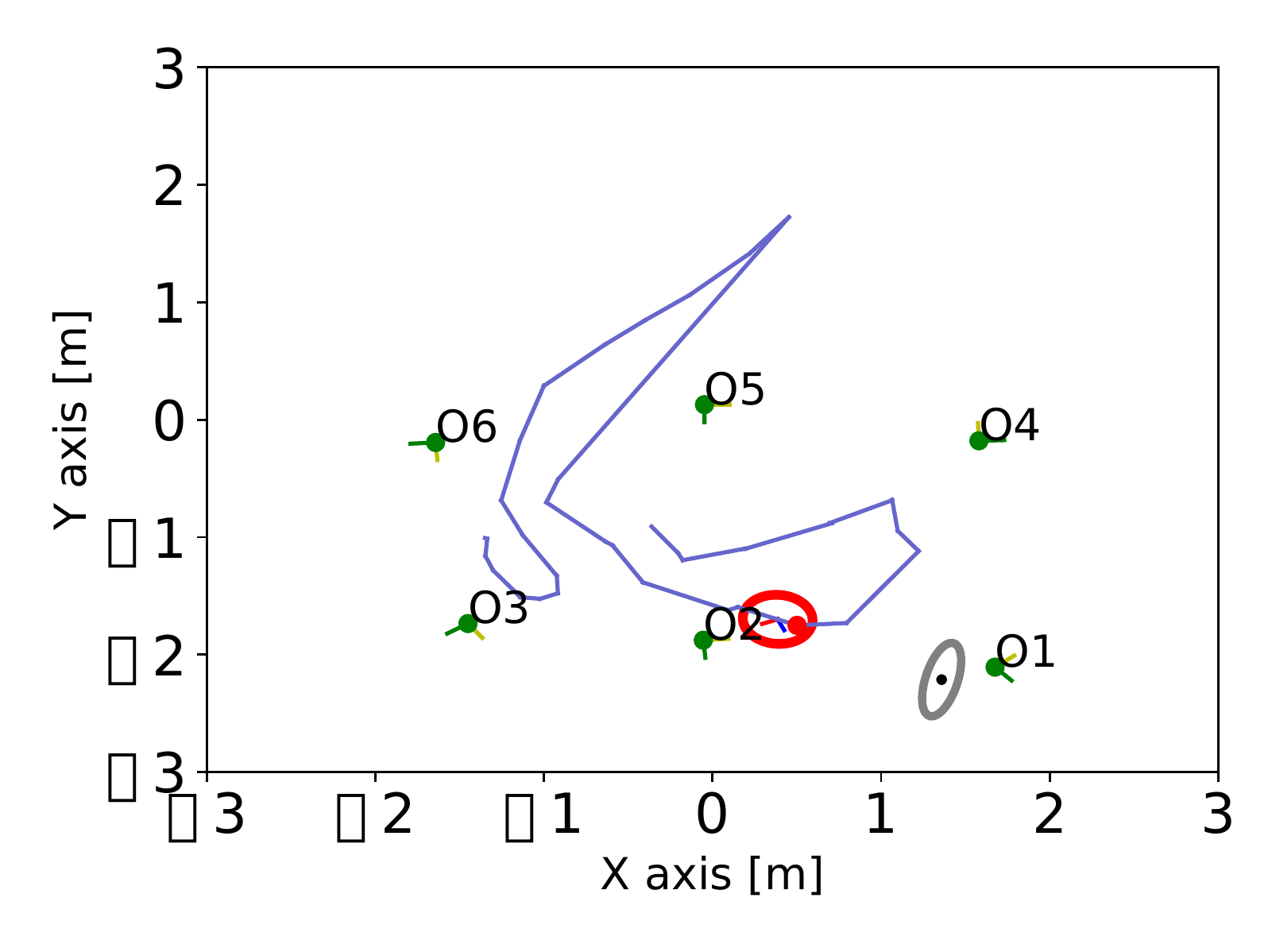}
		\caption{Local SLAM $r_3$}\label{fig:real_Robot_3_x15}
	\end{subfigure}
	\begin{subfigure}[b]{0.24\textwidth}
		\includegraphics[width=\textwidth]{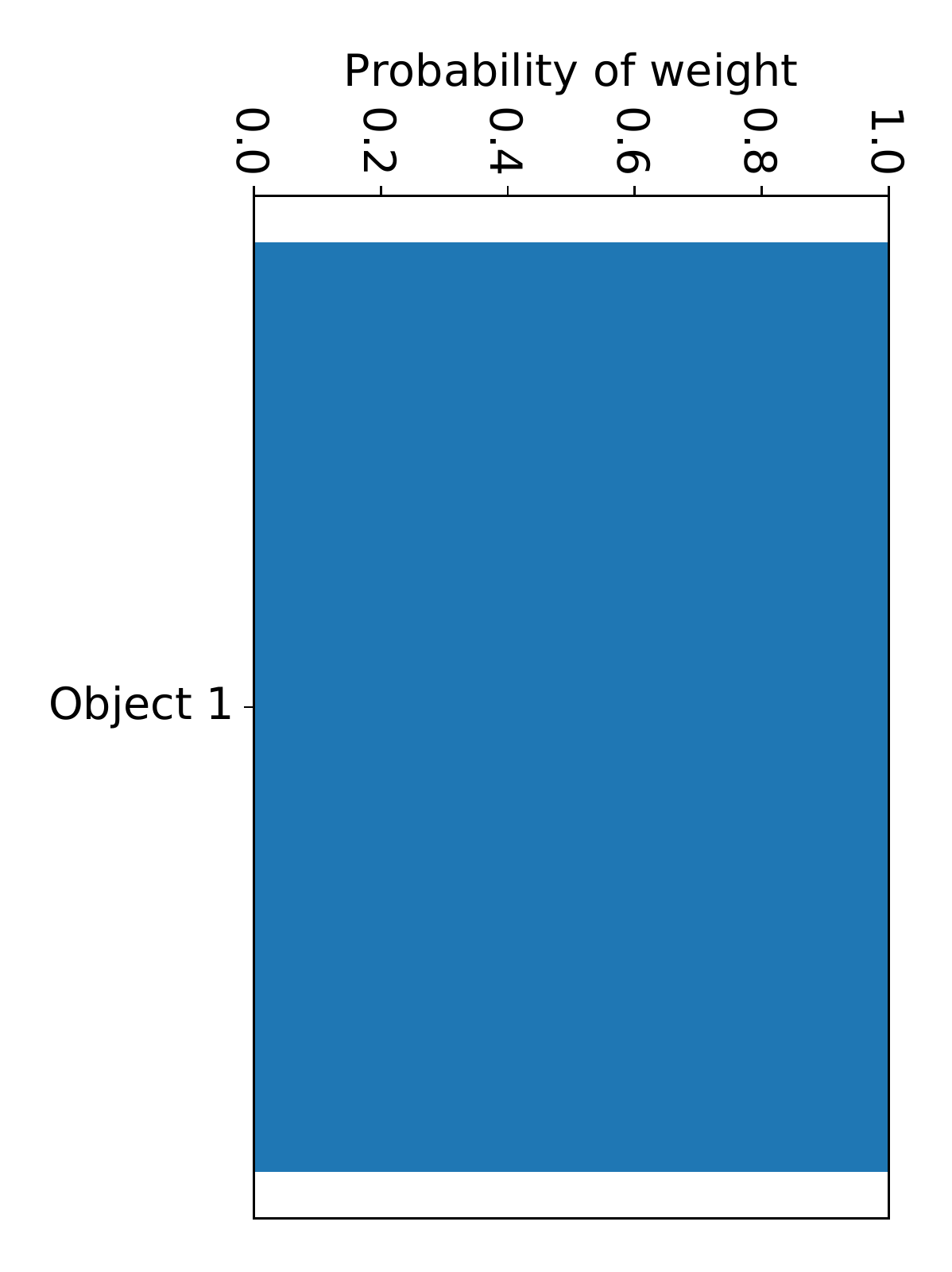}
		\caption{Local classification $r_3$}\label{fig:real_Robot_3_x15_cls}
	\end{subfigure}
	\begin{subfigure}[b]{0.24\textwidth}
		\includegraphics[width=\textwidth]{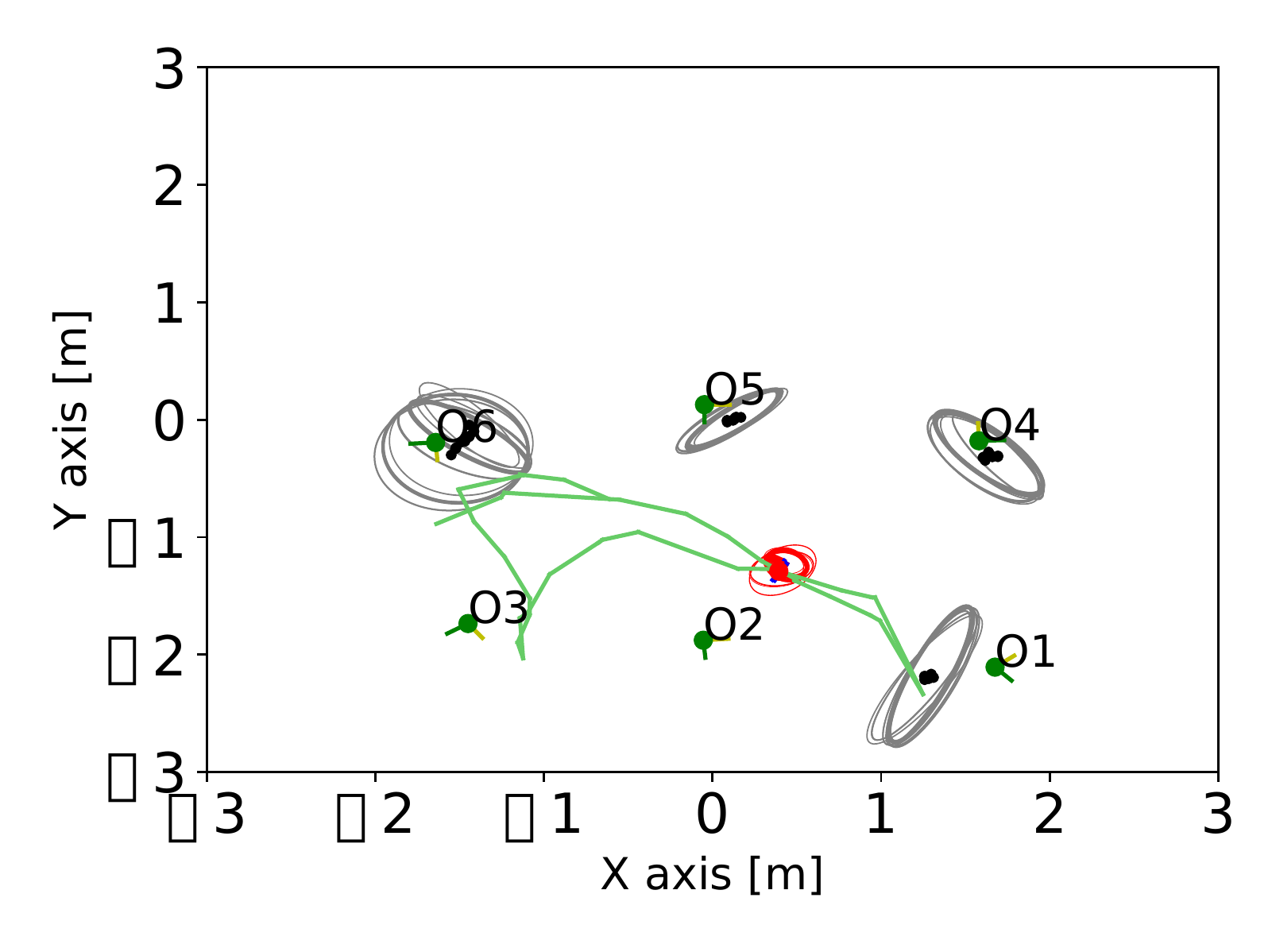}
		\caption{Local SLAM $r_2$}\label{fig:real_Robot_2_x20}
	\end{subfigure}
	\begin{subfigure}[b]{0.24\textwidth}
		\includegraphics[width=\textwidth]{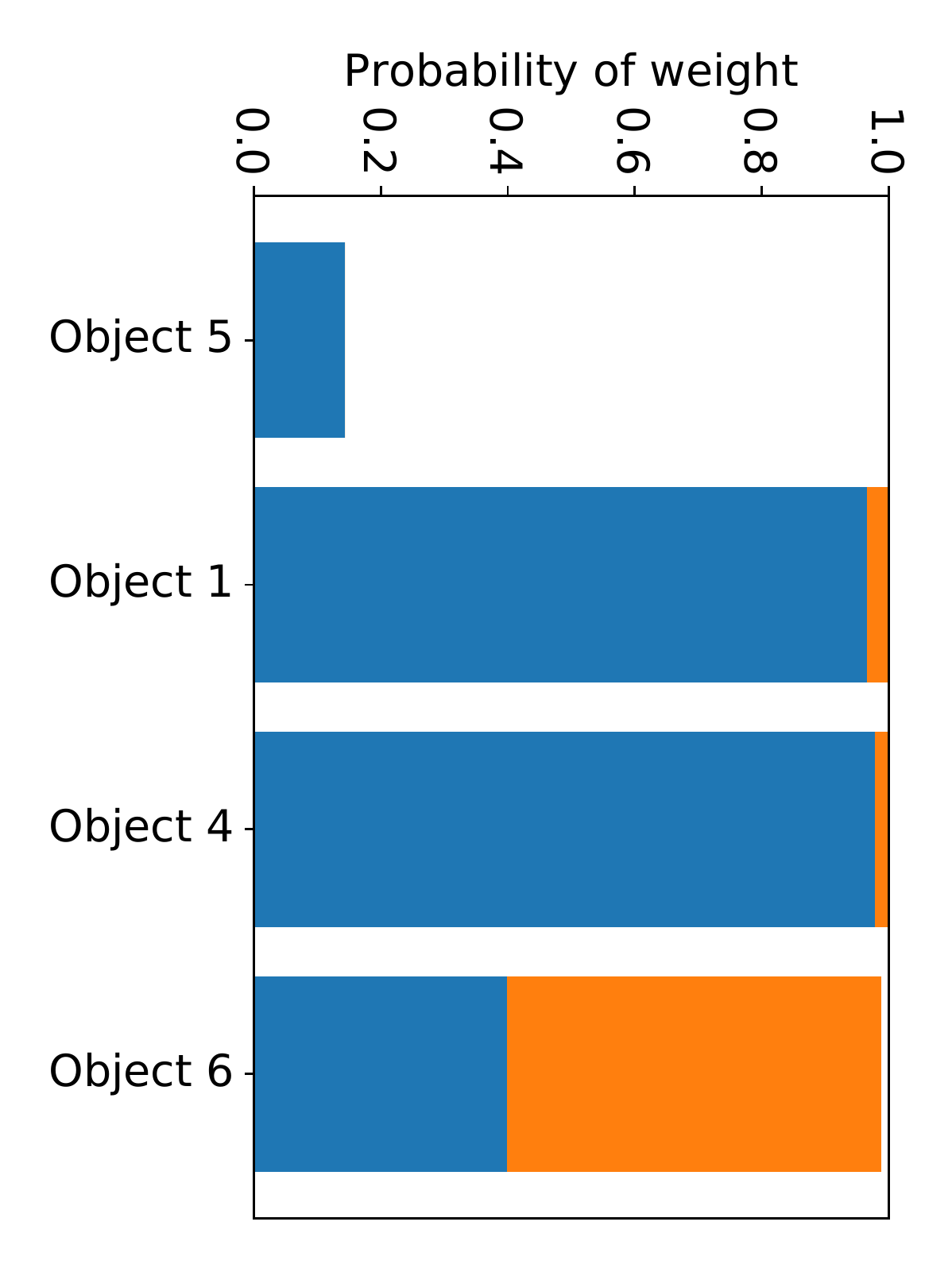}
		\caption{Local classification $r_2$}\label{fig:real_Robot_2_x20_cls}
	\end{subfigure}
	
	\caption{Figures for robot $r_3$ and $r_2$, local beliefs for time $k=15$ and $k=20$ respectively. \textbf{(a)} and \textbf{(b)} show results for $r_3$, \textbf{(c)} and \textbf{(d)} for $r_2$. \textbf{(a)} and \textbf{(c)} present SLAM results, \textbf{(b)} and \textbf{(d)} present classification results.}
	\label{fig:Figures_real_local_1}
\end{figure}

\begin{figure}[!htbp]
	
	\begin{subfigure}[b]{0.24\textwidth}
		\includegraphics[width=\textwidth]{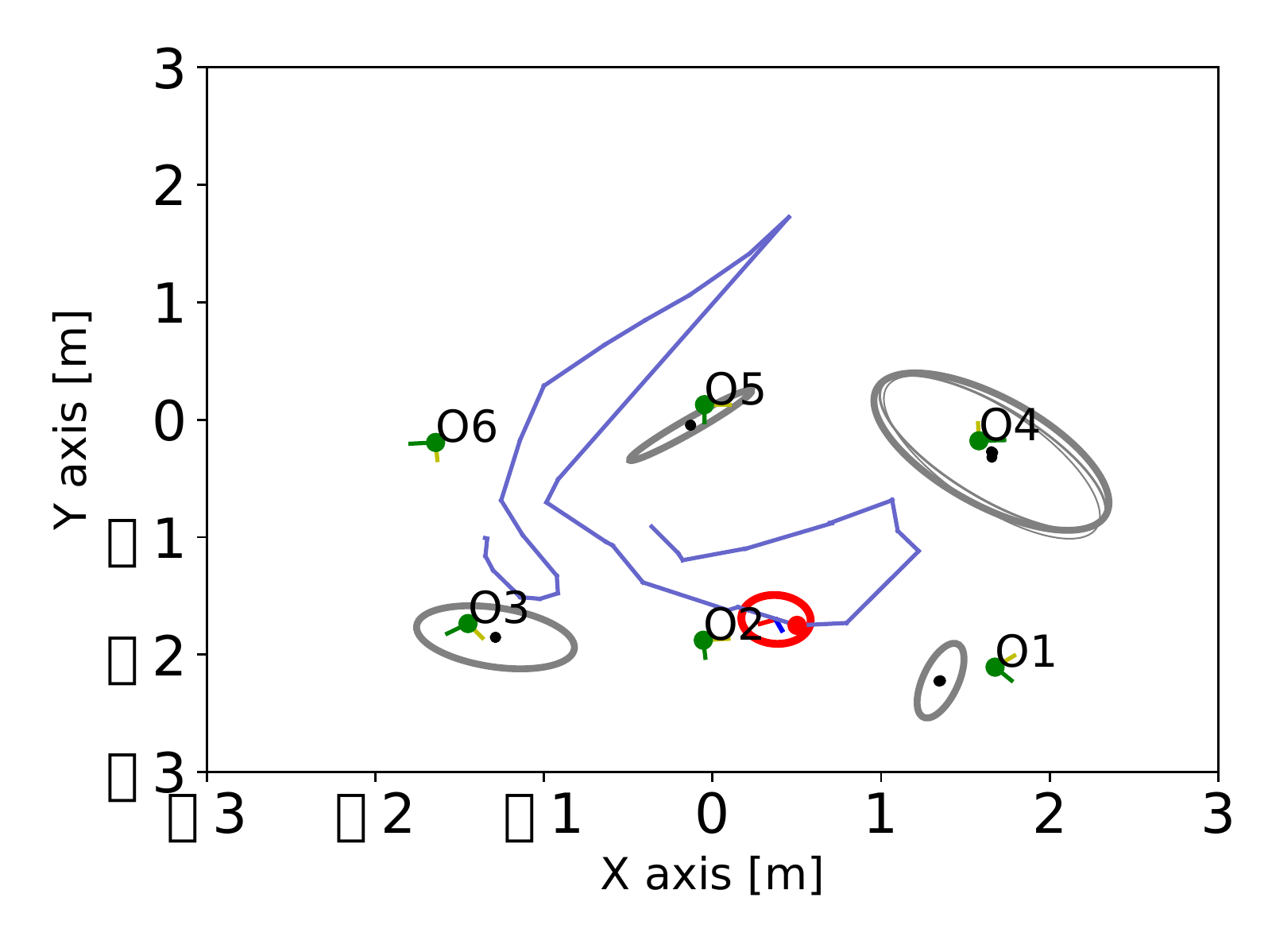}
		\caption{Dis. SLAM $r_3$}\label{fig:real_d_Robot_3_x15}
	\end{subfigure}
	\begin{subfigure}[b]{0.24\textwidth}
		\includegraphics[width=\textwidth]{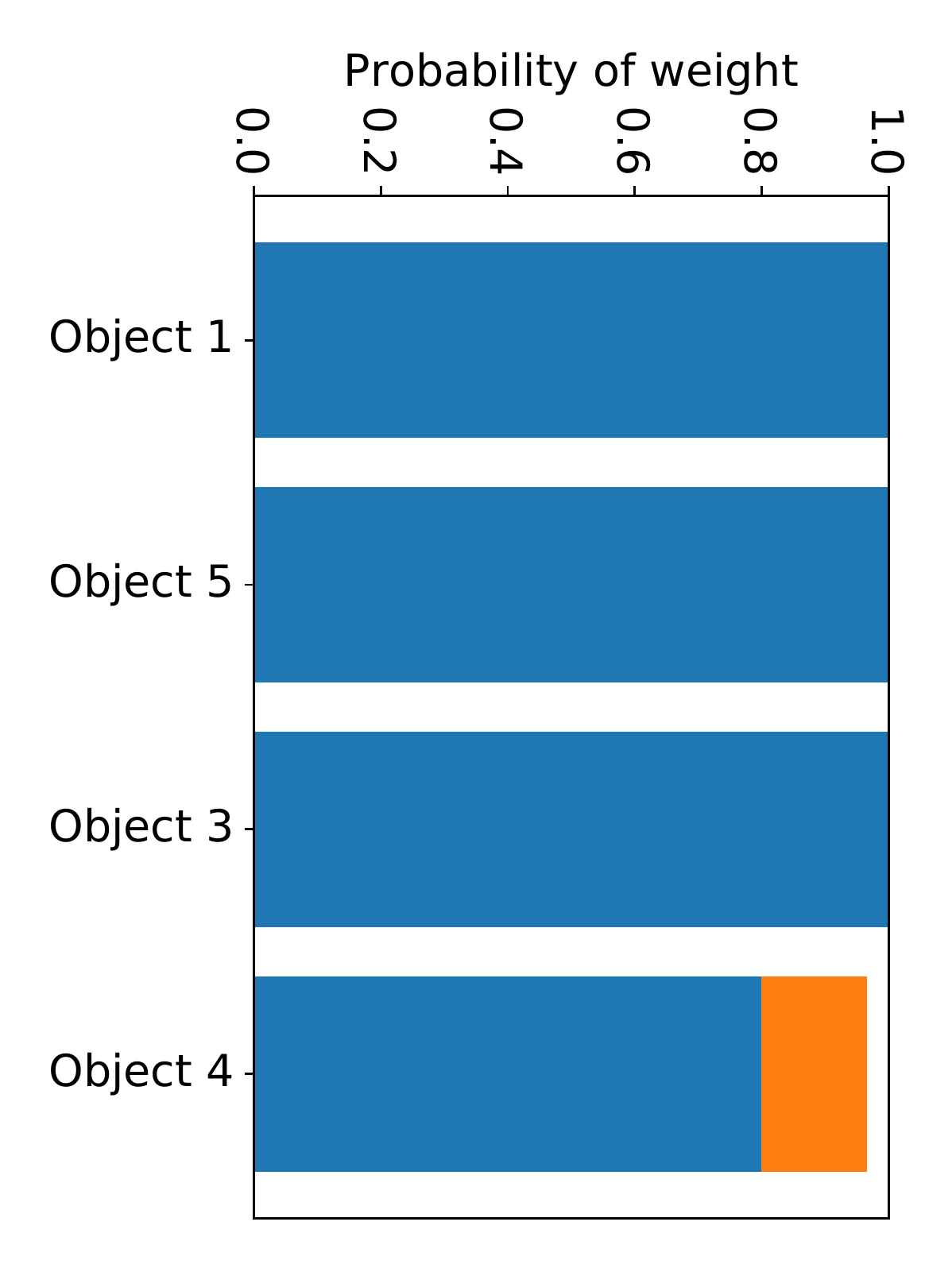}
		\caption{Dis. classification $r_3$}\label{fig:real_d_Robot_3_x15_cls}
	\end{subfigure}
	\begin{subfigure}[b]{0.24\textwidth}
		\includegraphics[width=\textwidth]{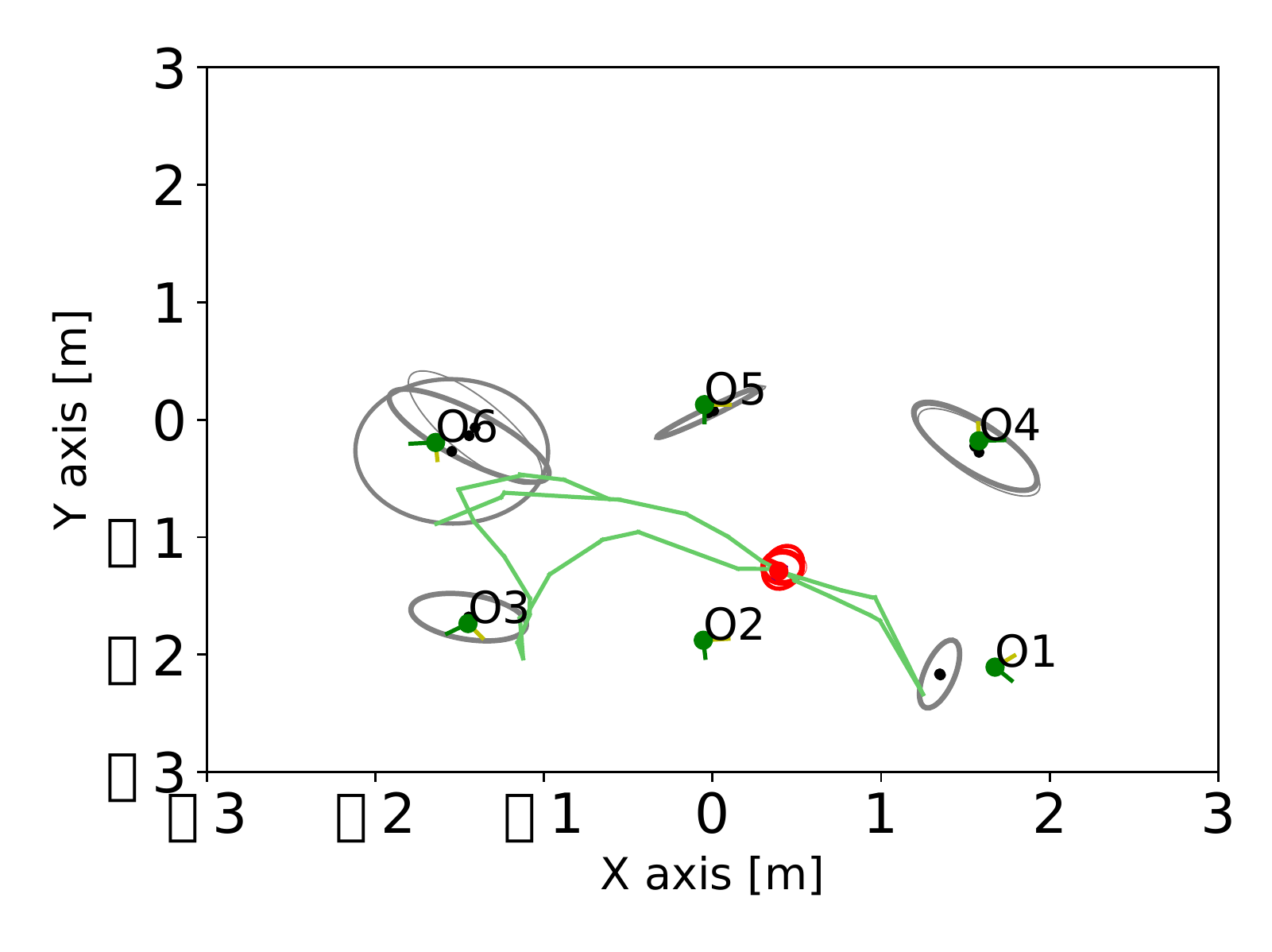}
		\caption{Dis. SLAM $r_2$}\label{fig:real_d_Robot_2_x20}
	\end{subfigure}
	\begin{subfigure}[b]{0.24\textwidth}
		\includegraphics[width=\textwidth]{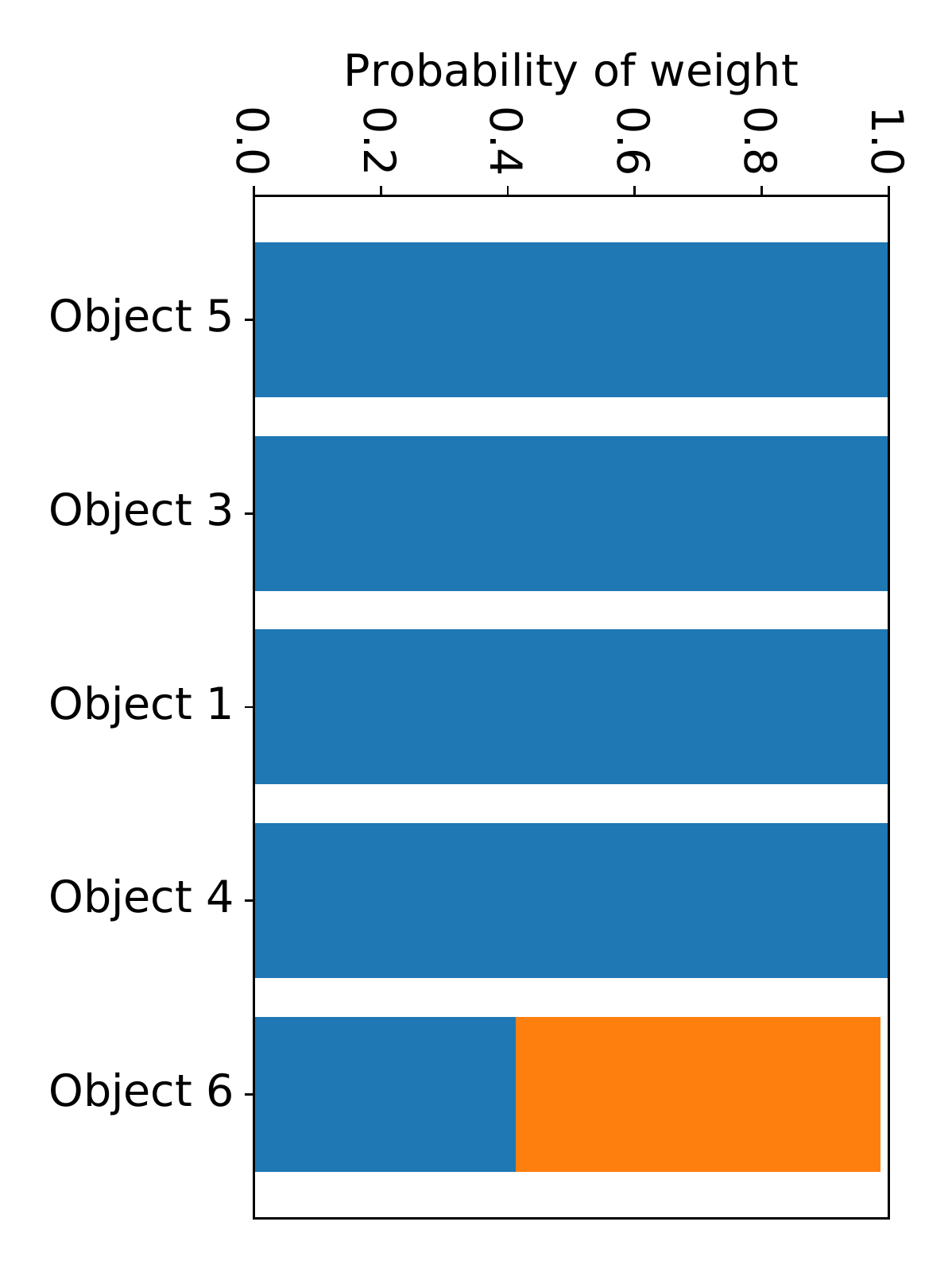}
		\caption{Dis. classification $r_2$}\label{fig:real_d_Robot_2_x20_cls}
	\end{subfigure}
	
	\caption{Figures for robot $r_2$ and $r_1$, distributed beliefs for time $k=15$ and $k=20$ respectively. \textbf{(a)} and \textbf{(b)} show results for $r_3$, \textbf{(c)} and \textbf{(d)} for $r_2$. \textbf{(a)} and \textbf{(c)} present SLAM results, \textbf{(b)} and \textbf{(d)} present classification results.}
	\label{fig:Figures_real_d_1}
\end{figure}

\begin{figure}[!htbp]
	
	\begin{subfigure}[b]{0.24\textwidth}
		\includegraphics[width=\textwidth]{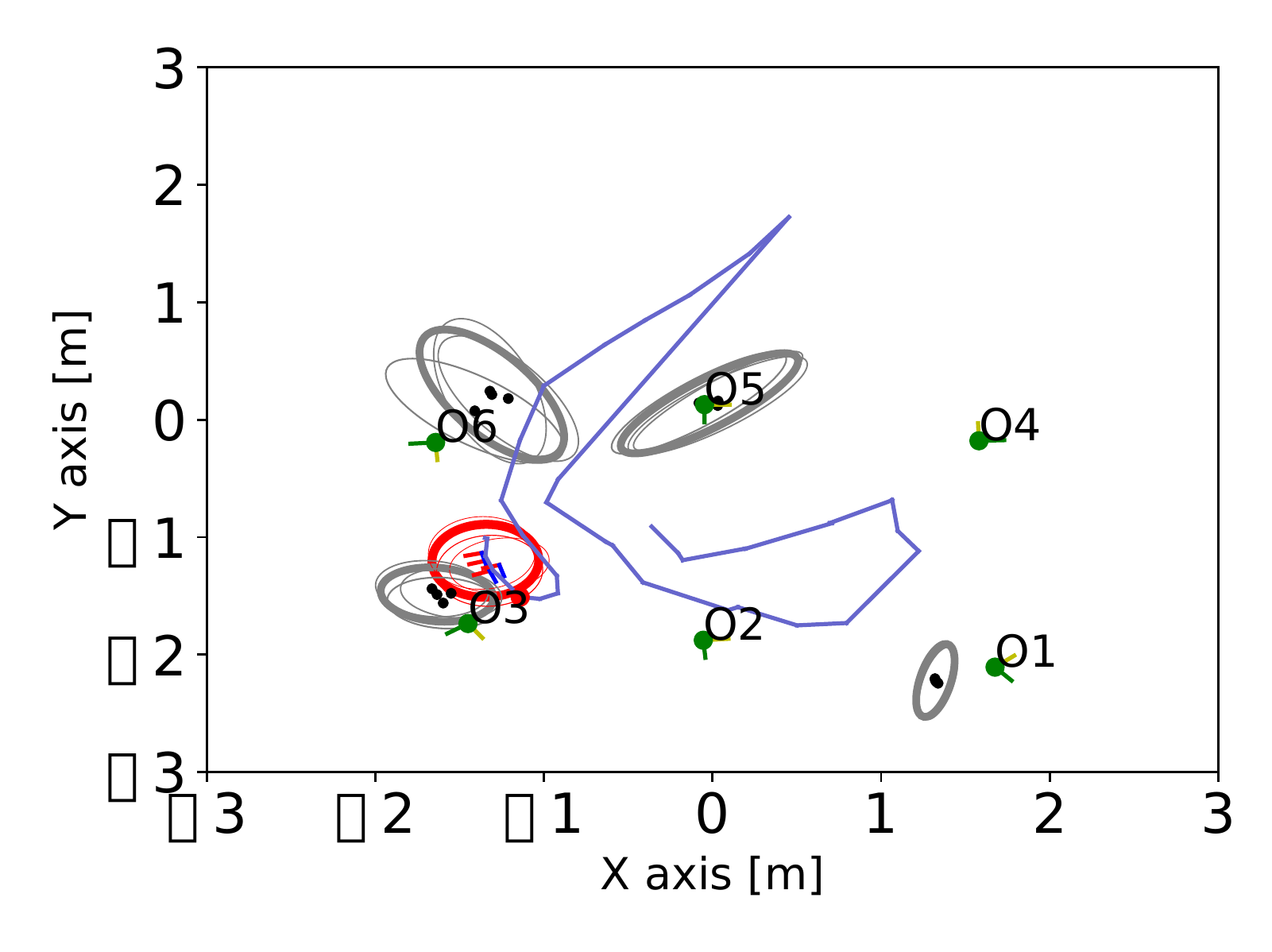}
		\caption{Local SLAM $r_1$}\label{fig:real_Robot_3_x35}
	\end{subfigure}
	\begin{subfigure}[b]{0.24\textwidth}
		\includegraphics[width=\textwidth]{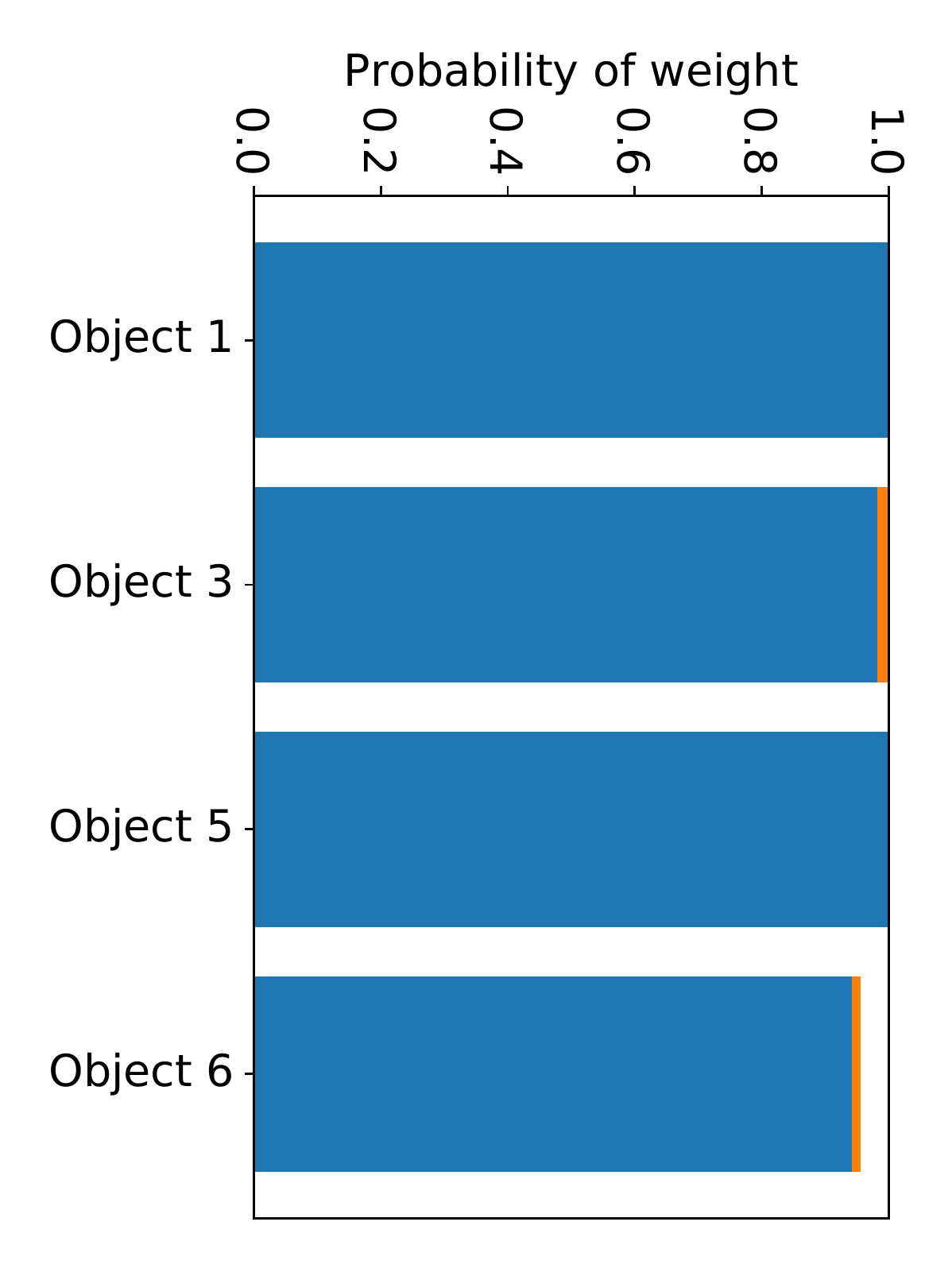}
		\caption{Local classification $r_1$}\label{fig:real_Robot_3_x35_cls}
	\end{subfigure}
	\begin{subfigure}[b]{0.24\textwidth}
		\includegraphics[width=\textwidth]{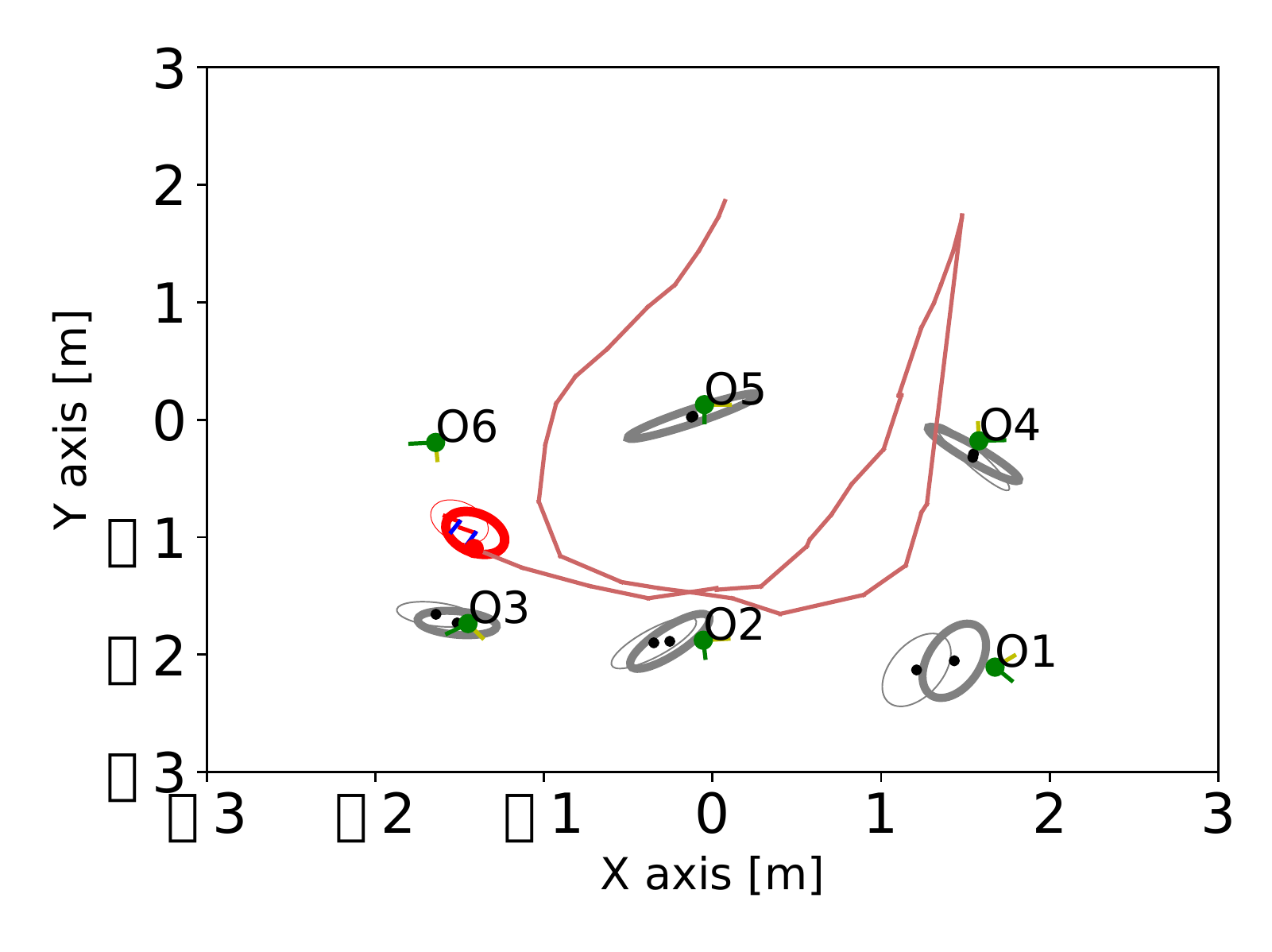}
		\caption{Local SLAM $r_3$}\label{fig:real_Robot_1_x40}
	\end{subfigure}
	\begin{subfigure}[b]{0.24\textwidth}
		\includegraphics[width=\textwidth]{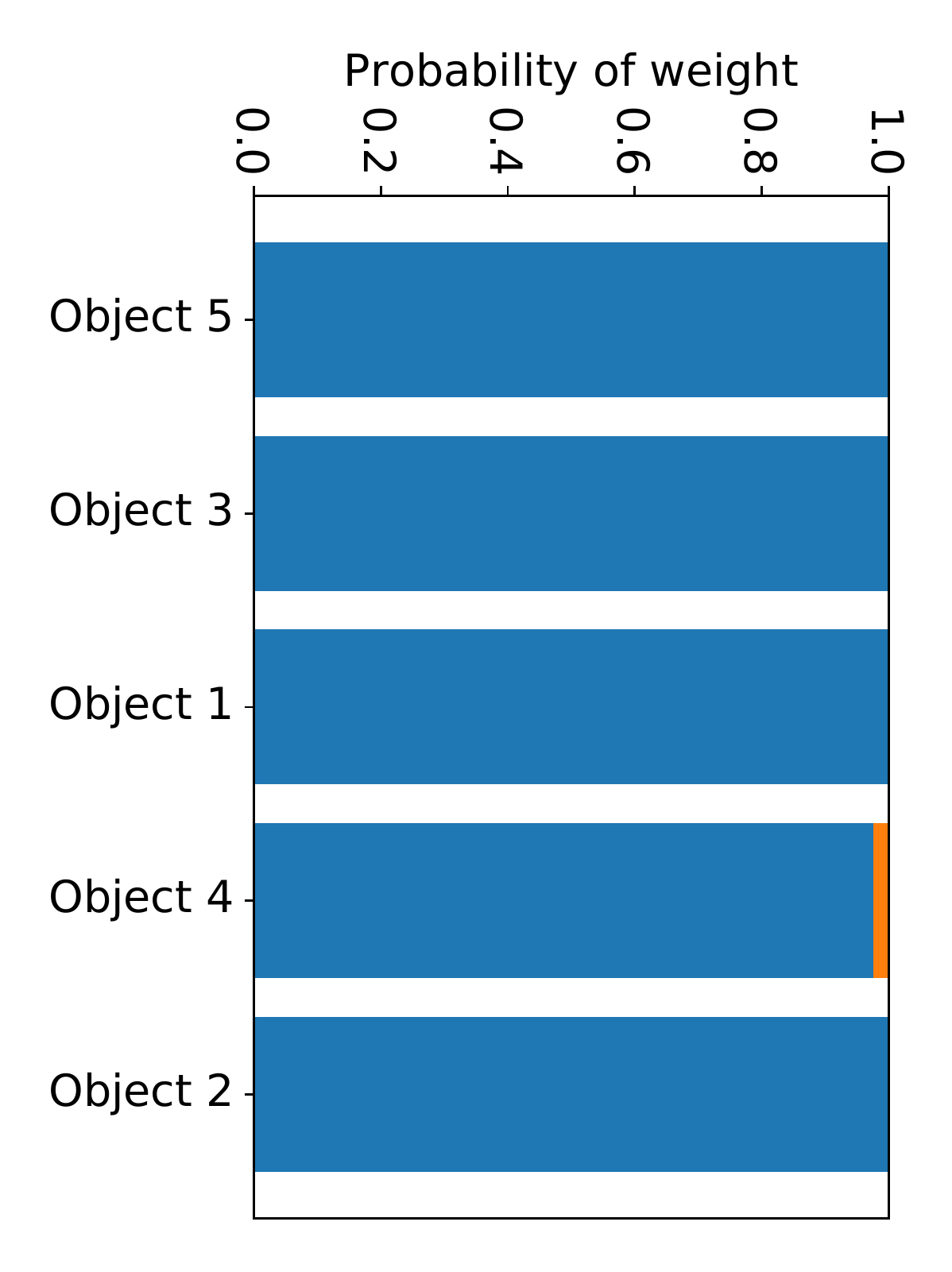}
		\caption{Local classification $r_3$}\label{fig:real_Robot_1_x40_cls}
	\end{subfigure}
	
	\caption{Figures for robot $r_3$ and $r_1$, local beliefs for time $k=35$ and $k=40$ respectively. \textbf{(a)} and \textbf{(b)} show results for $r_3$, \textbf{(c)} and \textbf{(d)} for $r_1$. \textbf{(a)} and \textbf{(c)} present SLAM results, \textbf{(b)} and \textbf{(d)} present classification results.}
	\label{fig:Figures_real_local_2}
\end{figure}

\begin{figure}[!htbp]
	
	\begin{subfigure}[b]{0.24\textwidth}
		\includegraphics[width=\textwidth]{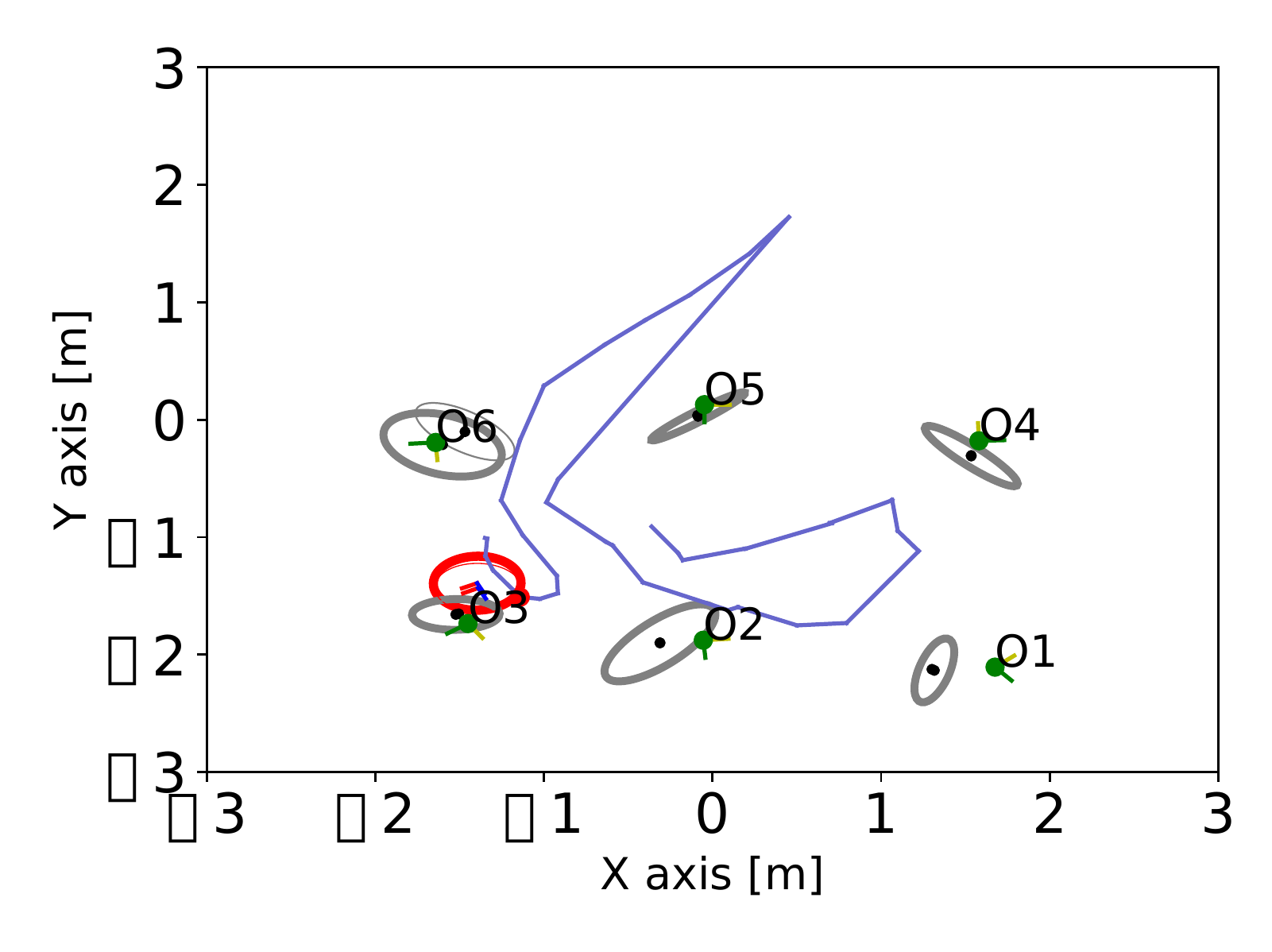}
		\caption{Dis. SLAM $r_3$}\label{fig:real_d_Robot_3_x35}
	\end{subfigure}
	\begin{subfigure}[b]{0.24\textwidth}
		\includegraphics[width=\textwidth]{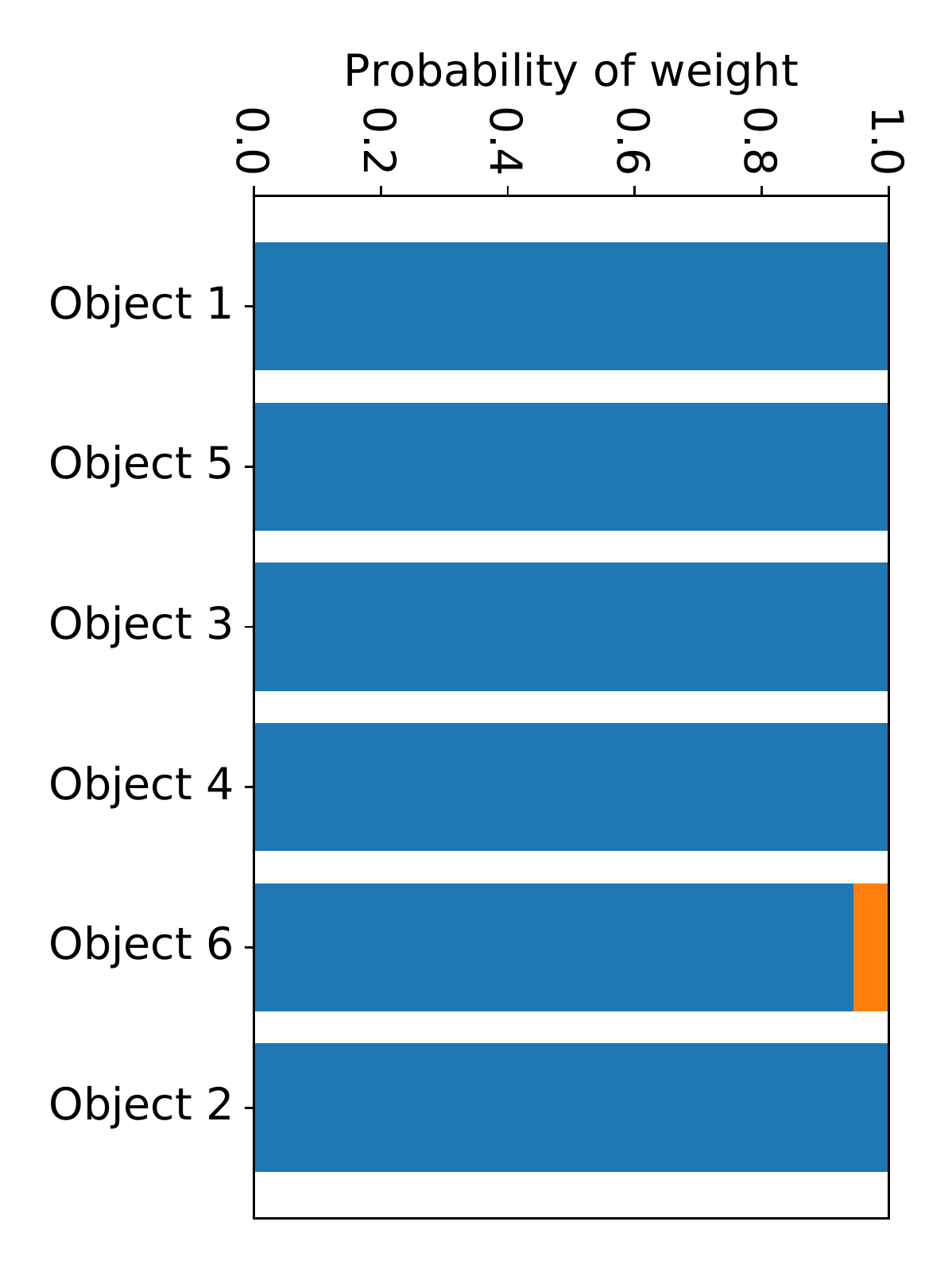}
		\caption{Dis. classification $r_3$}\label{fig:real_d_Robot_3_x35_cls}
	\end{subfigure}
	\begin{subfigure}[b]{0.24\textwidth}
		\includegraphics[width=\textwidth]{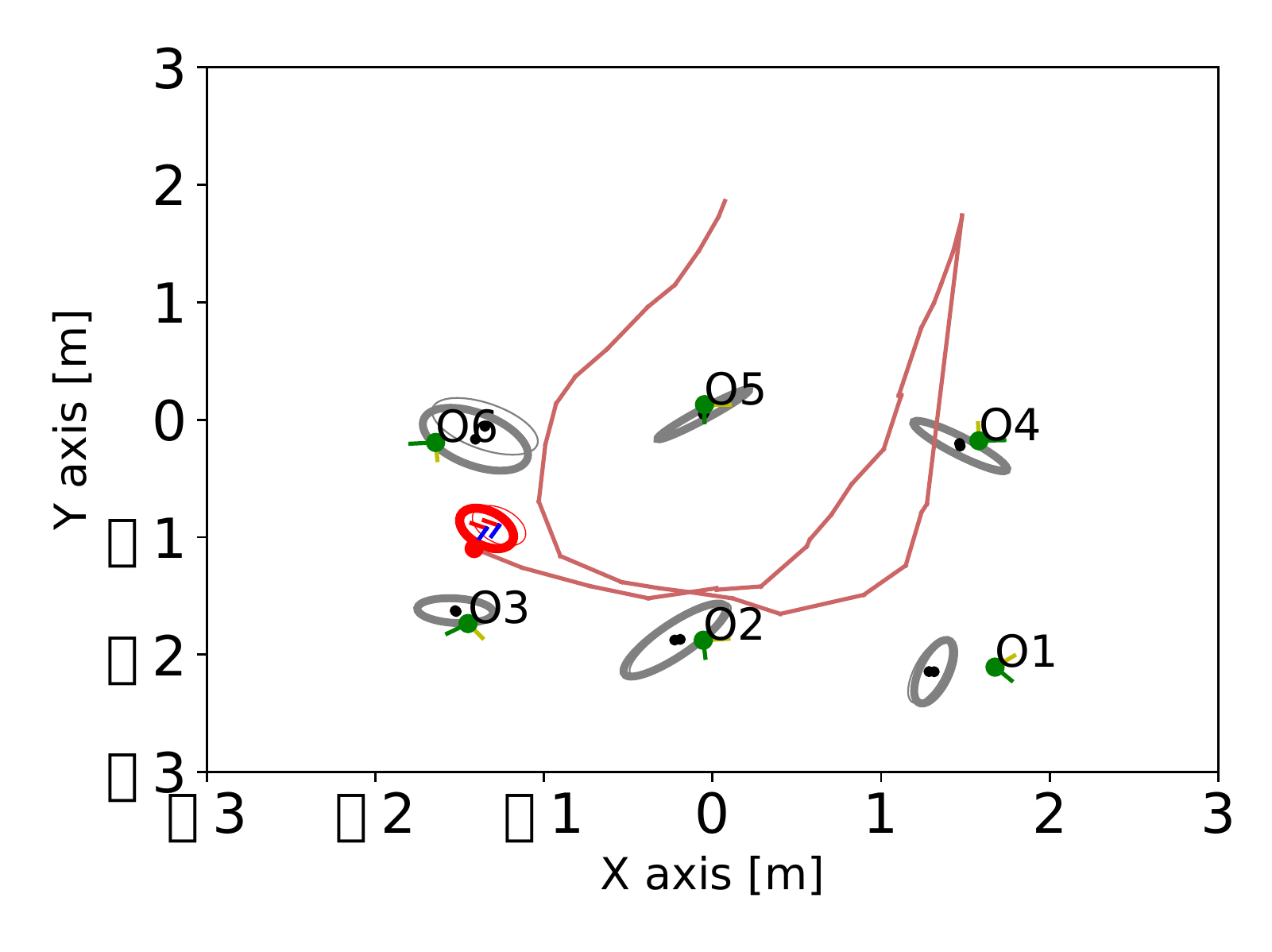}
		\caption{Dis. SLAM $r_1$}\label{fig:real_d_Robot_1_x40}
	\end{subfigure}
	\begin{subfigure}[b]{0.24\textwidth}
		\includegraphics[width=\textwidth]{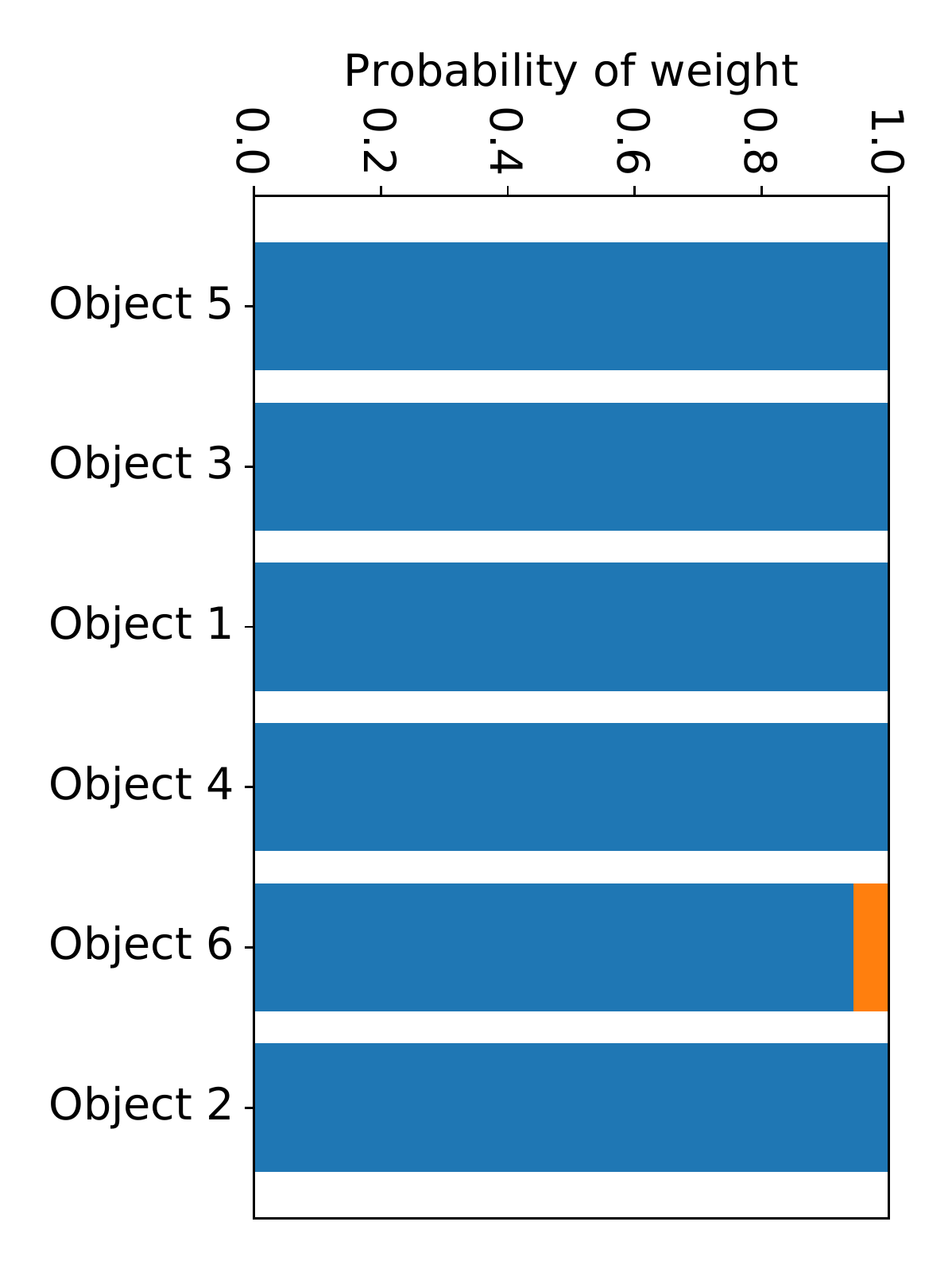}
		\caption{Dis. classification $r_1$}\label{fig:real_d_Robot_1_x40_cls}
	\end{subfigure}
	
	\caption{Figures for robot $r_3$ and $r_1$, distributed beliefs for time $k=35$ and $k=40$ respectively. \textbf{(a)} and \textbf{(b)} show results for $r_3$, \textbf{(c)} and \textbf{(d)} for $r_1$. \textbf{(a)} and \textbf{(c)} present SLAM results, \textbf{(b)} and \textbf{(d)} present classification results.}
	\label{fig:Figures_real_d_2}
\end{figure}

The results of all the graphs support the paper results as well, where both classification and SLAM in general are more accurate for the distributed belief. In addition, the robots inferring the distributed belief take into account objects that they didn't observe directly.

\begin{figure}[!htbp]
	
	\includegraphics[width=0.5\textwidth]{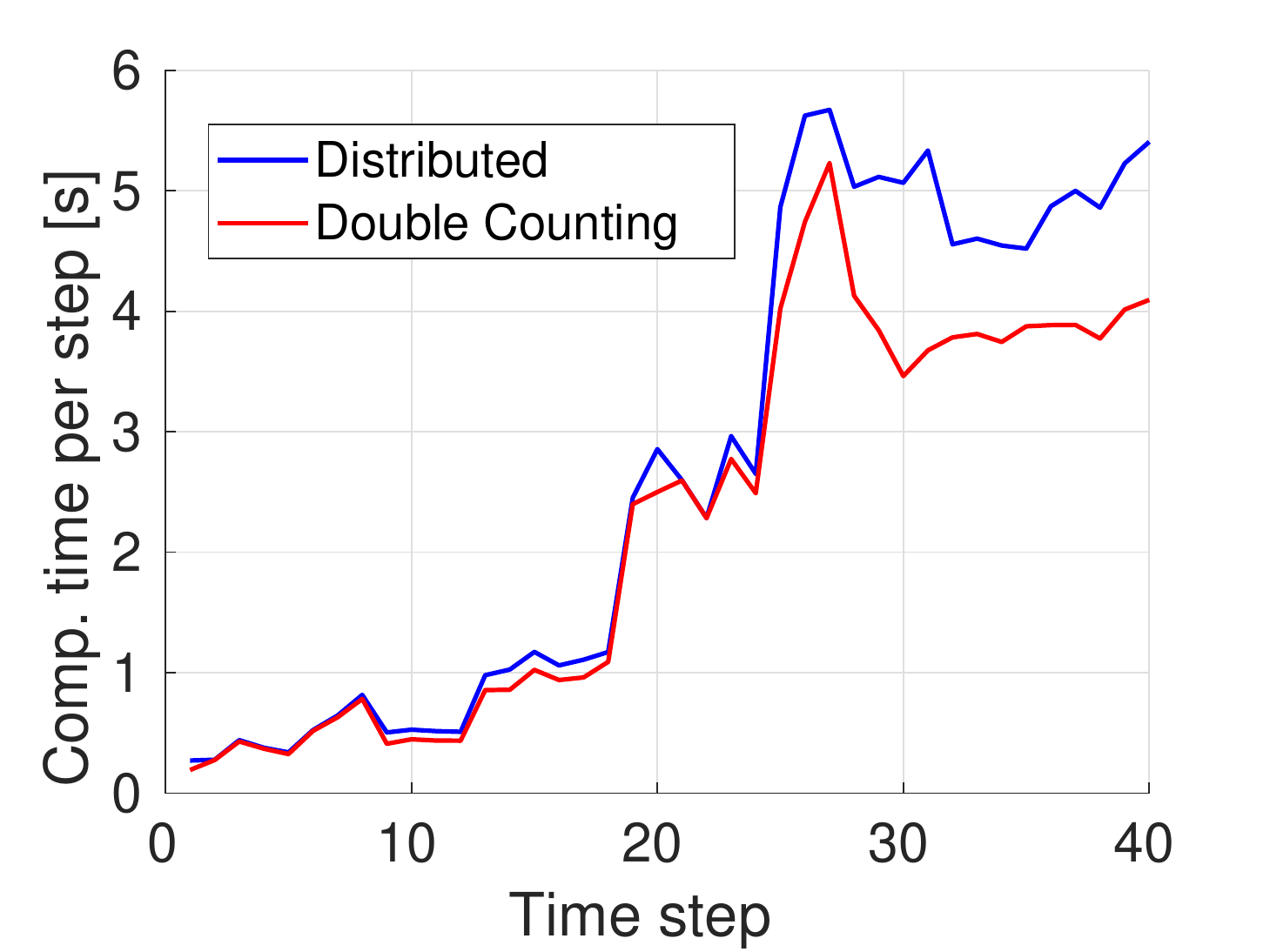}
	\caption{Calculation time as a function of the time step in seconds.}\label{fig:Time_Real}
	
\end{figure}

In Fig.~\ref{fig:Time_Real} we show the time each inference time-step takes to compute for the distributed case, without and with double-counting. In general, computation time is influenced by the number of class realizations that aren't pruned, and is higher when robots communicate between each other. For each newly observed object the algorithm must consider all realizations for the said object, thus the computation time "spikes" at the first step the new object is observed. Because the classifier model in the experiment uses deep neural networks, the computation is slower than in the simulation where hand crafted models were used.

	\bibliographystyle{unsrt}
	\bibliography{../../../../References/refs}